\documentclass{article}
\usepackage[utf8]{inputenc}
\usepackage[T1]{fontenc}

\usepackage{microtype}
\usepackage{subfig}
\usepackage{graphicx}
\usepackage{setspace}
\usepackage[normalem]{ulem}

\usepackage{booktabs} %

\usepackage{hyperref}

\usepackage{mathrsfs}

\usepackage[accepted]{icml2024}
\usepackage[noend]{algpseudocode}

\usepackage{microtype}
\usepackage{xcolor,colortbl}
\usepackage{natbib,bm}
\usepackage{amsmath,amsfonts,amssymb,amsthm}
\usepackage{mathtools}
\usepackage[capitalize]{cleveref}
\usepackage{tablefootnote}
\crefformat{footnote}{#2\footnotemark[#1]#3}
\Crefname{section}{Sec.}{Secs.}
\usepackage{pifont}
\usepackage{tabularx}
\usepackage{makecell,array,multirow,graphicx}
\usepackage{tikz,pgfplots}
\pgfplotsset{compat=1.18}
\usetikzlibrary{arrows,arrows.meta,positioning, fit,decorations.markings}
\usepackage{threeparttable}
\usepackage{booktabs} 

\definecolor{fhcolor}{rgb}{0.523, 0.235, 0.625}
\definecolor{tabblue}{HTML}{1F77B4}
\definecolor{taborange}{HTML}{FF7F0E}
\definecolor{tabred}{HTML}{D62728}
\definecolor{tabcyan}{HTML}{17BECF}
\definecolor{tabgreen}{HTML}{2CA02C}

\definecolor{ytcolor}{rgb}{1.0, 0.49, 0.0}

\newcommand{\cmark}{\ding{51}}%
\newcommand{\xmark}{\ding{55}}%

\newcommand{\lp}[2]{\left|\left|#1\right|\right|_{#2}}

\providecommand{\realnum}{\mathbb{R}}

\DeclareMathOperator*{\argmin}{arg\min}
\DeclareMathOperator*{\argmax}{arg\max}

\newcommand{\methodname}{\texttt{Charmer}}
\newcommand{\llama}{Llama 2}
\newcommand{\llamachat}{{Llama 2-Chat 7B}}

\newcolumntype{v}{>{\columncolor{blue!10}}c}
\newcolumntype{g}{>{\columncolor{black!10}}c}
\newcolumntype{G}{>{\columncolor{black!10}}l}

\theoremstyle{plain}
\newtheorem{theorem}{Theorem}[section]
\newtheorem{proposition}[theorem]{Proposition}

\newtheorem{corollary}[theorem]{Corollary}
\theoremstyle{definition}
\newtheorem{definition}[theorem]{Definition}

\theoremstyle{remark}
\newtheorem{remark}[theorem]{Remark}
\newtheorem{example}[theorem]{Example}

\icmltitlerunning{\methodname{}: Revisiting Character-level Adversarial Attacks for Language Models}

\begin{document}

\twocolumn[

\icmltitle{Revisiting Character-level Adversarial Attacks for Language Models}

\icmlsetsymbol{equal}{*}

\begin{icmlauthorlist}
\icmlauthor{Elias Abad Rocamora}{raclette}
\icmlauthor{Yongtao Wu}{raclette}
\icmlauthor{Fanghui Liu}{fishandchips}
\icmlauthor{Grigorios G. Chrysos}{burgersandguns}
\icmlauthor{Volkan Cevher}{raclette}
\end{icmlauthorlist}

\icmlaffiliation{raclette}{LIONS, École Polytechnique Fédérale de Lausanne, Switzerland}
\icmlaffiliation{fishandchips}{Department of Computer Science, University of Warwick, United Kingdom}
\icmlaffiliation{burgersandguns}{Department of Electrical and Computer Engineering, University of Wisconsin-Madison, USA}

\icmlcorrespondingauthor{Yongtao Wu}{yongtao.wu@epfl.ch}

\icmlkeywords{Machine Learning, ICML}

\vskip 0.3in
]

\printAffiliationsAndNotice{}  %

\begin{abstract}
Adversarial attacks in Natural Language Processing apply perturbations in the character or token levels. \emph{Token-level} attacks, gaining prominence for their use of gradient-based methods, are susceptible to altering sentence semantics, leading to invalid adversarial examples. %
While \emph{character-level} attacks easily maintain semantics, they have received less attention as they cannot easily adopt popular gradient-based methods, and are thought to be easy to defend. 
Challenging these beliefs, we introduce \methodname{}, an efficient query-based %
adversarial attack capable of achieving high attack success rate (ASR) while generating highly similar adversarial examples. Our method successfully targets both small (BERT) and large (\llama) models. %
Specifically, on BERT with SST-2, \methodname{} improves the ASR in $4.84\%$ points and the USE similarity in $8\%$ points with respect to the previous art. Our implementation is available in \href{https://github.com/LIONS-EPFL/Charmer}{github.com/LIONS-EPFL/Charmer}. %
\end{abstract}

\section{Introduction}
\label{sec:intro}
Language Models (LMs) have rapidly become the go-to tools for Natural Language Processing (NLP) tasks like language translation \citep{sutskever2014sequence}, code development \citep{chen2021copilot} and even general counseling via chat interfaces \citep{openai2023gpt4}. However, several failures concerning robustness to natural and adversarial noise have been discovered \citep{belinkov2018synthetic,alzantot2018generating}. %
Adversarial attacks have been widely adopted in the computer vision community %
to discover the worst-case performance of Machine Learning models \citep{Szegedy2014,Goodfellow2015} or %
be used to defend against such failure cases \citep{madry2018AT,Zhang2019TRADES}. 

The application of adversarial attacks in LMs is not straight-forward due to algorithmic \citep{guo2021gradient} and imperceptibility constraints \citep{morris2020reevaluating}.
Unlike the computer vision tasks, where inputs consist of tensors of real numbers, %
in NLP tasks, we work with sequences of discrete non-numerical inputs. This results in adversarial attacks %
being an NP-hard problem even for convex classifiers \citep{lei2019discrete}. %
This fact also hardens the use of popular gradient-based methods for obtaining adversarial examples \citep{guo2021gradient}. To tackle this problem, attackers adopt %
gradient based strategies in the embedding space, restricting the attack to the token vocabulary \citep{ebrahimi-etal-2018-hotflip,liu2022character,hou2023textgrad} or the \emph{black-box} setting, %
where only input-output access to the model is assumed \citep{alzantot2018generating,gao2018deepwordbug,Jin2020textfooler,li2020bertattack,garg2020bae,wallace2020imitationMT}.

\begin{figure}[!t]
    \centering
    \includegraphics[width=0.48\textwidth]{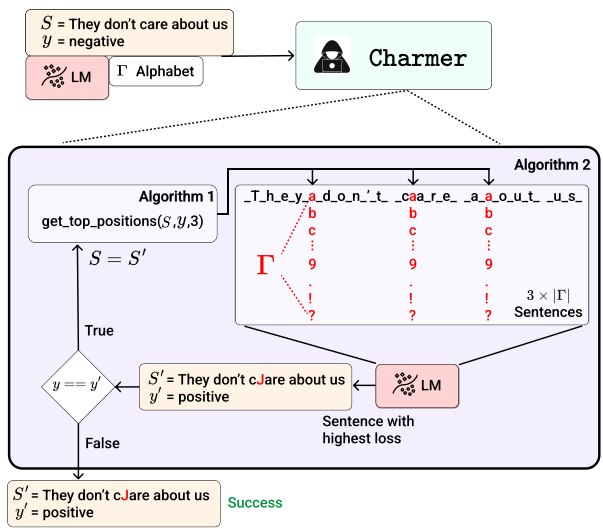}
    \vspace{-6mm}
    \caption{%
    \textbf{Schematic of the proposed method, \methodname{}:} Example of our attack in the sentiment classification task with the positions subset size $n=3$. %
    At each iteration, our attack computes the most important positions in the sentence via \cref{alg:position_selection}. Then, we generate all possible sentences replacing a character in the top positions, to get the one with the highest loss. If this sentence is misclassified, the process is finished. 
    }
    
    \label{fig:schematicllm}
        \vspace{-3mm}
\end{figure}

Another difficult analogy to make with the computer vision world is imperceptibility. Adversarial examples should be by definition imperceptible, in the sense that the attack should not modify the human prediction or allow to think that an attack has been done \citep{Szegedy2014}. Given an input $\bm{x} \in \realnum^{d}$, in the numerical-input setting, imperceptibility is controlled by looking for an adversarial example $\hat{\bm{x}}\in \realnum^d$ in an $\ell_p$ ball centered at $\bm x$ with perturbation radius $\epsilon$, i.e., $\lp{\bm{x} - \hat{\bm{x}}}{p} \leq \epsilon$, where $\epsilon$ can be set arbitrarily small. 

In NLP, \citet{morris2020reevaluating} suggest different strategies for controlling imperceptibility according to the attack level: \\
- \emph{Character:} Constrain the attack to have a low Levenshtein (edit) distance. However, character-level attacks have lost relevance due to the strength of robust word recognition defenses \citep{pruthi2019misspellings,jones2020robencodings}.\\
- \emph{Token:} Constrain the embedding similarity\footnote{Similarity is commonly measured via the cosine similarity of the USE embeddings \citep{cer2018USE}.} of replaced words and of the overall sentence to be high. Nevertheless, \citet{dyrmishi2023humans} conclude that  state-of-the-art attacks do not produce imperceptible attacks in practice. To be specific, \citet{hou2023textgrad} report $56.5 \%$ of their attacks change the semantics of the sentence.

The overall attack desiderata is summarized in \cref{tab:desiderata}.
Existing defenses against character-level attacks rely on robust word recognition modules, which assume the attacker adopts unrealistic constraints, not allowing simple modifications such as insertion or deletion of blank spaces, which are adopted in practice \citep{li2019textbugger}. In this work, we revisit character-level adversarial attacks as a practical solution to imperceptibility. %
Our attack, \methodname{}, is based on a greedy approach combined with a position subset selection to further speed-up the attack, while minimally affecting performance. Our strately is summerized in \cref{fig:schematicllm}.

Our attack is able to achieve $>95\%$ ASR (Attack success rate) in every studied TextAttack benchmark %
and LLMs Llama-2 and Vicuna, obtaining up to a $23\%$-point ASR improvement with respect to the runner-up method. We show that existing adversarial training based defenses \citep{hou2023textgrad} degrade character-level robustness, i.e., increasing the ASR in $3.32\%$ points when compared to standard training. Our findings indicate typo-corrector defenses \citep{pruthi2019misspellings,jones2020robencodings} are only successful when a set of strict attack constraints is assumed, if just one of these constraints is relaxed, ASR can increase from $0.96\%$ to $98.09\%$. Overall, we find that character-level robustness cannot be easily achieved with typo-correctors.%

\textbf{Notation:} Sentences are sequences of characters in the set $\Gamma$. Sentences are denoted with uppercase letters $S$. For sets of sentences we use caligraphic uppercase  letters $\mathcal{S}$. %
We denote the concatenation operator of two sentences as $\oplus$. The empty character $\emptyset$ is defined so that sentences remain invariant to concatenations with it, i.e., $S\oplus \emptyset = S$. We use the shorthand $[n]$ for $ \{ 1,2,\dots, n\}$ for any positive integer $n$. We use bold lowercase letters for vectors $\bm{x} \in \realnum^{d}$, with the $i^{\text{th}}$ position being $x_{i}\in \realnum$.

\begin{table}[t]
    \centering
    \caption{\textbf{Attack desiderata \& comparison with state-of-the-art.} ASR stands for Attack Success Rate. %
    }
    \vspace{2pt}
    \resizebox{0.48\textwidth}{!}{
    \begin{tabular}{l|c|cc}
        \toprule
        \multirow{3}{*}{Attack level} & Token & \multicolumn{2}{c}{Char}\\
        &  &  & \methodname{} \\
        &  & Previous & (This work)\\
        \midrule
        ASR(\%) & $\textcolor{tabgreen}{95.16}$ & $\textcolor{tabred}{0.96}^{*}$ & $\textcolor{tabgreen}{100.00}^{*}$ \\
        \midrule
        High efficiency & \textcolor{tabgreen}{\textbf{\cmark}} & \textcolor{tabgreen}{\textbf{\cmark}} & \textcolor{tabgreen}{\textbf{\cmark}} \\
        \midrule
        Semantics preserving & \textcolor{tabred}{\textbf{\xmark}}$^{**}$ & \textcolor{tabgreen}{\textbf{\cmark}} & \textcolor{tabgreen}{\textbf{\cmark}} \\
        \bottomrule
        \multicolumn{4}{l}{\begin{footnotesize}
            $^{*}$Defended with \citep{jones2020robencodings}.
        \end{footnotesize}}\\
        \multicolumn{4}{l}{\begin{footnotesize}
            $^{**}$According to \citep{hou2023textgrad,dyrmishi2023humans}.
        \end{footnotesize}}\\
    \end{tabular}}
    \vspace{-20pt}
    \label{tab:desiderata}
\end{table}

\section{Related Work}
\label{sec:related_work}

We provide an overview of adversarial attacks in NLP. %
Adversarial attacks have been devised for producing missclassifications \citep{alzantot2018generating}, generating unfaithful machine translations \citep{cheng2020seq2sick,sadrizadeh2023targetedMT,sadrizadeh2023transfool,wallace2020imitationMT}, providing malicious outputs (jailbreaking) \citep{zou2023universal,zhu2023autodan,carlini2023are} or even extracting training data \citep{nasr2023scalable}. We distinguish these methods in two main branches: \emph{token-level} and \emph{character-level} attacks.

\textbf{Token based:} Early token-based attacks rely in black-box token replacement/insertion strategies based on heuristics for estimating the importance of each position, and the candidate tokens for the operation \citep{ren2019PWWS,Jin2020textfooler,li2020bertattack,garg2020bae,lee2022query}. \citet{ebrahimi-etal-2018-hotflip} and \citet{li2019textbugger} consider the token gradient information to select which token to replace. \citet{guo2021gradient} propose GBDA, the first, but inefficient, full gradient based text adversarial attack. 
TextGrad \cite{hou2023textgrad} is a more efficient variante proposed to be integrated within Adversarial Training.

\textbf{Character based:} \citet{belinkov2018synthetic} showcase that character-level Machine Translation models are sensible to natural character level perturbations (typos) and adversarially chosen ones. \citep{pruthi2019misspellings} propose to iteratively change the best possible character until success. However, this strategy can be inefficient for lengthy sentences. Addressing this issue, other methods propose pre-filtering the most important words/tokens in the sentence, to then introduce a random typo \citep{gao2018deepwordbug}, or the best typo among a random sample \citep{li2019textbugger}. In \citep{liu2022character}, a token-based attack with character-level Levenshtein distance constraints is considered. However, considering token-level information for assesing character-level importance can be sub-optimal. \citet{ebrahimi-etal-2018-hotflip} propose involving embedding gradient information for evaluating the importance of characters, making the strategy only valid for character-level models. \citet{Yang2020greedy} evaluate the relevance of each character by masking and evaluating the loss in each position, to then modify the top positions. This strategy does not consider character insertions, and does not take into acount the effect of indivudual changes in the importance of positions. Similarly to \citep{Yang2020greedy}, our method measures the importance of every position plus insertions. After a perturbation is done, importances are updated to consider the interaction between perturbations.

\section{Problem Formulation}
\label{sec:back}
In this section, we present the problem formulation of our attack based on Levenshtein distance and the operators for characterizing perturbations.

\subsection{The Sentence Space}
\label{subsec:sentence_space}
Let $\Gamma$ %
be the alphabet set. A sentence $S$ with $l$ characters (i.e., the length $\left|S\right| = l$) in $\Gamma$ is defined with $S =c_1c_2\cdots c_l \in \Gamma^{l}$. 
For notational simplicity, we denote $S_i=c_i$ as the character in the $i^{\text{th}}$ position and $S_{i:}= c_i c_{i+1} \cdots c_l$ ($S_{:i}= c_1 c_{2} \cdots c_i$) as the sequence obtained by taking the characters after (before) the $i^{\text{th}}$ position included.
We denote $\mathcal{S}(\Gamma)$ as the set of (all possible) sequences with characters in $\Gamma$ with length less than $L$.
Let $d_{\text{lev}}: \mathcal{S}(\Gamma)\times \mathcal{S}(\Gamma) \to \realnum^+$ be the Levenshtein distance \citep{levenshtein1966binary}, also known as the edit distance. %
To be specific, for any two sentences $S, S' \in \mathcal{S}(\Gamma)$, the Levenshtein distance is defined as:
\[
    \resizebox{0.48\textwidth}{!}{$
    d_{\text{lev}}(S, S') := \left\{\begin{matrix}
        |S| & \quad \text{if}~|S'| = 0\\
        |S'| & \quad \text{if}~|S| = 0\\
        d_{\text{lev}}(S_{2:}, S'_{2:}) & \quad \text{if}~S_{1} = S'_{1}\\
        1+\min\left\{\begin{matrix}
            d_{\text{lev}}(S_{2:}, S'_{2:})\\
            d_{\text{lev}}(S_{2:}, S')\\
            d_{\text{lev}}(S, S'_{2:})\\
        \end{matrix}\right\}& \quad \text{otherwise}\;.\\
    \end{matrix}\right.
    $}
    \label{eq:lev}
\]
\begin{example}[$d_{\text{lev}}$ from $S = \text{Hello}$ to several modifications.]
    \[
    \begin{matrix}
        d_{\text{lev}}(\text{Hello}, \text{Helo}) = 1 &  d_{\text{lev}}(\text{Hello}, \text{Hallo}) = 1\\
        d_{\text{lev}}(\text{Hello}, \text{Helloo}) = 1 & d_{\text{lev}}(\text{Hello}, \text{Haloo}) = 2\,.
    \end{matrix}
    \]
\end{example}
Note that $d_{\text{lev}}$ represents the minimum %
cost in number of character \emph{insertions}, \emph{deletions} and \emph{replacements} needed for $S$ to become equal to $S'$ or vice-versa. %
\subsection{Adversarial Robustness}
In this work, we tackle robustness restricted by the Levenshtein distance. This enables the search of highly similar, hard to detect and semantics-preserving adversarial examples \citep{morris2020reevaluating}. %

\begin{definition}[$k$-robustness at $S$]
    \label{def:robustness}
    Denote the set of sentences at distance up to $k$ as
    \[
        \mathcal{S}_{k}(S,\Gamma) = \left\{S'\in \mathcal{S}(\Gamma) : d_{\text{lev}}(S, S') \leq k\right\}\,.
    \]
     A learning model (e.g., neural networks) $f : \mathcal{S}(\Gamma) \to \mathcal{Y}$, where $\mathcal{Y}$ is the label space, is called $k$-robust at $S$ if $f(S) = f(S'), ~~ \forall S'\in \mathcal{S}_{k}(S,\Gamma)$. If $f(S) \neq f(S')$ for some $S'\in \mathcal{S}_{k}(S,\Gamma)$, we say $S'$ is an \emph{adversarial example}.%
\end{definition}

Without loss of generality, we focus on the classification task. In the adversarial robustness community, adversarial examples are usually sought by solving some optimization problem \citep{carlini2017towards}. Given a data sample $(S,y) \in \mathcal{S}(\Gamma)\times[o]$ and a classification model $\bm{f} : \mathcal{S}(\Gamma) \to \realnum^{o}$, with $o$ classes, we solve:
\begin{equation}
    \max_{S'\in \mathcal{S}_{k}(S,\Gamma)}\mathcal{L}\left(\bm{f}(S'), y\right)\,,
    \label{eq:attack_problem}
    \vspace{-5pt}
\end{equation}
where $\mathcal{L}$ is some loss function, e.g., the cross entropy loss. In the following, we elaborate on how \cref{eq:attack_problem} is solved.
\subsection{Characterizing the Perturbations}

To explore adversarial examples in $\mathcal{S}_{k}(S,\Gamma)$, we make use of the contraction, expansion (\cref{def:dual_sentence}) and replacement operators (\cref{def:replace}) to characterize this set. %
\begin{definition}[Expansion and contraction operators]
\label{def:dual_sentence}
Let $\mathcal{S}(\Gamma)$ be the space of sentences with alphabet $\Gamma$ and the special character $\xi \notin \Gamma$, the pair of expansion-contraction functions $\phi : \mathcal{S}(\Gamma) \to \mathcal{S}(\Gamma \cup \{\xi\})$ and $\psi : \mathcal{S}(\Gamma \cup \{\xi\}) \to \mathcal{S}(\Gamma)$ is defined as:
\[
\begin{aligned}
    \phi(S) \coloneqq \left\{\begin{matrix}
        \xi & \text{if}~ |S| = 0\\
        \xi,S_1 \oplus \phi(S_{2:}) & \text{otherwise}
        \,.
    \end{matrix}\right.\\
    \psi(S) \coloneqq \left\{\begin{matrix}
        \emptyset & \text{if}~ |S| = 0\\
        \psi(S_{2:}) & \text{if}~ S_1 = \xi\\
        S_1 \oplus \psi(S_{2:}) & \text{otherwise}\,.
    \end{matrix}\right.
\end{aligned}
\]
\end{definition}
Clearly, $\phi(S)$ aims to insert $\xi$ into $S$ in all possible positions between characters and at the beginning and end of the sentence, and thus we have $|\phi(S)| = 2\cdot|S| + 1$.
Similarly, $\psi(S)$ aims to remove all $\xi$ occurred in $S$. 
The $(\phi,\psi)$ pair satisfies the property that $\psi(\phi(S)) = S$. 
We give the following example for a better understanding.
\begin{example}
Let $\xi :=\textcolor{tabred}{\mathbf{\bot}}$ for visibility:
$$
\resizebox{0.48\textwidth}{!}{$
    \begin{matrix}
        \phi(\text{Hello}) = \text{\textcolor{tabred}{$\mathbf{\bot}$}H\textcolor{tabred}{$\mathbf{\bot}$}e\textcolor{tabred}{$\mathbf{\bot}$}l\textcolor{tabred}{$\mathbf{\bot}$}l\textcolor{tabred}{$\mathbf{\bot}$}o\textcolor{tabred}{$\mathbf{\bot}$}} & \psi( \text{\textcolor{tabred}{$\mathbf{\bot}$}H\textcolor{tabred}{$\mathbf{\bot}$}eel\textcolor{tabred}{$\mathbf{\bot}$}l\textcolor{tabred}{$\mathbf{\bot}$}o\textcolor{tabred}{$\mathbf{\bot}$}}) =  \text{Heello}\\
        \psi(\text{\textcolor{tabred}{$\mathbf{\bot}$}H\textcolor{tabred}{$\mathbf{\bot}$}e\textcolor{tabred}{$\mathbf{\bot}$}l\textcolor{tabred}{$\mathbf{\bot}$}\textcolor{tabred}{$\mathbf{\bot}$}\textcolor{tabred}{$\mathbf{\bot}$}o\textcolor{tabred}{$\mathbf{\bot}$}}) = \text{Helo} & \psi( \text{\textcolor{tabred}{$\mathbf{\bot}$}H\textcolor{tabred}{$\mathbf{\bot}$}el\textcolor{tabred}{$\mathbf{\bot}$}lo\textcolor{tabred}{$\mathbf{\bot}$}}) =  \text{Hello}\,.\\
    \end{matrix}$}
$$
\end{example}

\begin{definition}[Replacement operator]
\label{def:replace}
Let $S \in \mathcal{S}(\Gamma \cup \{\xi\})$, the integer $i\in [|S|]$ and the character $c$, the replacement operator $\overset{i}{\leftarrow} c$ of the $i^{\text{th}}$ position of $S$ with $c$ is defined as:
\vspace{-5pt}
\[
	S \overset{i}{\leftarrow} c \coloneqq S_{:i-1} \oplus c \oplus S_{i+1:}
\]
\end{definition}
\vspace{-10pt}
Thanks to \cref{def:dual_sentence,def:replace}, it is easy to check the following proposition.
\begin{proposition}[Characterization of $d_{\text{lev}}$-$1$ operations]
	Let $S \in \mathcal{S}(\Gamma)$ be a non-empty sentence, and $S'$ be another sentence satisfying $d_{\text{lev}}(S,S')=1$. Then we can find $i \in [2|S| + 1]$ and a character $c \in \Gamma\cup \{\xi\}$ such that
	\[
	S' = \psi\left(\phi\left(S\right)\overset{i}{\leftarrow} c\right)\,.
        \]
\label{prop:char_d1}
\end{proposition}
\begin{remark}[Non-uniqueness]
	The parametrization of the transformation from $S$ to $S'$ might not be unique. For example, for $S = \text{Hello}$ and $S' = \text{Helo}$, both pairs $(i = 6,c = \xi)$ and $(i = 8,c = \xi)$ are valid parametrizations.
\end{remark}
\begin{remark}[Intuition]
    Replacing a character in $\Gamma$ for $\xi$, and applying $\psi$ results in a deletion. Similarly, replacing a $\xi$ by a character in $\Gamma$ and applying $\xi$ results in an insertion.
\end{remark}
\begin{corollary}[Generating $\mathcal{S}_k$]\label{coro:gen}
	Let $S$ be a non-empty sentence in the volcabulary, with $|\Gamma|>1$, for any $k \geq 1$, the set $\mathcal{S}_k(S,\Gamma)$ (see \cref{def:robustness}) can be obtained by the following recursion: %
 \vspace{-4pt}
	\[
	\resizebox{0.48\textwidth}{!}{$
	\mathcal{S}_k(S,\Gamma) = \left\{\begin{matrix}
		\left\{ \psi\left(\phi\left(S\right)\overset{i}{\leftarrow} c\right) ~~\begin{matrix}
		    \forall i \in [2|S| + 1]\\
                \forall c \in \Gamma \cup \{\xi\}
		\end{matrix} \right\} & ,\text{if}~ k=1,\\
		\left\{ \psi\left(\phi\left(\hat{S}\right)\overset{i}{\leftarrow} c\right) ~~ \begin{matrix}
		    \forall i \in [2|\hat{S}| + 1]\\
                \forall c \in \Gamma \cup \{\xi\}\\
                \forall \hat{S} \in \mathcal{S}_{k-1}(S,\Gamma)
		\end{matrix} \right\} & ,\text{if}~ k>1\,.\\
	\end{matrix}\right.$}
 \vspace{-4pt}
	\]
	The size of these sets is bounded as:
	\[
		\frac{|\Gamma|^{k+1}-1}{|\Gamma|-1}\leq |\mathcal{S}_{k}(S,\Gamma)| \leq (|\Gamma|+1)^{k}\cdot \prod_{j=1}^{k}(2(|S| + k)-1)\,.
	\]
    \label{cor:Sk}
\end{corollary}
\vspace{-10pt}
\begin{remark}
    In the case $|\Gamma| = 1$, for any $S\in \mathcal{S}(\Gamma): |S|\geq k$, we trivially have that $|\mathcal{S}_k(S,\Gamma)| = 2k + 1$.
\end{remark}
\emph{Proof.} Refer to \cref{app:method}. %

Note that exactly computing $|\mathcal{S}_{k}(S,\Gamma)|$ is non-trivial and complex dynamic programming algorithms have been proposed for this task \citep{mihov2004fast,mitankin2005universal,touzet2016levenshtein}. The exponential dependence of $|\mathcal{S}_k(S,\Gamma)|$ on $k$ makes it unfeasible to evaluate every sentence in the set, therefore, smarter strategies are needed.

\section{Method}
\label{sec:method}
\begin{algorithm}[!t]
	\caption{Heuristic for Top-$n$ position selection.%
 } 
	\label{alg:position_selection}
	\begin{spacing}{1.3}
	\begin{algorithmic}[1]

	\State \textbf{Inputs:} model $f$, sentence $S$, test character $t$, special character $\xi$, number of positions $n$, loss $\mathcal{L}$ and label $y$. %
        \State $\bm{l} = \bm{0}$ \Comment{\textcolor{teal}{Initialize vector to store losses with zeros}}
	\For {$i=1,\dots, 2|S| + 1$}
            \State $P = \phi(S)$ \Comment{\textcolor{teal}{Expand sentence}}
            \If{$P_{i} = t$} $P_{i} \leftarrow \xi$
            \Else{} $P_{i} \leftarrow t$ \EndIf
            \State $l_{i} = \mathcal{L}(f(\psi(P)),y)$ \Comment{\textcolor{teal}{Contract and eval. loss}}
	\EndFor
        \State \Return $\text{Top-}n(\bm{l})$ \Comment{\textcolor{teal}{Indexes of the top $n$ values in $\bm{l}$}}
	\end{algorithmic} 
	\end{spacing}

\end{algorithm}
Let us now introduce our method (\methodname{}). %
In \cref{subsubsec:attack_classifier,subsubsec:attack_llm} we present our attack for both standard classifiers and LLM-based classifiers. %

To circumvent the exponential dependence of $|\mathcal{S}_k(S,\Gamma)|$ on $k$ as indicated by \cref{coro:gen}, we propose to greedily select the single-character perturbation with the highest loss. Furthermore, we reduce the search space for single-character perturbations by considering a subset of locations where characters can be replaced. %

\subsection{Pre-selection of Replacement Locations}
\label{subsubsec:position_selecion}
In \cref{prop:char_d1}, we consider all possible locations $(i\in [2\cdot|S| + 1])$, leading to $|\mathcal{S}_1(S,\Gamma)| \leq (|\Gamma| + 1)\cdot(2\cdot|S| + 1)$ by \cref{coro:gen}, which can be relatively big for lengthy sentences (e.g., $|S|$ can be up to $844$ for AG-News). We propose considering a subset of $n$ locations in order to remove the dependency on the length of the sentence. %
To select the top $n$ locations, we propose testing the relevance of each position by replacing each character with a ``test" character 
and looking at the change in the loss. In \cref{alg:position_selection} we formalize our proposed strategy. Note that if the test character is going to be replaced by itself in a certain position, we replace by the special character $\xi$ (Line 5 in \cref{alg:position_selection}). In practice we use the white space (U+0020) as the test character. Overall, \cref{alg:position_selection} performs $\mathcal{O}(|S|)$ forward passes trough the language model.

\begin{table*}[!ht]
    \centering
    \caption{\textbf{Attack evaluation in the TextAttack BERT and RoBERTa models:} Token-level and character-level attacks are highlighted with \textcolor{tabcyan}{\ding{108}} and \textcolor{tabred}{\ding{108}} respectively. For each metric, the best method is highlighted in \textbf{bold} and the runner-up is \underline{underlined}. \methodname{} consistently achieves highest Attack Success Rate (ASR)%
    Additionally, the similarity between the original and attacked sentences is the highest or runner-up in 8/10 cases. %
    }
    \vspace{2pt}
    \setlength{\tabcolsep}{0.2\tabcolsep}
    \resizebox{\textwidth}{!}{
    \begin{tabular}{cg|vggg|vggg}
        \toprule
        \rowcolor{white}
            & & \multicolumn{4}{c}{BERT} & \multicolumn{4}{c}{RoBERTa} \\
        \rowcolor{white}
        & Method & ASR (\%) $\bm{\uparrow}$ & $d_{\text{lev}}(S,S')~ \bm{\downarrow}$ & $\text{Sim}(S,S')~ \bm{\uparrow}$ & $\text{Time (s)} ~ \bm{\downarrow}$ & ASR (\%) $\bm{\uparrow}$ & $d_{\text{lev}}(S,S')~ \bm{\downarrow}$ & $\text{Sim}(S,S')~ \bm{\uparrow}$ & $\text{Time (s)} ~ \bm{\downarrow}$\\
        \midrule
        \parbox[t]{3mm}{\multirow{9}{*}{\rotatebox[origin=c]{90}{AG-News}}}
        & GBDA \textcolor{tabcyan}{\ding{108}} & $42.09$ & $17.76_{\pm (9.33)}$ & $0.93_{\pm (0.05)}$ & $13.86_{\pm (3.14)}$ & - & - & - & - \\
        & \cellcolor{white} BAE-R \textcolor{tabcyan}{\ding{108}} & \cellcolor{white} $17.09$ & \cellcolor{white} $15.07_{\pm (10.59)}$ & \cellcolor{white} $\mathbf{0.97}_{\pm (0.02)}$ & \cellcolor{white} $1.61_{\pm (1.36)}$ & \cellcolor{white} $18.27$ & \cellcolor{white} $15.29_{\pm (10.34)}$ & \cellcolor{white} $\mathbf{0.97}_{\pm (0.02)}$ & \cellcolor{white} $2.14_{\pm (1.81)}$ \\
        & BERT-attack \textcolor{tabcyan}{\ding{108}} & $29.90$ & $20.66_{\pm (16.91)}$ & $0.93_{\pm (0.05)}$ & $5.58_{\pm (12.92)}$ & $27.55$ & $16.96_{\pm (12.95)}$ & $0.94_{\pm (0.04)}$ & $\underline{1.44}_{\pm (1.76)}$ \\
        & \cellcolor{white} DeepWordBug \textcolor{tabred}{\ding{108}} & \cellcolor{white} $60.51$ & \cellcolor{white} $11.75_{\pm (8.00)}$ & \cellcolor{white} $0.78_{\pm (0.18)}$ & \cellcolor{white} $\mathbf{0.81}_{\pm (0.52)}$ & \cellcolor{white} $56.81$ & \cellcolor{white} $11.81_{\pm (7.69)}$ & \cellcolor{white} $0.79_{\pm (0.16)}$ & \cellcolor{white} $\mathbf{0.69}_{\pm (0.35)}$ \\
        & TextBugger \textcolor{tabred}{\ding{108}} & $50.85$ & $19.79_{\pm (17.93)}$ & $0.90_{\pm (0.06)}$ & $\underline{1.53}_{\pm (1.13)}$ & $51.21$ & $21.42_{\pm (19.28)}$ & $0.90_{\pm (0.06)}$ & $2.30_{\pm (1.61)}$ \\
        & \cellcolor{white} TextFooler \textcolor{tabcyan}{\ding{108}} & \cellcolor{white} $78.98$ & \cellcolor{white} $53.18_{\pm (39.30)}$ & \cellcolor{white} $0.84_{\pm (0.11)}$ & \cellcolor{white} $3.75_{\pm (2.76)}$ & \cellcolor{white} $84.48$ & \cellcolor{white} $52.45_{\pm (36.97)}$ & \cellcolor{white} $0.84_{\pm (0.11)}$ & \cellcolor{white} $3.84_{\pm (2.77)}$ \\
        & TextGrad \textcolor{tabcyan}{\ding{108}} & $85.85$ & $55.38_{\pm (30.33)}$ & $0.77_{\pm (0.11)}$ & $7.98_{\pm (9.24)}$ & $78.75$ & $31.94_{\pm (15.57)}$ & $0.86_{\pm (0.07)}$ & $9.86_{\pm (9.74)}$ \\
        & \cellcolor{white} CWBA \textcolor{tabred}{\ding{108}} & \cellcolor{white} $86.72$ & \cellcolor{white} $15.71_{\pm (7.17)}$ & \cellcolor{white} $0.65_{\pm (0.19)}$ & \cellcolor{white} $174.15_{\pm (130.91)}$ & \cellcolor{white} $81.39$ & \cellcolor{white} $13.73_{\pm (11.24)}$ & \cellcolor{white} $0.86_{\pm (0.11)}$ & \cellcolor{white} $55.33_{\pm (43.19)}$ \\
        & \citep{pruthi2019misspellings} \textcolor{tabred}{\ding{108}} & $90.02$ & $6.25_{\pm (4.69)}$ & $0.86_{\pm (0.14)}$ & $49.47_{\pm (48.26)}$ & $88.91$ & $6.55_{\pm (5.13)}$ & $0.86_{\pm (0.14)}$ & $29.75_{\pm (24.53)}$ \\
        \rowcolor{white} & \methodname{}\texttt{-Fast} (Ours) \textcolor{tabred}{\ding{108}} & $\underline{95.86}$ & $\underline{4.85}_{\pm (3.96)}$ & $0.92_{\pm (0.08)}$ & $3.12_{\pm (3.88)}$ & $\underline{91.87}$ & $\underline{4.87}_{\pm (4.07)}$ & $0.91_{\pm (0.09)}$ & $3.15_{\pm (3.83)}$ \\
        & \methodname{} (Ours) \textcolor{tabred}{\ding{108}} & $\mathbf{98.51}$ & $\mathbf{3.68}_{\pm (3.08)}$ & $\underline{0.95}_{\pm (0.06)}$ & $8.74_{\pm (11.10)}$ & $\mathbf{96.88}$ & $\mathbf{3.73}_{\pm (3.07)}$ & $\underline{0.95}_{\pm (0.05)}$ & $9.45_{\pm (11.20)}$ \\
        \midrule
        \parbox[t]{2mm}{\multirow{8}{*}{\rotatebox[origin=c]{90}{MNLI-m}}} & \cellcolor{white} GBDA \textcolor{tabcyan}{\ding{108}} & \cellcolor{white} $\underline{97.97}$ & \cellcolor{white} $11.45_{\pm (6.52)}$ & \cellcolor{white} $0.73_{\pm (0.16)}$ & \cellcolor{white} $11.68_{\pm (3.23)}$ & \cellcolor{white} - & \cellcolor{white} - & \cellcolor{white} - & \cellcolor{white} - \\
        & BAE-R \textcolor{tabcyan}{\ding{108}} & $70.00$ & $6.46_{\pm (3.32)}$ & $\underline{0.83}_{\pm (0.15)}$ & $0.53_{\pm (0.44)}$ & $67.39$ & $6.47_{\pm (3.31)}$ & $\mathbf{0.84}_{\pm (0.14)}$ & $0.54_{\pm (0.37)}$ \\
        & \cellcolor{white} BERT-attack \textcolor{tabcyan}{\ding{108}} & \cellcolor{white} $92.41$ & \cellcolor{white} $6.95_{\pm (6.57)}$ & \cellcolor{white} $\underline{0.83}_{\pm (0.13)}$ & \cellcolor{white} $26.53_{\pm (204.23)}$ & \cellcolor{white} $\underline{97.62}$ & \cellcolor{white} $6.56_{\pm (4.16)}$ & \cellcolor{white} $\underline{0.83}_{\pm (0.14)}$ & \cellcolor{white} $2.60_{\pm (15.58)}$ \\
        & DeepWordBug \textcolor{tabred}{\ding{108}} & $84.88$ & $2.30_{\pm (1.68)}$ & $0.75_{\pm (0.18)}$ & $\underline{0.23}_{\pm (0.12)}$ & $78.41$ & $2.79_{\pm (2.02)}$ & $0.71_{\pm (0.21)}$ & $\underline{0.27}_{\pm (0.15)}$ \\
        & \cellcolor{white} TextBugger \textcolor{tabred}{\ding{108}} & \cellcolor{white} $85.36$ & \cellcolor{white} $4.17_{\pm (4.33)}$ & \cellcolor{white} $\underline{0.83}_{\pm (0.13)}$ & \cellcolor{white} $0.44_{\pm (0.32)}$ & \cellcolor{white} $86.36$ & \cellcolor{white} $5.41_{\pm (5.40)}$ & \cellcolor{white} $0.80_{\pm (0.15)}$ & \cellcolor{white} $0.51_{\pm (0.38)}$ \\
        & TextFooler \textcolor{tabcyan}{\ding{108}} & $92.26$ & $9.83_{\pm (6.87)}$ & $0.82_{\pm (0.14)}$ & $0.52_{\pm (0.41)}$ & $90.23$ & $10.50_{\pm (7.79)}$ & $0.81_{\pm (0.14)}$ & $0.54_{\pm (0.42)}$ \\
        & \cellcolor{white} TextGrad \textcolor{tabcyan}{\ding{108}} & \cellcolor{white} $93.69$ & \cellcolor{white} $9.98_{\pm (5.66)}$ & \cellcolor{white} $0.75_{\pm (0.13)}$ & \cellcolor{white} $2.50_{\pm (1.97)}$ & \cellcolor{white} $95.44$ & \cellcolor{white} $9.10_{\pm (5.30)}$ & \cellcolor{white} $0.79_{\pm (0.12)}$ & \cellcolor{white} $3.56_{\pm (2.83)}$ \\
        & \citep{pruthi2019misspellings} \textcolor{tabred}{\ding{108}} & $57.62$ & $1.32_{\pm (0.64)}$ & $0.83_{\pm (0.12)}$ & $4.48_{\pm (3.73)}$ & $52.84$ & $\underline{1.36}_{\pm (0.63)}$ & $0.82_{\pm (0.13)}$ & $7.48_{\pm (6.49)}$ \\
        \rowcolor{white} & \methodname{}\texttt{-Fast} (Ours) \textcolor{tabred}{\ding{108}} & $\mathbf{100.00}$ & $\underline{1.23}_{\pm (0.58)}$ & $\mathbf{0.85}_{\pm (0.14)}$ & $\mathbf{0.21}_{\pm (0.17)}$ & $\mathbf{100.00}$ & $\underline{1.36}_{\pm (0.78)}$ & $0.82_{\pm (0.15)}$ & $\mathbf{0.23}_{\pm (0.19)}$ \\
        & \methodname{} (Ours) \textcolor{tabred}{\ding{108}} & $\mathbf{100.00}$ & $\mathbf{1.14}_{\pm (0.42)}$ & $\mathbf{0.85}_{\pm (0.13)}$ & $1.45_{\pm (0.81)}$ & $\mathbf{100.00}$ & $\mathbf{1.17}_{\pm (0.46)}$ & $\mathbf{0.84}_{\pm (0.13)}$ & $1.49_{\pm (0.82)}$ \\
        \midrule
        \parbox[t]{2mm}{\multirow{8}{*}{\rotatebox[origin=c]{90}{QNLI}}}  & \cellcolor{white} GBDA \textcolor{tabcyan}{\ding{108}} & \cellcolor{white} $47.16$ & \cellcolor{white} $11.88_{\pm (7.02)}$ & \cellcolor{white} $0.93_{\pm (0.06)}$ & \cellcolor{white} $13.85_{\pm (2.94)}$ & \cellcolor{white} - & \cellcolor{white} - & \cellcolor{white} - & \cellcolor{white} - \\
        & BAE-R \textcolor{tabcyan}{\ding{108}} & $40.04$ & $11.44_{\pm (8.30)}$ & $\mathbf{0.95}_{\pm (0.07)}$ & $2.31_{\pm (2.36)}$ & $41.66$ & $10.37_{\pm (9.05)}$ & $\mathbf{0.96}_{\pm (0.04)}$ & $2.18_{\pm (2.77)}$ \\
        &\cellcolor{white} BERT-attack \textcolor{tabcyan}{\ding{108}} & \cellcolor{white} $70.21$ & \cellcolor{white} $16.21_{\pm (12.44)}$ & \cellcolor{white} $0.90_{\pm (0.08)}$ & \cellcolor{white} $239.86_{\pm (1395.15)}$ & \cellcolor{white} $70.65$ & \cellcolor{white} $17.78_{\pm (13.28)}$ & \cellcolor{white} $0.89_{\pm (0.12)}$ & \cellcolor{white} $2.70_{\pm (12.58)}$ \\
        & DeepWordBug \textcolor{tabred}{\ding{108}} & $71.57$ & $4.52_{\pm (4.04)}$ & $0.86_{\pm (0.15)}$ & $\mathbf{0.50}_{\pm (0.34)}$ & $64.34$ & $5.07_{\pm (4.67)}$ & $0.85_{\pm (0.17)}$ & $\mathbf{0.59}_{\pm (0.41)}$ \\
        &\cellcolor{white} TextBugger \textcolor{tabred}{\ding{108}} & \cellcolor{white} $75.77$ & \cellcolor{white} $8.16_{\pm (9.93)}$ & \cellcolor{white} $0.89_{\pm (0.10)}$ & \cellcolor{white} $\underline{0.99}_{\pm (0.78)}$ & \cellcolor{white} $67.39$ & \cellcolor{white} $9.08_{\pm (10.31)}$ & \cellcolor{white} $0.90_{\pm (0.10)}$ & \cellcolor{white} $\underline{0.90}_{\pm (0.72)}$ \\
        & TextFooler \textcolor{tabcyan}{\ding{108}} & $80.64$ & $23.42_{\pm (21.56)}$ & $0.87_{\pm (0.12)}$ & $1.90_{\pm (1.70)}$ & $76.01$ & $25.74_{\pm (28.53)}$ & $0.87_{\pm (0.12)}$ & $2.00_{\pm (2.09)}$ \\
        & \cellcolor{white} TextGrad \textcolor{tabcyan}{\ding{108}} & \cellcolor{white} $77.35$ & \cellcolor{white} $30.03_{\pm (20.41)}$ & \cellcolor{white} $0.82_{\pm (0.10)}$ & \cellcolor{white} $4.54_{\pm (3.82)}$ & \cellcolor{white} $76.80$ & \cellcolor{white} $21.56_{\pm (15.74)}$ & \cellcolor{white} $0.87_{\pm (0.08)}$ & \cellcolor{white} $5.80_{\pm (4.58)}$ \\
        & \citep{pruthi2019misspellings} \textcolor{tabred}{\ding{108}} & $17.70$ & $\mathbf{1.57}_{\pm (0.81)}$ & $0.93_{\pm (0.07)}$ & $7.22_{\pm (4.91)}$ & $17.45$ & $\mathbf{1.54}_{\pm (0.88)}$ & $\underline{0.93}_{\pm (0.08)}$ & $7.46_{\pm (5.33)}$ \\
        \rowcolor{white}& \methodname{}\texttt{-Fast} (Ours) \textcolor{tabred}{\ding{108}} & $\underline{94.69}$ & $2.21_{\pm (1.69)}$ & $0.93_{\pm (0.09)}$ & $1.33_{\pm (1.55)}$ & $\underline{96.95}$ & $2.73_{\pm (2.15)}$ & $0.90_{\pm (0.12)}$ & $1.72_{\pm (2.27)}$ \\
        & \methodname{} (Ours) \textcolor{tabred}{\ding{108}} & $\mathbf{97.68}$ & $\underline{1.94}_{\pm (1.48)}$ & $\underline{0.94}_{\pm (0.07)}$ & $9.19_{\pm (9.60)}$ & $\mathbf{97.86}$ & $\underline{2.20}_{\pm (1.69)}$ & $0.92_{\pm (0.08)}$ & $10.55_{\pm (9.69)}$ \\
        \midrule
        \parbox[t]{2mm}{\multirow{8}{*}{\rotatebox[origin=c]{90}{RTE}}} & \cellcolor{white} GBDA \textcolor{tabcyan}{\ding{108}} & \cellcolor{white} $76.62$ & \cellcolor{white} $8.99_{\pm (4.74)}$ & \cellcolor{white} $0.78_{\pm (0.13)}$ & \cellcolor{white} $16.71_{\pm (7.29)}$ & \cellcolor{white} - & \cellcolor{white} - & \cellcolor{white} - & \cellcolor{white} - \\
        & BAE-R \textcolor{tabcyan}{\ding{108}} & $64.68$ & $6.98_{\pm (3.39)}$ & $\underline{0.87}_{\pm (0.09)}$ & $0.84_{\pm (0.65)}$ & $64.06$ & $6.11_{\pm (3.84)}$ & $\mathbf{0.89}_{\pm (0.08)}$ & $0.76_{\pm (0.69)}$ \\
        & \cellcolor{white} BERT-attack \textcolor{tabcyan}{\ding{108}} & \cellcolor{white} $68.00$ & \cellcolor{white} $10.06_{\pm (9.75)}$ & \cellcolor{white} $0.78_{\pm (0.19)}$ & \cellcolor{white} $23.67_{\pm (51.78)}$ & \cellcolor{white} $31.34$ & \cellcolor{white} $6.00_{\pm (3.96)}$ & \cellcolor{white} $0.86_{\pm (0.08)}$ & \cellcolor{white} $5.36_{\pm (20.47)}$ \\
        & DeepWordBug \textcolor{tabred}{\ding{108}} & $65.67$ & $1.64_{\pm (0.82)}$ & $0.85_{\pm (0.10)}$ & $\mathbf{0.12}_{\pm (0.03)}$ & $62.67$ & $1.83_{\pm (1.09)}$ & $0.82_{\pm (0.14)}$ & $\mathbf{0.13}_{\pm (0.04)}$ \\
        & \cellcolor{white} TextBugger \textcolor{tabred}{\ding{108}} & \cellcolor{white} $74.13$ & \cellcolor{white} $3.38_{\pm (3.48)}$ & \cellcolor{white} $\mathbf{0.88}_{\pm (0.09)}$ & \cellcolor{white} $0.35_{\pm (0.18)}$ & \cellcolor{white} $71.43$ & \cellcolor{white} $3.91_{\pm (3.73)}$ & \cellcolor{white} $\mathbf{0.89}_{\pm (0.08)}$ & \cellcolor{white} $\underline{0.38}_{\pm (0.40)}$ \\
        & TextFooler \textcolor{tabcyan}{\ding{108}} & $79.60$ & $7.42_{\pm (6.10)}$ & $\mathbf{0.88}_{\pm (0.09)}$ & $0.47_{\pm (0.59)}$ & $74.19$ & $7.68_{\pm (5.94)}$ & $\underline{0.88}_{\pm (0.09)}$ & $0.47_{\pm (0.59)}$ \\
        & \cellcolor{white} TextGrad \textcolor{tabcyan}{\ding{108}} & \cellcolor{white} $81.77$ & \cellcolor{white} $10.02_{\pm (5.69)}$ & \cellcolor{white} $0.76_{\pm (0.13)}$ & \cellcolor{white} $2.44_{\pm (1.06)}$ & \cellcolor{white} $73.97$ & \cellcolor{white} $6.40_{\pm (3.30)}$ & \cellcolor{white} $0.84_{\pm (0.08)}$ & \cellcolor{white} $3.57_{\pm (3.61)}$ \\
        & \citep{pruthi2019misspellings} \textcolor{tabred}{\ding{108}} & $62.19$ & $\mathbf{1.18}_{\pm (0.41)}$ & $0.86_{\pm (0.08)}$ & $8.45_{\pm (6.25)}$ & $49.31$ & $\mathbf{1.21}_{\pm (0.54)}$ & $0.87_{\pm (0.08)}$ & $12.23_{\pm (8.84)}$ \\
        \rowcolor{white} & \methodname{}\texttt{-Fast} (Ours) \textcolor{tabred}{\ding{108}} & $\underline{89.55}$ & $\underline{1.36}_{\pm (0.93)}$ & $\underline{0.87}_{\pm (0.12)}$ & $\underline{0.29}_{\pm (0.27)}$ & $\underline{91.71}$ & $1.78_{\pm (1.71)}$ & $0.82_{\pm (0.15)}$ & $0.41_{\pm (0.54)}$ \\
        & \methodname{} (Ours) \textcolor{tabred}{\ding{108}} &  $\mathbf{97.01}$ & $1.55_{\pm (1.42)}$ & $0.86_{\pm (0.13)}$ & $2.50_{\pm (2.33)}$ & $\mathbf{97.24}$ & $\underline{1.61}_{\pm (1.39)}$ & $0.85_{\pm (0.13)}$ & $2.74_{\pm (2.87)}$ \\
        \midrule
        \parbox[t]{2mm}{\multirow{9}{*}{\rotatebox[origin=c]{90}{SST-2}}} & \cellcolor{white} GBDA \textcolor{tabcyan}{\ding{108}} &\cellcolor{white} $83.37$ &\cellcolor{white} $12.20_{\pm (6.94)}$ &\cellcolor{white} $0.85_{\pm (0.11)}$ & \cellcolor{white} $9.32_{\pm (1.78)}$ & \cellcolor{white} - & \cellcolor{white} - & \cellcolor{white} - & \cellcolor{white} - \\
        & BAE-R \textcolor{tabcyan}{\ding{108}} & $66.38$ & $10.10_{\pm (7.00)}$ & $0.83_{\pm (0.18)}$ & $1.24_{\pm (0.86)}$ & $63.16$ & $10.22_{\pm (6.33)}$ & $0.85_{\pm (0.16)}$ & $0.73_{\pm (0.62)}$ \\
        & \cellcolor{white} BERT-attack \textcolor{tabcyan}{\ding{108}} &\cellcolor{white} $69.57$ &\cellcolor{white} $12.19_{\pm (9.55)}$ & \cellcolor{white}$0.87_{\pm (0.09)}$ & \cellcolor{white}$239.80_{\pm (1763.30)}$ &\cellcolor{white} $64.21$ &\cellcolor{white} $11.26_{\pm (7.18)}$ & \cellcolor{white}$0.86_{\pm (0.10)}$ &\cellcolor{white} $18.12_{\pm (32.34)}$ \\
        & DeepWordBug \textcolor{tabred}{\ding{108}} & $81.39$ & $3.74_{\pm (2.95)}$ & $0.80_{\pm (0.17)}$ & $\mathbf{0.22}_{\pm (0.12)}$ & $84.27$ & $4.61_{\pm (3.47)}$ & $0.75_{\pm (0.20)}$ & $\mathbf{0.28}_{\pm (0.16)}$ \\
        & \cellcolor{white} TextBugger \textcolor{tabred}{\ding{108}} &\cellcolor{white} $68.49$ & \cellcolor{white}$5.97_{\pm (5.87)}$ &\cellcolor{white} $\mathbf{0.91}_{\pm (0.06)}$ & \cellcolor{white}$1.75_{\pm (0.91)}$ & \cellcolor{white}$61.10$ &\cellcolor{white} $6.85_{\pm (6.54)}$ & \cellcolor{white}$\mathbf{0.90}_{\pm (0.05)}$ & \cellcolor{white} $1.82_{\pm (0.97)}$ \\
        & TextFooler \textcolor{tabcyan}{\ding{108}} & $\underline{95.16}$ & $17.17_{\pm (12.51)}$ & $0.82_{\pm (0.15)}$ & $0.90_{\pm (0.57)}$ & $95.00$ & $17.76_{\pm (12.45)}$ & $0.82_{\pm (0.15)}$ & $1.16_{\pm (0.76)}$ \\
        & \cellcolor{white} TextGrad \textcolor{tabcyan}{\ding{108}} & \cellcolor{white}$94.04$ & \cellcolor{white}$21.61_{\pm (11.30)}$ & \cellcolor{white}$0.75_{\pm (0.13)}$ &\cellcolor{white} $19.94_{\pm (22.32)}$ &\cellcolor{white} $95.49$ & \cellcolor{white}$17.07_{\pm (9.57)}$ & \cellcolor{white}$0.81_{\pm (0.10)}$ &\cellcolor{white} $3.75_{\pm (2.83)}$ \\
        & CWBA \textcolor{tabred}{\ding{108}} & $72.92$ & $8.55_{\pm (3.78)}$ & $0.53_{\pm (0.26)}$ & $33.81_{\pm (33.86)}$ & $49.84$ & $8.88_{\pm (3.94)}$ & $0.65_{\pm (0.17)}$ & $56.35_{\pm (46.42)}$ \\
        \rowcolor{white}
        & \citep{pruthi2019misspellings} \textcolor{tabred}{\ding{108}} & $90.94$ & $2.22_{\pm (1.35)}$ & $0.85_{\pm (0.14)}$ & $4.86_{\pm (4.02)}$ & $92.93$ & $2.52_{\pm (1.57)}$ & $0.84_{\pm (0.14)}$ & $5.29_{\pm (4.68)}$ \\
        & \methodname{}\texttt{-Fast} (Ours) \textcolor{tabred}{\ding{108}} & $\mathbf{100.00}$ & $\underline{1.74}_{\pm (1.02)}$ & $0.89_{\pm (0.13)}$ & $\underline{0.34}_{\pm (0.31)}$ & $\underline{99.39}$ & $\underline{2.29}_{\pm (1.53)}$ & $0.84_{\pm (0.15)}$ & $\underline{0.47}_{\pm (0.49)}$ \\
        \rowcolor{white} & \methodname{} (Ours) \textcolor{tabred}{\ding{108}} & $\mathbf{100.00}$ & $\mathbf{1.47}_{\pm (0.74)}$ & $\underline{0.90}_{\pm (0.11)}$ & $1.27_{\pm (0.84)}$ & $\mathbf{99.51}$ & $\mathbf{1.76}_{\pm (1.12)}$ & $\underline{0.89}_{\pm (0.12)}$ & $1.52_{\pm (1.25)}$ \\
        \bottomrule
    \end{tabular}
    }
    \label{tab:textattack_benchmark}
\end{table*}

\subsection{Attack Classifier}
\label{subsubsec:attack_classifier}
In the case of using a classifier $\bm{f}: \mathcal{S}(\Gamma) \to \realnum^{o}$, where the predicted class is given by $\hat{y} = \argmax_{y \in [o]} f(S)_{y}$ with $o$ classes, %
we follow \citet{hou2023textgrad} and use the Carlini-Wagner Loss\footnote{In the original paper, \citet{carlini2017towards} clip the value of the loss to be $0$ at maximum. We do not clip in order to deal with cases where the loss is positive for different adversarial examples.} \citep{carlini2017towards}:
\vspace{-4pt}
\begin{equation}
\mathcal{L}(\bm{f}(S),y) = \max_{\hat{y}\neq y}f(S)_{\hat{y}} - f(S)_{y}\,.
\label{eq:loss_cw}
\vspace{-4pt}
\end{equation}
In this case, a sentence $S'$ is an adversarial example when $\mathcal{L}(\bm{f}(S'),y) \geq 0$.
To search the closest sentence in Levenshtein distance that produces a misclassification, we iteratively solve \cref{eq:bp_problem} with $k=1$ until the adversarial sentence $S'$ is misclassified. Our attack pseudo-code is presented in \cref{alg:ours}.

\begin{algorithm}[t]
	\caption{\methodname{} Adversarial Attack} 
	\label{alg:ours}
	\begin{spacing}{1.3}

	\begin{algorithmic}[1]

	\State \textbf{Inputs:} model $f$, sentence $S$, alphabet of characters $\Gamma$, max Levenshtein distance $k$, candidate positions $n$, loss function $\mathcal{L}$ and label $y$. %
        \State $S' = S$ \Comment{\textcolor{teal}{Initialize attack}}
	\For {$i=1,\dots, k$}
                \State $Z = \text{get\_top\_locations}(f,S',y,n)$ \Comment{\textcolor{teal}{\cref{alg:position_selection}}}
                \State $\mathcal{S}' = \{\psi(\phi(S') \overset{j}{\leftarrow} c), ~ \forall j \in Z, \forall c \in \Gamma \cup \{\xi\}\}$ \Comment{\textcolor{teal}{All sentences with modifications in $Z$}}
                \State $\bm{l} = \mathcal{L}(f(\mathcal{S}'),y)$ \Comment{\textcolor{teal}{Batch of $n\cdot(|\Gamma|+1)$ sentences}}
                \State $j^{*}  = \underset{j\in [|\mathcal{S'}|]}{\argmax}~l_{j}$
                \State $S' = \mathcal{S}'_{j^{*}}$ \Comment{\textcolor{teal}{Sentence with highest loss in the batch}}
                \If{$\underset{\hat{y}\in [o]}{\argmax}~f(S')\neq y$} \Return $S'$ \Comment{\textcolor{green}{Successful}}
                \EndIf
	\EndFor
        \State \Return $S'$ \Comment{\textcolor{red}{Unsuccessful}}
	
	\end{algorithmic} 
	
	\end{spacing}
\end{algorithm}

\subsection{Attack LLM}
\label{subsubsec:attack_llm}
Now we illustrate how to apply our method on attacking LLM-based classifiers. Given a data sample $(S,y) \in \mathcal{S}\times[C]$, the input to LLMs is formulated by concatenating the original sentences with some instructive prompts $S_{P_1}, S_{P_2}$ in the format of $ S_{P_1} \oplus S\oplus S_{P_2}$.  A schematic for illustration and additional details on the prompting strategy can be found in \cref{app:appendix_promptllm}.
Similar to \cref{eq:attack_problem}, we aim to solve:
\begin{equation*}
    \max_{S'\in \mathcal{S}_{k}(S,\Gamma)}
    \mathcal{L}\left(\bm{f} (
     S_{P_1} \oplus S' \oplus S_{P_2}
    )
    , y\right)\,,
    \label{eq:prob_llm}
\end{equation*}
where the model output ${f} (S_{P_1} \oplus S' \oplus S_{P_2}
    )_i:=\mathbb{P}(i| S_{P_1} \oplus S' \oplus S_{P_2})$ is the conditional probability of the next token $i$.
We can still use the Carlini-Wagner Loss defined in \cref{eq:loss_cw} by considering the next token probability for the classes.

\section{Experiments}
\label{sec:exp}

Our experiments are conducted in the publicly available\footnote{\url{https://huggingface.co/textattack}} TextAttack models \citep{morris2020textattack} and open-source large language models including \llamachat{}~\citep{touvron2023llama} and Vicuna 7B~\citep{chiang2023vicuna}. We evaluate our attack in the text (or text pair) classification datasets SST-2 \citep{socher2013sst}, MNLI-m \citep{williams2018mnli}, RTE \citep{dagan2006rte,wang2018glue}, QNLI \citep{rajpurkar2016qnli} and AG-News \citep{gulli2005agnews,zhang2015agnews}.

In the text pair classification tasks (MNLI-m, RTE, and QNLI), we perturb only the \emph{hypothesis} sentence. If the length of the test dataset is more than $1,000$, we restrict to the first $1,000$ samples. If a test dataset is not available for a benchmark, we evaluate in the validation dataset, this is a standard practice \citep{morris2020textattack}. For comparison with other attacks, we use the default hyperparameters of those methods. For \methodname{} we use $n=20$ positions (see \cref{alg:position_selection}) and $k=10$ except for AG-news where we use $k=20$ because of the much longer sentences present in the dataset. \methodname{}\texttt{-Fast} simply takes $n=1$ to speed-up the attack. For the alphabet $\Gamma$, in order to not introduce out-of-distribution characters, we take the characters present in each evaluation dataset.
All of our experiments were conducted in a machine with a single NVIDIA A100 SXM4 GPU. 
For better illustration between token-level and character-level attacks, we mark them with \textcolor{tabcyan}{\ding{108}} and \textcolor{tabred}{\ding{108}} respectively.
We note that alternative design decisions can enable the usage of projected gradient ascent (PGA) attacks. However, we observed a worse performance in comparison with our strategy, see \cref{app:pga}. %

\subsection{Selecting the Number of Positions}
\label{subsec:ablation_positions}
To select the appropriate number of candidate positions $n$ for \cref{alg:position_selection}, we evaluate the ASR and runtime of the attack with $n \in \{1,5,10,20,30,40,50,60,70\}$. We conduct the experiment with the fine-tuned BERT on SST-2 from TextAttack at $k=1$. For the SST-2 test sentences, the maximum number of positions across the dataset is $489$ and the average is $213.72$. We would like a value of $n$ much smaller than these values. As a comparison, we report the ASR computed by exploring all possible positions (ASR Upper Bound). Additionally, to test the effect of our heuristic, we evaluate the performance when randomly selecting $n$ positions (Random).

\begin{figure}
    \centering
    \resizebox{0.48\textwidth}{!}{
    \input{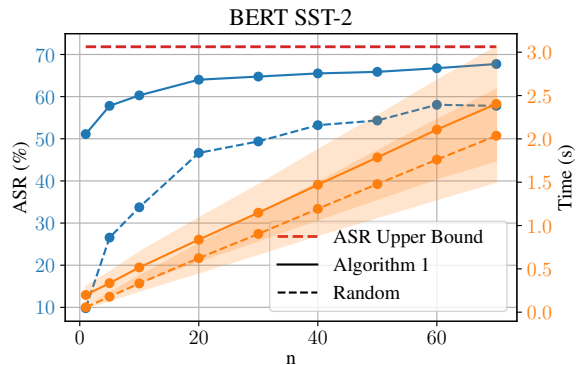}}
    \vspace{-30pt}
    \caption{\textbf{Selection of the number of candidate positions:} Attack Success Rate (ASR) at $k=1$ (\textcolor{tabblue}{\ding{108}} left axis) and runtime (\textcolor{taborange}{\ding{108}} right axis) for our candidate position selection strategy (\cref{alg:position_selection}, bold lines) and a random selection (Random, dotted lines). Our strategy improves the random baseline at a small cost ($\approx 0.25 s$). %
    }
    \vspace{-15pt}
    \label{fig:n_positions}
\end{figure}

In \cref{fig:n_positions}, we can observe that the ASR consistently grows when increasing the number of candidate positions. However, the increase is less noticeable for $n>20$, therefore, the increase in runtime does not pay off the increase in ASR. This leads us to choose $n=20$ for the rest of our experiments. When compared with the random position selection, our method greatly improves the ASR for all the studied $n$, while introducing a minor time increase of $0.25$ seconds on average. %

\subsection{Comparison Against State-of-the-art Attacks}
\label{subsec:sota}
Firstly, we compare against the following state-of-the-art attacks \textbf{(i) token level:} BAE-R \citep{garg2020bae}, TextFooler \citep{Jin2020textfooler}, BERT-attack \citep{li2020bertattack}, GBDA \citep{guo2021gradient} and TextGrad \citep{hou2023textgrad}, \textbf{(ii) character level:} DeepWordBug \cite{gao2018deepwordbug}, TextBugger \citep{li2019textbugger} and CWBA \cite{liu2022character}. For each attack method, we evaluate the attach success rate (ASR), the average Levenshtein distance measured at character level ($d_{\text{lev}}(S,S')$) and the cosine similarity ($\text{Sim}(S,S')$) measured as in \cite{guo2021gradient}, i.e., computing the cosine similarity of the USE encodings \citep{cer2018USE}. We evaluate the performance of the attacks in the finetuned BERT \citep{devlin2019bert} and RoBERTa \citep{liu2019roberta} from TextAttack. Additional experiments with ALBERT \citep{Lan2020albert} can be found in \cref{app:experiments}.

Secondly, we test the performance of the proposed method in \llamachat{}~\citep{touvron2023llama}. Additional results on Vicuna 7B \citep{chiang2023vicuna} are deferred to \cref{app:appendix_promptllm}. Note that in the case of LLMs, the inference process is extremely costly. As a result, we only use the fast version of \methodname{}, i.e., $n=1$. Moreover, we perform an additional position selection framework to further accelerate. Specifically, we first tokenize the input sentence and mask each token to determine the most important one based on the loss. Next, we perform \cref{alg:position_selection} for the position in these important tokens. An ablation study of such token selection procedure can be found in \cref{app:appendix_promptllm}.

In \cref{tab:textattack_benchmark,tab:llmclasifier}, we can observe \methodname{} consistently achieves the highest ASR with $>95\%$ in every benchmark. At the same time, our method obtains the lowest Levenshtein distance ($d_{\text{lev}}$). Regarding the similarity (Sim), our \methodname{} attains the highest or runner up similarity in $8/10$ cases, proving its ability to generate highly similar adversarial examples. With respect to time, \methodname{} is not as fast as the simple DeepWordBug, however, the runtime is comparable to previous state-of-the-art token-level TextGrad. If speed is preferred to adversarial example quality, we can set $n=1$ (\methodname{}\texttt{-Fast}), which attains a runtime closer to DeepWordBug at the cost of a higher $d_{\text{lev}}$\footnote{Except for RTE where the average $d_{\text{lev}}$ is smaller due to failure in hard examples, where higher $d_{\text{lev}}$ is needed.} and lower ASR. This phenomenon is aligned with the results of \cref{subsec:ablation_positions}, as the ASR at $k=1$ is lower when $n$ is lower. %
    
\begin{table}[ht]
\tabcolsep=0.06cm
    \centering
    \caption{\textbf{Attack evaluation in \llamachat{}.} \methodname{} \texttt{-Fast} outperforms baselines in terms of attack success rate, Levenshtein distance%
, and achieves comparable similarity and speed.
}
\vspace{2pt}
    \resizebox{0.48\textwidth}{!}{
    \begin{tabular}{l|g|vggg}
        \toprule
         \rowcolor{white}
         & Method &   ASR (\%)&     $d_{\text{lev}}(S,S')$ &      $\text{Sim}(S,S')$ &Time        \\
        \midrule
        \parbox[t]{2mm}{\multirow{7}{*}{\rotatebox[origin=c]{90}{SST-2}}} 
        & BAE-R \textcolor{tabcyan}{\ding{108}} &            $60.13$ &  $10.55$ &  $0.82$ &   $2.31$ \\
        \rowcolor{white}
            & BERT-attack \textcolor{tabcyan}{\ding{108}} &            $57.86$ &  $12.05$ &  $\underline{0.86}$ &   $1.61$ \\
           & DeepWordBug \textcolor{tabred}{\ding{108}} &            $50.82$ &   $\underline{5.24}$ &  $0.73$ &   $\mathbf{1.01}$ \\
            & \cellcolor{white} TextBugger \textcolor{tabred}{\ding{108}} &        \cellcolor{white}     $41.89$ & \cellcolor{white}   $8.99$ &  \cellcolor{white} $\mathbf{0.89}$ &  \cellcolor{white}  $1.63$ \\
            & TextFooler \textcolor{tabcyan}{\ding{108}} &            $\underline{85.79}$ &  $20.91$ &  $0.79$ &   $3.54$ \\
            \rowcolor{white}
            & \methodname{}\texttt{-Fast} \textcolor{tabred}{\ding{108}} &            $\mathbf{95.47}$ &   $\mathbf{2.55}$ &  $0.83$ &   $\underline{1.47}$ \\
                    \midrule
        \parbox[t]{3mm}{\multirow{7}{*}{\rotatebox[origin=c]{90}{QNLI}}} 
        & BAE-R \textcolor{tabcyan}{\ding{108}} &            $47.16$ &  $10.36$ &  $\mathbf{0.95}$ &   $2.46$\\
        \rowcolor{white}
        & BERT-attack \textcolor{tabcyan}{\ding{108}} &            $60.30$ &  $14.07$ &  $0.91$ &   $2.72$ \\
        & DeepWordBug \textcolor{tabred}{\ding{108}} &            $49.93$ &   $\underline{3.86}$ &  $0.88$ &   $\mathbf{1.24}$ \\
            &  \cellcolor{white} TextBugger \textcolor{tabred}{\ding{108}} &             \cellcolor{white} $58.92$ &  \cellcolor{white}  $10.59$ &  \cellcolor{white} $0.91$ &   \cellcolor{white} $\underline{2.44}$ \\
            & TextFooler \textcolor{tabcyan}{\ding{108}} &            $\underline{64.04}$ &  $18.03$ &  $0.91$ &   $4.05$ \\
         \rowcolor{white} 
            & \methodname{}\texttt{-Fast} \textcolor{tabred}{\ding{108}} &          
            $\mathbf{93.51}$&	$\mathbf{2.40}$	&$\underline{0.93}$	&$5.66$
            \\
        \midrule
        \parbox[t]{2mm}{\multirow{7}{*}{\rotatebox[origin=c]{90}{RTE}}} 
         & BAE-R \textcolor{tabcyan}{\ding{108}} &            $66.02$ &   $6.96$ &  $\mathbf{0.88}$ &   $1.39$ \\
         \rowcolor{white}
            & BERT-attack \textcolor{tabcyan}{\ding{108}} &            $\underline{90.78}$ &   $8.77$ &  $0.82$ &   $1.41$ \\
           & DeepWordBug \textcolor{tabred}{\ding{108}} &            $50.97$ &   $\underline{2.67}$ &  $0.76$ &   $\mathbf{0.61}$ \\
            & \cellcolor{white} TextBugger \textcolor{tabred}{\ding{108}} &            \cellcolor{white} $79.61$ & \cellcolor{white}   $7.76$ & \cellcolor{white}  $0.80$ &\cellcolor{white}    $\underline{1.19}$ \\
            & TextFooler \textcolor{tabcyan}{\ding{108}} &            ${86.41}$ &   $8.92$ &  $\underline{0.84}$ &   $1.73$ \\
            \rowcolor{white}
            & \methodname{}\texttt{-Fast} \textcolor{tabred}{\ding{108}} &            $\mathbf{97.10}$ &   $\mathbf{1.68}$ &  $0.82$ &   $2.06$ \\
        \bottomrule
    \end{tabular}}
\label{tab:llmclasifier}
\end{table}

\subsection{Adversarial Training}
\label{subsec:AT}
In this section, we analyze the performance of models trained with adversarial training defenses \citep{madry2018AT}. Following the insights of \citet{hou2023textgrad}, we use the TRADES objective \citep{Zhang2019TRADES}. We compare the use of a token-level attack, TextGrad, v.s. a character-level attack, \methodname{}, for solving the inner maximization problem. We use TextGrad with the recommended hyperparameters for training and \methodname{} with the standard hyperparameters and $k=1$. Every $100$ training steps, we measure the clean, TextFooler and \methodname{} ($k=1$) adversarial accuracies. We train on $5$ random initializations of BERT-base \citep{devlin2019bert} for $1$ epoch in SST-2.
\begin{table}
    \centering
    \setlength{\tabcolsep}{0.2\tabcolsep}
    \caption{\textbf{Adversarial Training defenses:} \methodname{} is an effective defense against character-level attacks, minimally affects clean accuracy and does not improve token-level robustness. On the contrary, TextGrad hinders character-level robustness and clean accuracy to improve token-level robustness.}
    \vspace{2pt}
    \resizebox{0.48\textwidth}{!}{
    \begin{tabular}{l|g|gg}
    \toprule
    Method & \cellcolor{white} Acc. (\%) $\uparrow$ & \cellcolor{white} ASR-Char (\%) $\downarrow$ & \cellcolor{white} ASR-Token (\%) $\downarrow$\\
    \midrule
    Standard & $\mathbf{92.43}$ & $\underline{64.02}$ & $\underline{95.16}$\\
    \methodname{} \textcolor{tabred}{\ding{108}} & \cellcolor{white} $\underline{87.20}_{\pm (1.34)}$ & \cellcolor{white}$\mathbf{20.34}_{\pm (1.17)}$ & \cellcolor{white}$95.17_{\pm (1.15)}$\\
    TextGrad \textcolor{tabcyan}{\ding{108}} & $80.94_{\pm (0.60)}$ & $67.34_{\pm (4.87)}$ & $\mathbf{71.36}_{\pm (3.63)}$\\
    \bottomrule
    \end{tabular}}
    \label{tab:at_results}
\end{table}

\begin{figure}[t]
    \centering
    \resizebox{0.75\columnwidth}{!}{
\def\mathdefault#1{#1}
\begingroup%
\makeatletter%
\begin{pgfpicture}%
\pgfpathrectangle{\pgfpointorigin}{\pgfqpoint{4.000000in}{1.000000in}}%
\pgfusepath{use as bounding box, clip}%
\begin{pgfscope}%
\pgfsetbuttcap%
\pgfsetmiterjoin%
\definecolor{currentfill}{rgb}{1.000000,1.000000,1.000000}%
\pgfsetfillcolor{currentfill}%
\pgfsetlinewidth{0.000000pt}%
\definecolor{currentstroke}{rgb}{1.000000,1.000000,1.000000}%
\pgfsetstrokecolor{currentstroke}%
\pgfsetdash{}{0pt}%
\pgfpathmoveto{\pgfqpoint{0.000000in}{0.000000in}}%
\pgfpathlineto{\pgfqpoint{4.000000in}{0.000000in}}%
\pgfpathlineto{\pgfqpoint{4.000000in}{1.000000in}}%
\pgfpathlineto{\pgfqpoint{0.000000in}{1.000000in}}%
\pgfpathlineto{\pgfqpoint{0.000000in}{0.000000in}}%
\pgfpathclose%
\pgfusepath{fill}%
\end{pgfscope}%
\begin{pgfscope}%
\pgfsetbuttcap%
\pgfsetmiterjoin%
\definecolor{currentfill}{rgb}{1.000000,1.000000,1.000000}%
\pgfsetfillcolor{currentfill}%
\pgfsetfillopacity{0.800000}%
\pgfsetlinewidth{1.003750pt}%
\definecolor{currentstroke}{rgb}{0.800000,0.800000,0.800000}%
\pgfsetstrokecolor{currentstroke}%
\pgfsetstrokeopacity{0.800000}%
\pgfsetdash{}{0pt}%
\pgfpathmoveto{\pgfqpoint{0.112318in}{0.375000in}}%
\pgfpathlineto{\pgfqpoint{3.887682in}{0.375000in}}%
\pgfpathquadraticcurveto{\pgfqpoint{3.915459in}{0.375000in}}{\pgfqpoint{3.915459in}{0.402778in}}%
\pgfpathlineto{\pgfqpoint{3.915459in}{0.597222in}}%
\pgfpathquadraticcurveto{\pgfqpoint{3.915459in}{0.625000in}}{\pgfqpoint{3.887682in}{0.625000in}}%
\pgfpathlineto{\pgfqpoint{0.112318in}{0.625000in}}%
\pgfpathquadraticcurveto{\pgfqpoint{0.084541in}{0.625000in}}{\pgfqpoint{0.084541in}{0.597222in}}%
\pgfpathlineto{\pgfqpoint{0.084541in}{0.402778in}}%
\pgfpathquadraticcurveto{\pgfqpoint{0.084541in}{0.375000in}}{\pgfqpoint{0.112318in}{0.375000in}}%
\pgfpathlineto{\pgfqpoint{0.112318in}{0.375000in}}%
\pgfpathclose%
\pgfusepath{stroke,fill}%
\end{pgfscope}%
\begin{pgfscope}%
\pgfsetbuttcap%
\pgfsetroundjoin%
\definecolor{currentfill}{rgb}{0.121569,0.466667,0.705882}%
\pgfsetfillcolor{currentfill}%
\pgfsetlinewidth{1.003750pt}%
\definecolor{currentstroke}{rgb}{0.121569,0.466667,0.705882}%
\pgfsetstrokecolor{currentstroke}%
\pgfsetdash{}{0pt}%
\pgfsys@defobject{currentmarker}{\pgfqpoint{-0.069444in}{-0.069444in}}{\pgfqpoint{0.069444in}{0.069444in}}{%
\pgfpathmoveto{\pgfqpoint{0.000000in}{-0.069444in}}%
\pgfpathcurveto{\pgfqpoint{0.018417in}{-0.069444in}}{\pgfqpoint{0.036082in}{-0.062127in}}{\pgfqpoint{0.049105in}{-0.049105in}}%
\pgfpathcurveto{\pgfqpoint{0.062127in}{-0.036082in}}{\pgfqpoint{0.069444in}{-0.018417in}}{\pgfqpoint{0.069444in}{0.000000in}}%
\pgfpathcurveto{\pgfqpoint{0.069444in}{0.018417in}}{\pgfqpoint{0.062127in}{0.036082in}}{\pgfqpoint{0.049105in}{0.049105in}}%
\pgfpathcurveto{\pgfqpoint{0.036082in}{0.062127in}}{\pgfqpoint{0.018417in}{0.069444in}}{\pgfqpoint{0.000000in}{0.069444in}}%
\pgfpathcurveto{\pgfqpoint{-0.018417in}{0.069444in}}{\pgfqpoint{-0.036082in}{0.062127in}}{\pgfqpoint{-0.049105in}{0.049105in}}%
\pgfpathcurveto{\pgfqpoint{-0.062127in}{0.036082in}}{\pgfqpoint{-0.069444in}{0.018417in}}{\pgfqpoint{-0.069444in}{0.000000in}}%
\pgfpathcurveto{\pgfqpoint{-0.069444in}{-0.018417in}}{\pgfqpoint{-0.062127in}{-0.036082in}}{\pgfqpoint{-0.049105in}{-0.049105in}}%
\pgfpathcurveto{\pgfqpoint{-0.036082in}{-0.062127in}}{\pgfqpoint{-0.018417in}{-0.069444in}}{\pgfqpoint{0.000000in}{-0.069444in}}%
\pgfpathlineto{\pgfqpoint{0.000000in}{-0.069444in}}%
\pgfpathclose%
\pgfusepath{stroke,fill}%
}%
\begin{pgfscope}%
\pgfsys@transformshift{0.278985in}{0.513889in}%
\pgfsys@useobject{currentmarker}{}%
\end{pgfscope}%
\end{pgfscope}%
\begin{pgfscope}%
\definecolor{textcolor}{rgb}{0.000000,0.000000,0.000000}%
\pgfsetstrokecolor{textcolor}%
\pgfsetfillcolor{textcolor}%
\pgftext[x=0.528985in,y=0.465278in,left,base]{\color{textcolor}{\rmfamily\fontsize{10.000000}{12.000000}\selectfont\catcode`\^=\active\def^{\ifmmode\sp\else\^{}\fi}\catcode`\%=\active\def
\end{pgfscope}%
\begin{pgfscope}%
\pgfsetbuttcap%
\pgfsetroundjoin%
\definecolor{currentfill}{rgb}{0.172549,0.627451,0.172549}%
\pgfsetfillcolor{currentfill}%
\pgfsetlinewidth{1.003750pt}%
\definecolor{currentstroke}{rgb}{0.172549,0.627451,0.172549}%
\pgfsetstrokecolor{currentstroke}%
\pgfsetdash{}{0pt}%
\pgfsys@defobject{currentmarker}{\pgfqpoint{-0.069444in}{-0.069444in}}{\pgfqpoint{0.069444in}{0.069444in}}{%
\pgfpathmoveto{\pgfqpoint{0.000000in}{-0.069444in}}%
\pgfpathcurveto{\pgfqpoint{0.018417in}{-0.069444in}}{\pgfqpoint{0.036082in}{-0.062127in}}{\pgfqpoint{0.049105in}{-0.049105in}}%
\pgfpathcurveto{\pgfqpoint{0.062127in}{-0.036082in}}{\pgfqpoint{0.069444in}{-0.018417in}}{\pgfqpoint{0.069444in}{0.000000in}}%
\pgfpathcurveto{\pgfqpoint{0.069444in}{0.018417in}}{\pgfqpoint{0.062127in}{0.036082in}}{\pgfqpoint{0.049105in}{0.049105in}}%
\pgfpathcurveto{\pgfqpoint{0.036082in}{0.062127in}}{\pgfqpoint{0.018417in}{0.069444in}}{\pgfqpoint{0.000000in}{0.069444in}}%
\pgfpathcurveto{\pgfqpoint{-0.018417in}{0.069444in}}{\pgfqpoint{-0.036082in}{0.062127in}}{\pgfqpoint{-0.049105in}{0.049105in}}%
\pgfpathcurveto{\pgfqpoint{-0.062127in}{0.036082in}}{\pgfqpoint{-0.069444in}{0.018417in}}{\pgfqpoint{-0.069444in}{0.000000in}}%
\pgfpathcurveto{\pgfqpoint{-0.069444in}{-0.018417in}}{\pgfqpoint{-0.062127in}{-0.036082in}}{\pgfqpoint{-0.049105in}{-0.049105in}}%
\pgfpathcurveto{\pgfqpoint{-0.036082in}{-0.062127in}}{\pgfqpoint{-0.018417in}{-0.069444in}}{\pgfqpoint{0.000000in}{-0.069444in}}%
\pgfpathlineto{\pgfqpoint{0.000000in}{-0.069444in}}%
\pgfpathclose%
\pgfusepath{stroke,fill}%
}%
\begin{pgfscope}%
\pgfsys@transformshift{1.292875in}{0.513889in}%
\pgfsys@useobject{currentmarker}{}%
\end{pgfscope}%
\end{pgfscope}%
\begin{pgfscope}%
\definecolor{textcolor}{rgb}{0.000000,0.000000,0.000000}%
\pgfsetstrokecolor{textcolor}%
\pgfsetfillcolor{textcolor}%
\pgftext[x=1.542875in,y=0.465278in,left,base]{\color{textcolor}{\rmfamily\fontsize{10.000000}{12.000000}\selectfont\catcode`\^=\active\def^{\ifmmode\sp\else\^{}\fi}\catcode`\%=\active\def
\end{pgfscope}%
\begin{pgfscope}%
\pgfsetbuttcap%
\pgfsetroundjoin%
\definecolor{currentfill}{rgb}{1.000000,0.498039,0.054902}%
\pgfsetfillcolor{currentfill}%
\pgfsetlinewidth{1.003750pt}%
\definecolor{currentstroke}{rgb}{1.000000,0.498039,0.054902}%
\pgfsetstrokecolor{currentstroke}%
\pgfsetdash{}{0pt}%
\pgfsys@defobject{currentmarker}{\pgfqpoint{-0.069444in}{-0.069444in}}{\pgfqpoint{0.069444in}{0.069444in}}{%
\pgfpathmoveto{\pgfqpoint{0.000000in}{-0.069444in}}%
\pgfpathcurveto{\pgfqpoint{0.018417in}{-0.069444in}}{\pgfqpoint{0.036082in}{-0.062127in}}{\pgfqpoint{0.049105in}{-0.049105in}}%
\pgfpathcurveto{\pgfqpoint{0.062127in}{-0.036082in}}{\pgfqpoint{0.069444in}{-0.018417in}}{\pgfqpoint{0.069444in}{0.000000in}}%
\pgfpathcurveto{\pgfqpoint{0.069444in}{0.018417in}}{\pgfqpoint{0.062127in}{0.036082in}}{\pgfqpoint{0.049105in}{0.049105in}}%
\pgfpathcurveto{\pgfqpoint{0.036082in}{0.062127in}}{\pgfqpoint{0.018417in}{0.069444in}}{\pgfqpoint{0.000000in}{0.069444in}}%
\pgfpathcurveto{\pgfqpoint{-0.018417in}{0.069444in}}{\pgfqpoint{-0.036082in}{0.062127in}}{\pgfqpoint{-0.049105in}{0.049105in}}%
\pgfpathcurveto{\pgfqpoint{-0.062127in}{0.036082in}}{\pgfqpoint{-0.069444in}{0.018417in}}{\pgfqpoint{-0.069444in}{0.000000in}}%
\pgfpathcurveto{\pgfqpoint{-0.069444in}{-0.018417in}}{\pgfqpoint{-0.062127in}{-0.036082in}}{\pgfqpoint{-0.049105in}{-0.049105in}}%
\pgfpathcurveto{\pgfqpoint{-0.036082in}{-0.062127in}}{\pgfqpoint{-0.018417in}{-0.069444in}}{\pgfqpoint{0.000000in}{-0.069444in}}%
\pgfpathlineto{\pgfqpoint{0.000000in}{-0.069444in}}%
\pgfpathclose%
\pgfusepath{stroke,fill}%
}%
\begin{pgfscope}%
\pgfsys@transformshift{2.955582in}{0.513889in}%
\pgfsys@useobject{currentmarker}{}%
\end{pgfscope}%
\end{pgfscope}%
\begin{pgfscope}%
\definecolor{textcolor}{rgb}{0.000000,0.000000,0.000000}%
\pgfsetstrokecolor{textcolor}%
\pgfsetfillcolor{textcolor}%
\pgftext[x=3.205582in,y=0.465278in,left,base]{\color{textcolor}{\rmfamily\fontsize{10.000000}{12.000000}\selectfont\catcode`\^=\active\def^{\ifmmode\sp\else\^{}\fi}\catcode`\%=\active\def
\end{pgfscope}%
\end{pgfpicture}%
\makeatother%
\endgroup%
}\\
    \vspace{-15pt}
    \resizebox{\columnwidth}{!}{\input{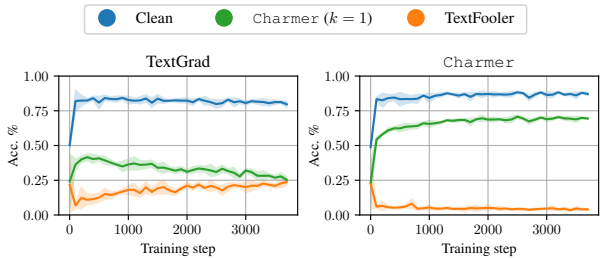}}
    \vspace{-25pt}
    \caption{\textbf{Adversarial Training evolution:} %
    When employing \methodname{} as a defense, clean and character-level accuracies grow consistently through training steps, while token-level (TextFooler) accuracy is unimproved. The TextGrad defense consistently improves the token-level accuracy at the cost of hindering clean and character-level accuracy, which grow in the first $\approx 400$ steps to then start decreasing.}
    \label{fig:at_progress}
\end{figure}

In \cref{tab:at_results} we can firstly observe that both the TextGrad and \methodname{} defenses improve the token-level and character-level robustness respectively when compared with the standard training baseline. This was expected as this is the objective each method is targetting. Interestingly, \methodname{} does not improve the token-level robustness and TextGrad hinders the character-level robustness. This observation is confirmed when looking at the training evolution in \cref{fig:at_progress}. It remains open to know if we should aim at character or token level robustness, nevertheless our results indicate character-level robustness is less conflicted with clean accuracy. %

\subsection{Bypassing Typo-correctors}
\label{subsec:typo}
We analyze the performance of our attack when attaking models defended by a typo-corrector \citep{pruthi2019misspellings}, or a robust encoding module \citep{jones2020robencodings}. We notice the success of these defenses can be attributed by the properties of the considered attacks. In \citet{pruthi2019misspellings,jones2020robencodings}, the studied attacks are constrained to\footnote{\citet{pruthi2019misspellings}, further constrain the attack by only considering replacements of nearby characters in the English keyboard.}:
\vspace{-8pt}
\begin{itemize}
\setlength\itemsep{0.0\itemsep}
    \item \texttt{Repeat}: Not perturb the same word twice.
    \item \texttt{First}: Not perturb the first character of a word.
    \item \texttt{Last}: Not perturb the last character of a word.
    \item \texttt{Length}: Not perturb words with less than $4$ chars.
    \item \texttt{LowEng}: Only perturb lowercase English letters.
\end{itemize}
\vspace{-8pt}
A word is anything between blank spaces. We denote these as the \emph{Pruthi-Jones Constraints} (\texttt{PJC}). While these constraints aim at preserving the meaning of \emph{every individual word} \citep{rawlinson1976words,davis2003words}, in sentence classification, we might sacrifice the meaning of a word during the attack, if the global meaning of the sentence is preserved. We analyze the performance of our attack with and without the \texttt{PJC} constraints. We train the strongest typo-corrector \citep{pruthi2019misspellings} and use it in front of the BERT-base model from TextAttack. For the robust encoding defense we train a BERT-base model over the agglomerative clusters \citep{jones2020robencodings}.
\begin{table}
    \centering
    \setlength{\tabcolsep}{0.2\tabcolsep}
    \caption{\textbf{Robust word recognition defenses:} \methodname{} is able to break the studied defenses with $100\%$ ASR. Robust word recognition defenses are effective only when considering \texttt{PCJ} constraints.}
    \vspace{2pt}
    \resizebox{0.48\textwidth}{!}{
    \begin{tabular}{lc|gggg}
    \toprule
    \rowcolor{white} Defense & Acc. (\%) & \texttt{PJC}? & ASR (\%) & $d_{\text{lev}}(S,S')$ &  $\text{Sim}(S,S')$ \\
    \midrule
    \multirow{2}{*}{None} & \multirow{2}{*}{$92.43$} & \textcolor{tabred}{\xmark} & $100.00$ & $1.47_{\pm (0.74)}$ & $0.90_{\pm (0.11)}$ \\
    & & \cellcolor{white} \textcolor{tabgreen}{\cmark} &\cellcolor{white} $96.65$ &\cellcolor{white} $1.86_{\pm (1.14)}$ &\cellcolor{white} $0.87_{\pm (0.14)}$ \\
    \midrule
    \multirow{2}{*}{\citep{pruthi2019misspellings}} & \multirow{2}{*}{$88.53$} & \textcolor{tabred}{\xmark} & $100.00$ & $1.28_{\pm (0.51)}$ & $0.90_{\pm (0.11)}$ \\
    & & \cellcolor{white} \textcolor{tabgreen}{\cmark} &\cellcolor{white} $70.34$ &\cellcolor{white} $2.08_{\pm (1.49)}$ &\cellcolor{white} $0.85_{\pm (0.14)}$ \\
    \midrule
    \multirow{2}{*}{\citep{jones2020robencodings}} &  \multirow{2}{*}{$83.94$} & \textcolor{tabred}{\xmark} & $100.00$ & $1.43_{\pm (0.71)}$ & $0.88_{\pm (0.11)}$ \\
    & & \cellcolor{white} \textcolor{tabgreen}{\cmark} &\cellcolor{white} $0.96$ &\cellcolor{white} $1.14_{\pm (0.38)}$ &\cellcolor{white} $0.92_{\pm (0.06)}$ \\
    \bottomrule
    \end{tabular}}
    \label{tab:typo_correctors}
    \vspace{-12pt}
\end{table}

In \cref{tab:typo_correctors}, we can observe that \methodname{} attains $100\%$ ASR when not considering the \texttt{PJC} constraints. It is only when considering \texttt{PJC} that robust word recognition defenses are effective. In \cref{tab:pjc} we analyze the effect of relaxing each of the \texttt{PJC} constraints while keeping the rest. We observe that by relaxing any of the \texttt{LowEng}, \texttt{End} or \texttt{Start} constraints, performance grows considerably for both defenses, e.g., from $0.96\%$ to $98.09\%$ ASR when relaxing \texttt{LowEng} in the robust encoding case. This result indicates that robust word recognition defenses provide a false sempsation of robustness. Together with the observations in \cref{subsec:AT}, we believe adversarial training based methods suppose a more promising avenue towards achieving character-level robustness.

\begin{table}
    \centering
    \setlength{\tabcolsep}{0.5\tabcolsep}
    \caption{\textbf{Effect of each \texttt{PJC} constraint:} \methodname{} ASR when individually removing each constraint while keeping the rest. The ASR drastically increases when removing the \texttt{LowEng}, \texttt{End} or \texttt{Start} constraints, proving the fragility of existing robust word recognition defenses.}
    \vspace{2pt}
    \renewcommand{\arraystretch}{1}
    \begin{tabular}{ccG|g}
    \toprule
    \rowcolor{white} Defense & Acc. (\%) & Attack constraint  & ASR (\%)\\
    \midrule
    \parbox[t]{3mm}{\multirow{6}{*}{\rotatebox[origin=c]{90}{\footnotesize{\citep{pruthi2019misspellings}}}}} & \multirow{6}{*}{$88.53$} &\texttt{PJC} & $70.34$ \\
    &&\cellcolor{white}~~\texttt{-LowEng} &\cellcolor{white} $99.22$ \\
    &&~~\texttt{-Length} & $74.61$ \\
    &&\cellcolor{white}~~\texttt{-End} &\cellcolor{white} $93.91$ \\
    &&~~\texttt{-Start} & $98.58$ \\
    &&\cellcolor{white}~~\texttt{-Repeat} & \cellcolor{white}$74.09$ \\
    \midrule
    \parbox[t]{3mm}{\multirow{6}{*}{\rotatebox[origin=c]{90}{\footnotesize{\citep{jones2020robencodings}}}}} & \multirow{6}{*}{$83.94$} &\texttt{PJC} & $0.96$\\
    &&\cellcolor{white}~~\texttt{-LowEng} &\cellcolor{white} $98.09$ \\
    &&~~\texttt{-Length} & $0.96$ \\
    &&\cellcolor{white}~~\texttt{-End} &\cellcolor{white} $71.72$ \\
    &&~~\texttt{-Start} & $88.93$ \\
    &&\cellcolor{white}~~\texttt{-Repeat} &\cellcolor{white} $5.46$ \\
    \bottomrule
    \end{tabular}
    \vspace{-8pt}
    \label{tab:pjc}
\end{table}

\section{Conclusion}
\label{sec:conclusion}
We have proposed an efficient character-level attack based on a novel strategy to select the best positions to perturb at each iteration. Our attack (\methodname{}) is able to obtain close to $100\%$ ASR both in BERT-like models and LLMs like Llama-2. \methodname{} defeats both token-based adversarial training defenses \citep{hou2023textgrad} and robust word recognition defenses \citep{pruthi2019misspellings,jones2020robencodings}. When integrated within adversarial training, our attack is able to improve the robustness against character-level attacks. We believe defending agains character-level attacks is an interesting open problem, with adversarial training posing as a promising avenue for defenses.

\clearpage
\section*{Acknowledgements}
Authors acknowledge the constructive feedback of reviewers and the work of ICML'24 program and area chairs. We thank Zulip\footnote{\url{https://zulip.com}} for their project organization tool. ARO - Research was sponsored by the Army Research Office and was accomplished under Grant Number W911NF-24-1-0048.
Hasler AI - This work was supported by Hasler Foundation Program: Hasler Responsible AI (project number 21043).
SNF project – Deep Optimisation - This work was supported by the Swiss National Science Foundation (SNSF) under grant number 200021\_205011.
Fanghui Liu is supported by the Alan Turing Institute under the UK-Italy Trustworthy AI Visiting Researcher Programme.

\section*{Impact Statement}
In this work, we revisit character-level adversarial attacks and improve upon the prior art performance. %
We believe that showing that character-level attacks cannot easily be defended, is important to warn about the need of defenses.
Otherwise, malicious individuals or organizations could take advantage of this unawareness. However, we note that our algorithm could empower individuals to achieve malicious purposes. We will release our code to allow defenders to assess their performance against our attack.

\bibliography{bibliography.bib}
\bibliographystyle{icml2024}

\clearpage
\appendix
\renewcommand{\thefigure}{S\arabic{figure}} %
\renewcommand{\thetable}{S\arabic{table}} %
\renewcommand{\thetheorem}{S\arabic{theorem}} %

\begin{onecolumn}
\section*{Contents of the Appendix}
In \cref{app:background} we provide additional background in NLP, character-level adversarial attacks and the employed datasets. In \cref{app:experiments} we provide additional experimental validation of \methodname{}. In \cref{app:method}, we provide the proof of \cref{cor:Sk}. Lastly, \cref{app:pga} contains possible gradient-based methods for solving \cref{eq:attack_problem}.

\section{Additional Background} 
\label{app:background}
We introduce additional information about other character-level attacks in \cref{subsec:comparison_char_attacks}, the employed datasets in \cref{subsec:data} and about Neural Networks (NNs) for NLP in \cref{subsec:transformers}.

\subsection{Comparison with Other Character-level Attacks}
\label{subsec:comparison_char_attacks}

In this section we analyze in detail the relationship between existing character-level attacks and \methodname{}. As shown in \cref{fig:n_positions}, the key improvement in performance, without sacrificing much ASR, comes from our position subset selection strategy (\cref{alg:position_selection}). In addition to this technique, we analyze the main differences between methods with the following characteristics:
\begin{itemize}
    \item \texttt{Two-phase}: Does the method adopt the two-phase paradigm? I.e., evaluate the importance of chars/words to then greedily select the best change.
    \item \texttt{Any architecture}: Can the method handle any model architecture? E.g., can it handle char-level and token-level models?
    \item \texttt{Insertions and deletions}: Can the attack perform insertions and deletions of characters?
    \item \texttt{Importance re-evaluation}: Does the method re-evaluate the importance of a word/character after a perturbation is introduced?
    \item \texttt{PJC}: Does the method adopt the \texttt{PJC} constraints?
\end{itemize}

\begin{table}[]
    \centering
    \caption{\textbf{Qualitative comparison with other character-level methods:}}
    \resizebox{\textwidth}{!}{\begin{tabular}{c|ggggg}
         \rowcolor{white}Method & \texttt{Two-phase} & \texttt{Any architecture} & \texttt{Insertions and deletions} & \texttt{Importance re-evaluation} & \texttt{PJC} \\
         \toprule
         Hotflip & \textcolor{tabgreen}{\cmark} & \textcolor{tabred}{\xmark} & \textcolor{tabgreen}{\cmark} & \textcolor{tabgreen}{\cmark} & \textcolor{tabred}{\xmark}\\ 
         \rowcolor{white}DeepWordBug & \textcolor{tabgreen}{\cmark} & \textcolor{tabgreen}{\cmark} & \textcolor{tabgreen}{\cmark} & \textcolor{tabred}{\xmark} & \textcolor{tabred}{\xmark}\\
         TextBugger & \textcolor{tabgreen}{\cmark} & \textcolor{tabgreen}{\cmark} & \textcolor{tabgreen}{\cmark} & \textcolor{tabred}{\xmark} & \textcolor{tabred}{\xmark}\\
         \rowcolor{white}\citep{Yang2020greedy} & \textcolor{tabgreen}{\cmark} & \textcolor{tabgreen}{\cmark} & \textcolor{tabred}{\xmark} & \textcolor{tabred}{\xmark} & \textcolor{tabred}{\xmark}\\
         \citep{pruthi2019misspellings} & \textcolor{tabred}{\xmark} & \textcolor{tabgreen}{\cmark} & \textcolor{tabgreen}{\cmark} & \textbf{-} & \textcolor{tabgreen}{\cmark}\\
         \rowcolor{white}CWBA & \textcolor{tabred}{\xmark} & \textcolor{tabred}{\xmark} & \textcolor{tabgreen}{\cmark} & \textbf{-} & \textcolor{tabred}{\xmark}\\
         \methodname{} & \textcolor{tabgreen}{\cmark} & \textcolor{tabgreen}{\cmark} & \textcolor{tabgreen}{\cmark} & \textcolor{tabgreen}{\cmark} & \textcolor{tabred}{\xmark}\\
    \end{tabular}}
    
    \label{tab:char-level_methods}
\end{table}

In \cref{tab:char-level_methods}, we cover the characteristics of the character-level attacks studied in this work. Note that this table does not capture aspects like the specific strategy employed for every method to select the subset of changes and the final change in the two-stage paradigm.

\subsection{Datasets}
\label{subsec:data}
In \cref{tab:datasets_description} we provide the size, classes, alphabet and examples for all the studied datasets. All of our datasets are publicly available in \url{https://huggingface.co/datasets}.

\subsection{NN Architectures for NLP}
\label{subsec:transformers}

Unlike Computer Vision applications, where images can be directly fed into the NN, some pre-processing is needed in order to feed text into our models. %
A common practice is grouping characters into \emph{tokens} \citep{websterkit1992tokenization,palmer2000tokenisation,sennrich2015BPE,kudo2018sentencepiece,song2021fastwordpiece} and assigning a vector representation (\emph{embedding}) to each token in the text \citep{bengio2000embeddings,mikolov2013word2vec,pennington2014glove,bojanowski2017fasttext}. After a sequence of vector representations is obtained, an appropriate NN architecture can be used, e.g., RNNs or Transformers in an encoder and/or decoder fashion \cite{sutskever2014sequence,peters2018elmo,devlin2019bert,Brown2020gpt3}. 
Overall, the architecture will be:
\[
    f(S) = \hat{f}\left(\textbf{G}\left(S\right)\bm{T}\right)\,,
\]
where $\bm{E} = \textbf{G}\left(S\right)\bm{T}$ is the embedding representation of the sequence, $\text{G}: \mathcal{S}(\Gamma) \to \mathcal{S}(V_{\text{tok}})$ is the tokenizer with $V_{\text{tok}}$ as the token vocabulary and $\bm{T} \in \realnum^{|V_{\text{tok}}|\times d}$ is the matrix containing the embeddings of each token row-wise.

\begin{table}[h]
    \setlength{\tabcolsep}{0.1\tabcolsep}
    \centering
    \caption{\textbf{Attack transferability:} Adversarial examples are generated in the \emph{Source Model} to be evaluated in the \emph{Target Model}. For both \methodname{} and TextFooler, the ASR is consideraby lower when the target model is different from the source model. We observe no clear difference in transfer attack performance between TextFooler and \methodname{}. MNLI-m and AG-News are the easiest and hardest datasets for generating transfer attacks respectively.}
    \vspace{2pt}
    \begin{tabular}{c|ccccc}
        \toprule
         Attack & AG-News &  MNLI-m & QNLI & RTE & SST-2\\
         \midrule
         \raisebox{35pt}{\rotatebox[origin=c]{90}{TextFooler}} & \resizebox{0.18\textwidth}{!}{\input{figures/transfer/transfer_textfooler_agnews.pgf}} & \resizebox{0.18\textwidth}{!}{\input{figures/transfer/transfer_textfooler_mnli.pgf}} & \resizebox{0.18\textwidth}{!}{\input{figures/transfer/transfer_textfooler_qnli.pgf}} & \resizebox{0.18\textwidth}{!}{\input{figures/transfer/transfer_textfooler_rte.pgf}} & \resizebox{0.18\textwidth}{!}{\input{figures/transfer/transfer_textfooler_sst.pgf}} \\
         \midrule
         \raisebox{35pt}{\rotatebox[origin=c]{90}{\methodname{}}} & \resizebox{0.18\textwidth}{!}{\input{figures/transfer/transfer_chargrad_agnews.pgf}} & \resizebox{0.18\textwidth}{!}{\input{figures/transfer/transfer_chargrad_mnli.pgf}} & \resizebox{0.18\textwidth}{!}{\input{figures/transfer/transfer_chargrad_qnli.pgf}} & \resizebox{0.18\textwidth}{!}{\input{figures/transfer/transfer_chargrad_rte.pgf}} & \resizebox{0.18\textwidth}{!}{\input{figures/transfer/transfer_chargrad_sst.pgf}} \\
         \midrule
         & \multicolumn{5}{c}{\raisebox{3pt}{$0$~~}\resizebox{0.7\textwidth}{!}{\includegraphics{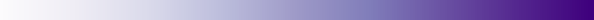}}\raisebox{3pt}{~~$100$}}\\
         \bottomrule
    \end{tabular}
    \label{tab:transfer_attacks}
\end{table}

\begin{table}
    \centering
    \caption{\textbf{Description of the employed datasets:} When a character is not printable in \LaTeX, we default to its Unicode encoding.}
    \begin{tabular}{ p{1.5cm}| p{14cm}}
         \multicolumn{2}{c}{\textbf{AG-News}}\\
         \toprule
         Test Size & 1,000\\
         Classes & 4 (World, Sports, Business, Sci/Tech)\\
         Alphabet $|\Gamma| = 82$ & \footnotesize{$\Gamma = $\{' ', '!', '"', '\#', '\$', '\&', ''', '(', ')', '*', ',', '-', '.', '/', '0', '1', '2', '3', '4', '5', '6', '7', '8', '9', ':', ';', '=', '?', 'A', 'B', 'C', 'D', 'E', 'F', 'G', 'H', 'I', 'J', 'K', 'L', 'M', 'N', 'O', 'P', 'Q', 'R', 'S', 'T', 'U', 'V', 'W', 'X', 'Y', 'Z', '\textbackslash', '\_', 'a', 'b', 'c', 'd', 'e', 'f', 'g', 'h', 'i', 'j', 'k', 'l', 'm', 'n', 'o', 'p', 'q', 'r', 's', 't', 'u', 'v', 'w', 'x', 'y', 'z'\}}\\
         Example & 
         $S = $'Fears for T N pension after talks Unions representing workers at Turner   Newall say they are 'disappointed' after talks with stricken parent firm Federal Mogul.', 
         $y = 3$ (Business)\\
         \bottomrule
         \multicolumn{2}{c}{\textbf{MNLI-m}}\\
         \toprule
         Test Size & 1,000\\
         Classes & 3 (Entailment, Neutral, Contradiction)\\
         Alphabet $|\Gamma| = 81$ & \footnotesize{$\Gamma = $\{' ', '!', '"', '\$', '\%', '\&', ''', '(', ')', ',', '-', '.', '/', '0', '1', '2', '3', '4', '5', '6', '7', '8', '9', ':', ';', '?', 'A', 'B', 'C', 'D', 'E', 'F', 'G', 'H', 'I', 'J', 'K', 'L', 'M', 'N', 'O', 'P', 'Q', 'R', 'S', 'T', 'U', 'V', 'W', 'X', 'Y', 'Z', 'a', 'b', 'c', 'd', 'e', 'f', 'g', 'h', 'i', 'j', 'k', 'l', 'm', 'n', 'o', 'p', 'q', 'r', 's', 't', 'u', 'v', 'w', 'x', 'y', 'z', '£', 'é', 'ô'\}}\\
         Example & $S_{\text{premise}} = $'The new rights are nice enough', 
         $S_{\text{hypothesis}} = $'Everyone really likes the newest benefits', $y = 2$ (Neutral)\\
         \bottomrule
         \multicolumn{2}{c}{\textbf{QNLI}}\\
         \toprule
         Test Size & 1,000\\
         Classes & 2 (Entailment, Not entailment)\\
         Alphabet $|\Gamma| = 233$ & \footnotesize{$\Gamma = $\{['21513', '21068', '25104', '8722', '8211', '8212', '8216', '8217', '8220', '8221', '24605', '27166', ' ', '!', '"', '\#', '\$', '\%', '\&', ''', '(', ')', '8230', '+', ',', '-', '.', '/', '0', '1', '2', '3', '4', '5', '6', '7', '8', '9', ':', ';', '8243', '=', '>', '?', '<', 'A', 'B', 'C', 'D', 'E', 'F', 'G', 'H', 'I', 'J', 'K', 'L', 'M', 'N', 'O', 'P', 'Q', 'R', 'S', 'T', 'U', 'V', 'W', 'X', 'Y', 'Z', '[', '8260', ']', '601', '\_', '`', 'a', 'b', 'c', 'd', 'e', 'f', 'g', 'h', 'i', 'j', 'k', 'l', 'm', 'n', 'o', 'p', 'q', 'r', 's', 't', 'u', 'v', 'w', 'x', 'y', 'z', '{', '|', '}', '~', '20094', '642', '8838', '21129', '650', '38498', '7841', '£', '7845', '7847', '8364', '20140', '8366', '°', '7857', '±', '·', '½', '38081', 'Å', 'Ç', '712', 'É', '7879', 'Î', '720', '40657', '7889', 'Ö', '×', 'Ü', 'ß', 'à', 'á', 'ä', '1063', 'æ', 'ç', 'è', 'é', 'ê', 'í', 'ï', 'ð', 'ñ', 'ó', '38515', 'õ', 'ö', 'ø', 'ù', '8801', 'û', 'ü', '65279', '7940', '263', '268', '269', '272', '1072', '275', '27735', '281', '283', '30494', '30495', '8478', '1075', '287', '22823', '26408', '299', '26413', '815', '305', '1079', '1080', '321', '322', '20803', '324', '333', '1085', '27491', '1089', '34157', '7547', '1093', '379', '626', '928', '592', '37941', '39340', '941', '942', '943', '594', '945', '946', '947', '948', '949', '432', '951', '952', '953', '954', '955', '956', '596', '950', '959', '957', '961', '962', '964', '966', '23494', '8134', '969', '973', '603', '8172', '8242']\}}\\
         Example & 
         $S_{\text{premise}} = $'What came into force after the new constitution was herald?', $S_{\text{hypothesis}} = $'As of that day, the new constitution heralding the Second Republic came into force.', $y = 1$ (Entailment)\\
         \bottomrule
         \multicolumn{2}{c}{\textbf{RTE}}\\
         \toprule
         Test Size & 277\\
         Classes & 2 (Entailment, Not entailment)\\
         Alphabet $|\Gamma| = 72$ & \footnotesize{$\Gamma = $\{' ', '"', '\$', '\%', '\&', ''', '(', ')', ',', '-', '.', '0', '1', '2', '3', '4', '5', '6', '7', '8', '9', 'A', 'B', 'C', 'D', 'E', 'F', 'G', 'H', 'I', 'J', 'K', 'L', 'M', 'N', 'O', 'P', 'Q', 'R', 'S', 'T', 'U', 'V', 'W', 'Y', 'Z', 'a', 'b', 'c', 'd', 'e', 'f', 'g', 'h', 'i', 'j', 'k', 'l', 'm', 'n', 'o', 'p', 'q', 'r', 's', 't', 'u', 'v', 'w', 'x', 'y', 'z'\}}\\
         Example & $S_{\text{premise}} = $'Dana Reeve, the widow of the actor Christopher Reeve, has died of lung cancer at age 44, according to the Christopher Reeve Foundation.', $S_{\text{hypothesis}} = $'Christopher Reeve had an accident.', $y = 2$ (Not entailment)\\
         \bottomrule
         \multicolumn{2}{c}{\textbf{SST-2}}\\
         \toprule
         Test Size & 872\\
         Classes & 2 (Negative, Positive)\\
         Alphabet $|\Gamma| = 55$ & \footnotesize{$\Gamma = $\{'æ', 'à', 'é', ' ', '!', '\$', '\%', ''', '(', ')', ',', '-', '.', '/', '0', '1', '2', '3', '4', '5', '6', '7', '8', '9', ':', ';', '?', '`', 'a', 'b', 'c', 'd', 'e', 'f', 'g', 'h', 'i', 'j', 'k', 'l', 'm', 'n', 'o', 'p', 'q', 'r', 's', 't', 'u', 'v', 'w', 'x', 'y', 'z', 'ö'\}}\\
         Example & $S = $'it 's a charming and often affecting journey .', $y = 2$ (Positive) \\
         \bottomrule
    \end{tabular}
    \label{tab:datasets_description}
\end{table}

\clearpage
\section{Additional Experimental Validation and Details}
\label{app:experiments}

\subsection{Perplexity Comparison}

To complete the analysis, we compare the perplexity of the attacked sentences for all the studied methods. Provided some methods directly optimize the perplexity of the attacked sentence \citep{guo2021gradient,hou2023textgrad}, we report the GPT-2 perplexity (PPL) \citep{radford2019gpt2}. Nevertheless, we do not consider PPL a good metric for assessing imperceptibility of adversarial examples. Consider the following sentences and their corresponding perplexity:
\begin{itemize}
    \item[\textbf{a)}] This film is good, 136.60
    \item[\textbf{b)}]This film is bad, 237.38
    \item[\textbf{c)}]This film is goodd, 1044.97
\end{itemize}

\textbf{a)} and \textbf{b)} are grammatically correct and, as expected, have a low PPL. However, they have a completely different meaning. Alternatively, \textbf{c)} is a typo of \textbf{a)} and shares its meaning, but has a much higher PPL. For this reason, we warn against the use of PPL for assessing imperceptibility.

\begin{table}[!ht]
    \centering
    \caption{\textbf{GPT-2 perplexity (PPL) and attack success rate (ASR) in the BERT-SST-2 TextAttack benchmark:}}
    \resizebox{\textwidth}{!}{
    \begin{tabular}{l|vg|vg|vg|vg|vg}
    \toprule
        \rowcolor{white} & \multicolumn{2}{c|}{SST-2} & \multicolumn{2}{c|}{QNLI} &\multicolumn{2}{c|}{MNLI-m} & \multicolumn{2}{c|}{RTE} & \multicolumn{2}{c}{AG-News} \\
        \rowcolor{white} Attack & ASR (\%) & PPL & ASR (\%) & PPL & ASR (\%) & PPL & ASR (\%) & PPL & ASR (\%) & PPL \\
        \midrule
        GBDA \textcolor{tabcyan}{\ding{108}} & 83.37 & 296.32 & 48.12 & 124.17 & 97.50 & 715.44 & 76.62 & 747.21 & 46.71 & 192.85 \\
        \rowcolor{white} BAE-R \textcolor{tabcyan}{\ding{108}} & 66.38 & 351.12 & 40.04 & 60.48 & 70.00 & 359.58 & 64.68 & 258.66 & 17.09 & 78.30 \\
        BertAttack \textcolor{tabcyan}{\ding{108}} & 69.57 & 346.92 & 70.21 & 102.12 & 92.41 & 289.52 & 68.00 & 442.74 & 29.90 & 118.09 \\
        \rowcolor{white} TextFooler \textcolor{tabcyan}{\ding{108}} & 95.16 & 539.19 & 80.64 & 210.61 & 92.26 & 576.84 & 79.60 & 971.69 & 78.98 & 334.05 \\
        TextGrad \textcolor{tabcyan}{\ding{108}} & 94.04 & 334.76 & 77.35 & 182.21 & 93.69 & 360.48 & 81.77 & 532.13 & 85.85 & 297.98 \\
        \rowcolor{white} DeepWordBug \textcolor{tabred}{\ding{108}} & 81.39 & 699.91 & 71.57 & 257.02 & 84.88 & 988.90 & 65.67 & 555.76 & 60.51 & 482.79 \\
        TextBugger \textcolor{tabred}{\ding{108}} & 68.49 & 396.88 & 75.77 & 200.58 & 85.36 & 665.93 & 74.13 & 439.07 & 50.85 & 224.74 \\
        \rowcolor{white} CWBA \textcolor{tabred}{\ding{108}} & 72.92 & 1206.91 & - & - & - & - & - & - & 86.72 & 758.29 \\
        \citep{pruthi2019misspellings} \textcolor{tabred}{\ding{108}} & 90.94 & 610.01 & 17.70 & 120.36 & 57.62 & 741.15 & 62.19 & 487.09 & 90.02 & 277.59 \\
        \rowcolor{white} \methodname{} \textcolor{tabred}{\ding{108}} & \textbf{100} & 569.17 & \textbf{97.68} & 153.73 & \textbf{100} & 998.59 & \textbf{97.01} & 987.32 & \textbf{98.51} & 161.07 \\
        \methodname{}\texttt{-Fast} \textcolor{tabred}{\ding{108}} & \textbf{100} & 644.66 & 94.69 & 191.42 & \textbf{100} & 1157.81 & 89.55 & 971.88 & 95.86 & 220.13 \\
    \bottomrule
    \end{tabular}}
    \label{tab:ppl}
\end{table}

In \cref{tab:ppl} we observe that \methodname{} and \methodname{}\texttt{-Fast} obtain similar PPLs to other character-level attacks. We notice token-level attacks present a lower PPL than character-level attacks, which was expected from the previous example.

\subsection{Qualitative Analysis of \methodname{}}
In this section we analyze the characteristics of the perturbations introduced by \methodname{}. 

\textbf{Common perturbations:} We display the most common operations for each dataset when attacking the corresponding TextAttack BERT model. In \cref{fig:common_operations} we can observe that across datasets, the most common operations correspond to insertions of punctuation marks such as paranthesis, dots, commas, question marks, percentages or dollar symbols.

\textbf{Location distribution:} We analyze the distribution of the location of perturbations across the sentences. From \cref{fig:location_distribution} on the one hand, we can conclude that there is no clear region where attacks are more common for the MNLI-m, RTE and SST-2 datasets. On the other hand, for AG-News and QNLI, perturbations appear to be more common closer to the beginning of the sentence. This inclination towards perturbations in AG-News, could be explained by the fact that most sentences in AG-News start with the news header, therefore, the model might be biased towards classifying based on the header.

\textbf{Attack examples:} For completeness, we provide in \cref{tab:examples_agnews,tab:examples_mnli,tab:examples_qnli,tab:examples_rte,tab:examples_sst} the BERT adversarial examples provided by every attack in \cref{tab:textattack_benchmark} in the first $3$ correctly classified sentences for each dataset. Additionally, we include examples of the successful BERT-SST-2 attacks from \cref{tab:typo_correctors} in \cref{tab:typo_examples}. We notice that these defenses specially struggle when a white-space is introduced in the middle of a word. For example, correcting “ch rming” and “asto nding” as “cold running” and “also nothing”, which greatly change the meaning of the sentence.

\textbf{Introduction of Out of Vocabulary (OOV) tokens:} We analyze the appearance of OOV tokens after attacking with all the studied methods, we compute the percentage of sentence with at least one OOV token after the attack in all the BERT models from TextAttack.

In \cref{tab:oov}, we can observe \methodname{} only introduces the OOV token [UNK] in 11\% of the attacked sentences of the QNLI dataset. In order to avoid introducing weird characters, we only consider the characters present in the dataset we are attacking. Nevertheless, characters tokenized as [UNK] like the the U-8366 character are present in the QNLI dataset. In comparison with other methods, we notice all attacks except TextBugger produce 0\% OOV tokens. This was expected as the strategy in TextBugger is to introduce the [UNK] to force a misclassification.

\begin{table}
    \centering
    \caption{\textbf{Percentage of attacked sentences with at least one OOV token in the BERT TextAttack models:}}
    \begin{tabular}{g|g|g|g|g|g}
    \toprule
        \rowcolor{white}Attack & SST-2 & QNLI & MNLI-m & RTE & AG-News \\ 
        \midrule
        GBDA \textcolor{tabcyan}{\ding{108}} & 0 & 0 & 0 & 0 & 0 \\
        \rowcolor{white}BAE-R \textcolor{tabcyan}{\ding{108}} & 0 & 0 & 0 & 0 & 0 \\        
        BertAttack \textcolor{tabcyan}{\ding{108}} & 0 & 0 & 0 & 0 & 0 \\ 
        \rowcolor{white}TextFooler \textcolor{tabcyan}{\ding{108}} & 0 & 0 & 0 & 0 & 0 \\ 
        TextGrad \textcolor{tabcyan}{\ding{108}} & 0 & 0 & 0 & 0 & 0 \\ 
        \rowcolor{white}CWBA \textcolor{tabred}{\ding{108}} & 0 & - & - & - & 0 \\ 
        \citep{pruthi2019misspellings} \textcolor{tabred}{\ding{108}} & 0 & 0 & 0 & 0 & 0 \\ 
        \rowcolor{white}DeepWordBug \textcolor{tabred}{\ding{108}} & 0 & 0 & 0 & 0 & 0 \\ 
        TextBugger \textcolor{tabred}{\ding{108}} & 28.99 & 49.93 & 31.66 & 16.78 & 47.60 \\ 
        \rowcolor{white}\methodname{} \textcolor{tabred}{\ding{108}} & 0 & 10.76 & 0 & 0 & 0 \\ 
        \methodname{}\texttt{-Fast} \textcolor{tabred}{\ding{108}} & 0 & 9.46 & 0 & 0 & 0 \\ 
        \bottomrule
    \end{tabular}
    \label{tab:oov}
\end{table}

To complete the analysis, we have removed the characters leading to the [UNK] token from the alphabet considered during the attack in the BERT-QNLI model. In \cref{tab:unk_no_unk} we can observe the performance of \methodname{} and \methodname{}\texttt{-Fast} is minimally affected.

\begin{table}[!ht]
    \centering
    \caption{\textbf{Performance comparison of \methodname{}(\texttt{-Fast}) when allowing and not allowing the introduction of OOV tokens:}}
    \begin{tabular}{l|ggg}
    \toprule
        \rowcolor{white} & [UNK]? & ASR(\%) & $d_{\text{lev}}$ \\
    \midrule
        \multirow{2}{*}{\methodname{}} & \textcolor{tabgreen}{\cmark} & 97.68 & 1.94 \\
        & \cellcolor{white}\textcolor{tabred}{\xmark} & \cellcolor{white} 97.68 & \cellcolor{white} 1.95 \\ 
        \multirow{2}{*}{\methodname{}\texttt{-Fast}} & \textcolor{tabgreen}{\cmark} & 94.69 & 2.21 \\ 
        & \cellcolor{white} \textcolor{tabred}{\xmark} & \cellcolor{white} 94.47 & \cellcolor{white} 2.19 \\
        \bottomrule
    \end{tabular}
    \label{tab:unk_no_unk}
\end{table}

\begin{figure}
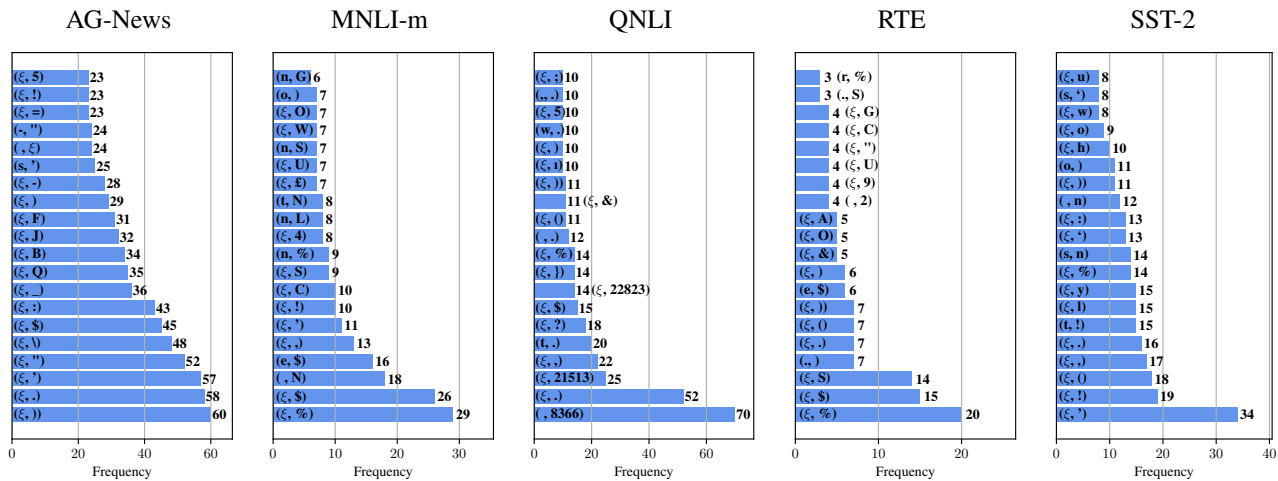

    \centering
    \setlength{\tabcolsep}{0.1\tabcolsep}
    \begin{tabular}{ccccc}
         AG-News & MNLI-m & QNLI & RTE & SST-2  \\
         \resizebox{0.2\textwidth}{!}{\input{figures/stats_attack/top_before_after_agnews_bert-base-uncased-ag-news.pgf}} & \resizebox{0.2\textwidth}{!}{\input{figures/stats_attack/top_before_after_mnli_bert-base-uncased-MNLI.pgf}} & \resizebox{0.2\textwidth}{!}{\input{figures/stats_attack/top_before_after_qnli_bert-base-uncased-QNLI.pgf}} & \resizebox{0.2\textwidth}{!}{\input{figures/stats_attack/top_before_after_rte_bert-base-uncased-RTE.pgf}} & \resizebox{0.2\textwidth}{!}{\input{figures/stats_attack/top_before_after_sst_bert-base-uncased-SST-2.pgf}}
    \end{tabular}
    \caption{\textbf{Top 20 most common replacements with \methodname{}:} The pair of characters $(c_1,c_2)$ indicates that $c_1$ is replaced by $c_2$ in the sentence. If $c_1 = \xi$, the replacement represents an insertion and if $c_2 = \xi$ the operation represents a deletion. The special character is denoted as $\xi$ as the Greek character $\xi$ did not appear in the most common operations. The most common operations are insertions of punctuation and special characters.}
    \label{fig:common_operations}
\end{figure}

\begin{figure}
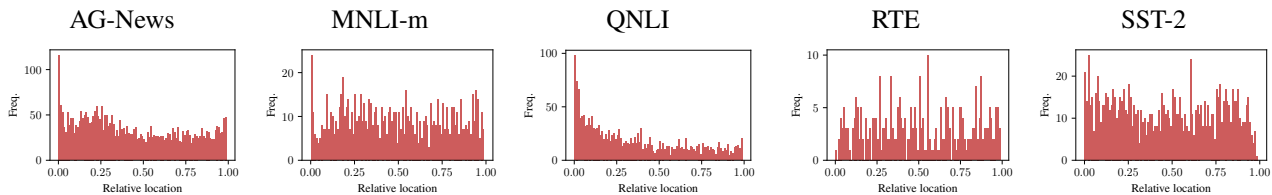

    \centering
    \setlength{\tabcolsep}{0\tabcolsep}
    \begin{tabular}{ccccc}
         AG-News & MNLI-m & QNLI & RTE & SST-2  \\
         \resizebox{0.2\textwidth}{!}{\input{figures/stats_attack/norm_loc_agnews_bert-base-uncased-ag-news.pgf}} & \resizebox{0.2\textwidth}{!}{\input{figures/stats_attack/norm_loc_mnli_bert-base-uncased-MNLI.pgf}} & \resizebox{0.2\textwidth}{!}{\input{figures/stats_attack/norm_loc_qnli_bert-base-uncased-QNLI.pgf}} & \resizebox{0.2\textwidth}{!}{\input{figures/stats_attack/norm_loc_rte_bert-base-uncased-RTE.pgf}} & \resizebox{0.2\textwidth}{!}{\input{figures/stats_attack/norm_loc_sst_bert-base-uncased-SST-2.pgf}}
    \end{tabular}
    \caption{\textbf{Distribution of the relative location of perturbations in the sentence with \methodname{}:} $0$ and $1$ represent an insertion before the first character and after the last character in the sentence respectively. We do not observe any tendency in the QNLI, RTE and SST-2 datasets. For AG-News and QNLI, the perturbations in locations closer to $0$ appear to be more common.}
    \label{fig:location_distribution}
\end{figure}

\begin{table}
    \centering
    \caption{\textbf{Attack and recognition examples in the BERT-SST-2 models defended with \citep{pruthi2019misspellings,jones2020robencodings}:}}
    \renewcommand{\arraystretch}{1.2}
    \resizebox{\textwidth}{!}{
    \begin{tabular}{g|gg|gg}
    \hline
        \toprule
        \rowcolor{white} & \multicolumn{2}{c|}{\citep{pruthi2019misspellings}} & \multicolumn{2}{c}{\citep{jones2020robencodings}} \\
        \rowcolor{white}Original & Attacked & Recognition & Attacked & Recognition \\ 
        \midrule
        \begin{minipage}{0.3\textwidth}
            it 's a charming and often affecting journey .
        \end{minipage} & \begin{minipage}{0.3\textwidth}it 's a ch rming and often affecting journey .\end{minipage} & \begin{minipage}{0.3\textwidth}it 's a cold running and often affecting journey .\end{minipage} & \begin{minipage}{0.3\textwidth}it 's a æharming and often aff ecting journ ey .\end{minipage} & \begin{minipage}{0.3\textwidth}it 's a [MASK] and often adolf eating [MASK] early .\end{minipage} \\ 
        \rowcolor{white}\begin{minipage}{0.3\textwidth}unflinchingly bleak and desperate\end{minipage} & \begin{minipage}{0.3\textwidth}unflinchingly bleak and des perate\end{minipage} & \begin{minipage}{0.3\textwidth}unflinching bleak and does pleasure\end{minipage} & \begin{minipage}{0.3\textwidth}unflinchingly bleak and desp rate\end{minipage} & \begin{minipage}{0.3\textwidth}unflinchingly break and deep role\end{minipage} \\
        \begin{minipage}{0.3\textwidth}allows us to hope that nolan is poised to embark a major career as a commercial yet inventive filmmaker .\end{minipage} & \begin{minipage}{0.3\textwidth}allows us to hope that nolan is poised td embark a major career as a commercial ye't inventive filmmaker .\end{minipage} & \begin{minipage}{0.3\textwidth}allows us to hope that nolan is polished bad embraces a major career as a commercial n't inventive filmmaker .\end{minipage} & \begin{minipage}{0.3\textwidth}allows us to hope that no an is poised to embark a major career as a commercial yet inventive filmmaker .\end{minipage} & \begin{minipage}{0.3\textwidth}angeles us to he that no an is played to embark a mother center as a commercial yet intense filmmaker .\end{minipage} \\
        \rowcolor{white} \begin{minipage}{0.3\textwidth}the acting , costumes , music , cinematography and sound are all astounding given the production 's austere locales .\end{minipage} & \begin{minipage}{0.3\textwidth}the acting , costumes , music , cinematography and sound are all antounding given the production 's austere locales .\end{minipage} & \begin{minipage}{0.3\textwidth}the acting , costumes , music , cinematography and sound are all annoying given the production 's a locales .\end{minipage} & \begin{minipage}{0.3\textwidth}the acting , costumes , music , cinematography and sound are all asto nding given the production 's austere locales .\end{minipage} & \begin{minipage}{0.3\textwidth}the among , costumes , music , cinematography and said are all also nothing given the production 's austere less .\end{minipage} \\ \bottomrule
    \end{tabular}}
    \label{tab:typo_examples}
\end{table}
\begin{table}
    \centering
    \caption{\textbf{Attack examples in the first 3 sentences of AG-News.}}
    \resizebox{\textwidth}{!}{
    \begin{tabular}{l| p{25cm} c}
        \toprule
        Method & Sentence & Prediction\\
        \midrule
\rowcolor{blue!10} Original & \texttt{Fears for T N pension after talks Unions representing workers at Turner   Newall say they are 'disappointed' after talks with stricken parent firm Federal Mogul.} &2\\ 
TextBugger \textcolor{tabred}{\ding{108}} & \texttt{Fears for T \textcolor{red}{p}\textcolor{red}{e}\textcolor{red}{r}\textcolor{red}{c}\textcolor{red}{e}\textcolor{red}{n}\textcolor{red}{t} pension after \textcolor{red}{c}\textcolor{red}{o}\textcolor{red}{n}\textcolor{red}{v}\textcolor{red}{e}\textcolor{red}{r}\textcolor{red}{s}\textcolor{red}{a}t\textcolor{red}{i}\textcolor{red}{o}\textcolor{red}{n}\textcolor{red}{ }\textcolor{red}{A}\textcolor{red}{s}s\textcolor{red}{o}\textcolor{red}{c}\textcolor{red}{i}\textcolor{red}{a}\textcolor{red}{t}ions representing workers at Turner   N\textcolor{red}{ }ewall say they are 'dsappointed' after talks with stricken parent firm Federal Mogul.} &3\\ 
TextGrad \textcolor{tabcyan}{\ding{108}} & \texttt{Fears for \textcolor{red}{t} \textcolor{red}{n} pension after \textcolor{red}{c}\textcolor{red}{h}a\textcolor{red}{t}\textcolor{red}{ }\textcolor{red}{c}\textcolor{red}{u}\textcolor{red}{s}\textcolor{red}{t}\textcolor{red}{o}\textcolor{red}{m}\textcolor{red}{e}\textcolor{red}{r}s representing \textcolor{red}{h}o\textcolor{red}{u}rs at \textcolor{red}{t}urner \textcolor{red}{n}ewall \textcolor{red}{c}\textcolor{red}{o}\textcolor{red}{m}\textcolor{red}{p}\textcolor{red}{l}a\textcolor{red}{i}\textcolor{red}{n} they are'disappointed'after \textcolor{red}{c}\textcolor{red}{h}\textcolor{red}{a}\textcolor{red}{t} with stricken parent \textcolor{red}{p}\textcolor{red}{r}\textcolor{red}{o}\textcolor{red}{v}i\textcolor{red}{d}\textcolor{red}{e}\textcolor{red}{r} \textcolor{red}{f}ederal \textcolor{red}{m}ogul.} &3\\ 
TextFooler \textcolor{tabcyan}{\ding{108}} & \texttt{Fears for T \textcolor{red}{p}\textcolor{red}{e}\textcolor{red}{r}\textcolor{red}{c}\textcolor{red}{e}\textcolor{red}{n}\textcolor{red}{t} pension after \textcolor{red}{d}\textcolor{red}{e}\textcolor{red}{b}a\textcolor{red}{t}\textcolor{red}{e}\textcolor{red}{ }\textcolor{red}{S}\textcolor{red}{y}n\textcolor{red}{d}i\textcolor{red}{c}\textcolor{red}{a}\textcolor{red}{t}\textcolor{red}{e}\textcolor{red}{s}\textcolor{red}{ }p\textcolor{red}{o}\textcolor{red}{r}\textcolor{red}{t}\textcolor{red}{r}\textcolor{red}{a}\textcolor{red}{y}\textcolor{red}{a}\textcolor{red}{l} worker at Turner   Newall say they are 'disappointed' after \textcolor{red}{c}\textcolor{red}{h}a\textcolor{red}{t}\textcolor{red}{t}\textcolor{red}{e}\textcolor{red}{r} with \textcolor{red}{b}\textcolor{red}{e}r\textcolor{red}{e}\textcolor{red}{a}\textcolor{red}{v}e\textcolor{red}{d} parent\textcolor{red}{i}\textcolor{red}{n}\textcolor{red}{g} \textcolor{red}{c}\textcolor{red}{o}\textcolor{red}{r}\textcolor{red}{p}\textcolor{red}{o}r\textcolor{red}{a}\textcolor{red}{t}\textcolor{red}{i}\textcolor{red}{o}\textcolor{red}{n}\textcolor{red}{s}\textcolor{red}{ }\textcolor{red}{C}\textcolor{red}{a}\textcolor{red}{n}a\textcolor{red}{d}\textcolor{red}{a} Mogul.} &0\\ 
DeepWordBug \textcolor{tabred}{\ding{108}} & \texttt{Fears for T N pension after alks Unions representing workers at Turner   Newall say they are 'disppointed' after ta\textcolor{red}{c}lks with stricken parent firm \textcolor{red}{G}Federal Mogul.} &2\\ 
BAE-R \textcolor{tabcyan}{\ding{108}} & \texttt{Fears for T \textcolor{red}{p}\textcolor{red}{l} pension after talks Unions representing workers at Turner   \textcolor{red}{c}\textcolor{red}{o}\textcolor{red}{n}\textcolor{red}{t}\textcolor{red}{r}\textcolor{red}{o}l\textcolor{red}{s} say they are 'disappointed' after talks with stricken parent firm Federal Mogul.} &3\\ 
BERT-attack \textcolor{tabcyan}{\ding{108}} & \texttt{Fears for T \textcolor{red}{e} pension after talks Unions representing workers at Turner   Newall say they are 'disappointed\textcolor{red}{,} after \textcolor{red}{d}\textcolor{red}{i}\textcolor{red}{s}\textcolor{red}{c}\textcolor{red}{u}\textcolor{red}{s}s\textcolor{red}{i}\textcolor{red}{o}\textcolor{red}{n} with stricken parent firm \textcolor{red}{g}\textcolor{red}{l}\textcolor{red}{o}\textcolor{red}{b}al Mogul.} &3\\ 
GBDA \textcolor{tabcyan}{\ding{108}} & \texttt{fears for t n pension after talks unions representing workers at turner newall say they are'disappointed'after talks with stricken parent \textcolor{red}{k}\textcolor{red}{n}\textcolor{red}{e}\textcolor{red}{s}\textcolor{red}{s}\textcolor{red}{e}\textcolor{red}{t} federal mogul\textcolor{red}{ }\textcolor{red}{'}} &0\\ 
CWBA \textcolor{tabred}{\ding{108}} & \texttt{fe\textcolor{red}{$\pm$}rs for t n pen\textcolor{red}{8314}ion af\textcolor{red}{1408}er ta\textcolor{red}{322}ks un\textcolor{red}{1603}ons representing wo\textcolor{red}{248}\textcolor{red}{y}ers at tu\textcolor{red}{650}ner newall say they are'dis\textcolor{red}{a}\textcolor{red}{x}a\textcolor{red}{c}\textcolor{red}{a}\textcolor{red}{l}\textcolor{red}{a}nted'after ta\textcolor{red}{38525}ks with st\textcolor{red}{f}\textcolor{red}{u}cken par\textcolor{red}{12459}nt firm federal mogul.} &3\\ 
\citep{pruthi2019misspellings} \textcolor{tabred}{\ding{108}} & \texttt{Fears for T N pension after talks Unions representing workers at T\textcolor{red}{u}urner   Ne\textcolor{red}{q}all say they are 'disappointed' after talks with stricken parent firm Federal Mogul.} &0\\ 
\methodname{} \textcolor{tabred}{\ding{108}} & \texttt{Fears for T \textcolor{red}{E} pension \textcolor{red}{:}fter talks Unions representing workers at Turner   Newall say they are 'disappointed' after talks with stricken parent firm Federal Mogul.} &3\\ 
\midrule 
\rowcolor{blue!10} Original & \texttt{The Race is On: Second Private Team Sets Launch Date for Human Spaceflight (SPACE.com) SPACE.com - TORONTO, Canada -- A second \textbackslash team of rocketeers competing for the  \#36;10 million Ansari X Prize, a contest for\textbackslash privately funded suborbital space flight, has officially announced the first\textbackslash launch date for its manned rocket.} &3\\ 
TextBugger \textcolor{tabred}{\ding{108}} & \texttt{The Race is On: Se\textcolor{red}{g}\textcolor{red}{u}nd\textcolor{red}{o}\textcolor{red}{ }\textcolor{red}{O}\textcolor{red}{w}\textcolor{red}{n} Team Sets \textcolor{red}{I}\textcolor{red}{n}\textcolor{red}{i}\textcolor{red}{t}\textcolor{red}{i}a\textcolor{red}{t}\textcolor{red}{e}\textcolor{red}{ }\textcolor{red}{S}\textcolor{red}{t}\textcolor{red}{a}\textcolor{red}{r}\textcolor{red}{d}ate for Hum\textcolor{red}{a}\textcolor{red}{n}\textcolor{red}{i}\textcolor{red}{t}\textcolor{red}{a}\textcolor{red}{r}\textcolor{red}{i}an Spaceflight (SPACE.com) S\textcolor{red}{A}\textcolor{red}{P}CE.com - \textcolor{red}{V}\textcolor{red}{A}\textcolor{red}{N}\textcolor{red}{C}O\textcolor{red}{U}\textcolor{red}{V}\textcolor{red}{E}\textcolor{red}{R}, Canad\textcolor{red}{i}a\textcolor{red}{n} -- \textcolor{red}{p}\textcolor{red}{e}\textcolor{red}{r} sec\textcolor{red}{s}\textbackslash team of rocketeers compet\textcolor{red}{e} for the  \#36;10 m\textcolor{red}{1110}llion Asari X Pri\textcolor{red}{ }ze, a contest for\textbackslash\textcolor{red}{s}\textcolor{red}{e}\textcolor{red}{c}\textcolor{red}{r}\textcolor{red}{e}tly funded suborbit\textcolor{red}{ }al space flight, has \textcolor{red}{s}\textcolor{red}{o}\textcolor{red}{l}\textcolor{red}{e}\textcolor{red}{m}\textcolor{red}{n} announced the first\textbackslash lau\textcolor{red}{1400}ch date for its man\textcolor{red}{ }ned rocket.} &3\\ 
TextGrad \textcolor{tabcyan}{\ding{108}} & \texttt{The \textcolor{red}{e}\textcolor{red}{l}\textcolor{red}{e}c\textcolor{red}{t}\textcolor{red}{i}\textcolor{red}{o}\textcolor{red}{n} is \textcolor{red}{o}n\textcolor{red}{ }: \textcolor{red}{s}econd \textcolor{red}{p}rivate \textcolor{red}{p}\textcolor{red}{a}\textcolor{red}{r}\textcolor{red}{t}\textcolor{red}{y} \textcolor{red}{s}ets \textcolor{red}{e}\textcolor{red}{v}\textcolor{red}{a} \textcolor{red}{d}ate for \textcolor{red}{h}uman \textcolor{red}{s}pace\textcolor{red}{ }\textcolor{red}{c}\textcolor{red}{a}\textcolor{red}{p}\textcolor{red}{s}\textcolor{red}{u}\textcolor{red}{l}\textcolor{red}{e} (\textcolor{red}{ }\textcolor{red}{m}\textcolor{red}{o}\textcolor{red}{o}\textcolor{red}{n}.\textcolor{red}{ }com\textcolor{red}{ }) \textcolor{red}{s}\textcolor{red}{p}\textcolor{red}{a}\textcolor{red}{c}\textcolor{red}{e}.\textcolor{red}{ }com - \textcolor{red}{p}\textcolor{red}{a}\textcolor{red}{r}\textcolor{red}{i}\textcolor{red}{s}, \textcolor{red}{c}anada -\textcolor{red}{ }- \textcolor{red}{a} \textcolor{red}{f}\textcolor{red}{o}\textcolor{red}{u}\textcolor{red}{r}\textcolor{red}{t}\textcolor{red}{h}\textcolor{red}{ }\textbackslash\textcolor{red}{ }t\textcolor{red}{r}\textcolor{red}{i}\textcolor{red}{o} of rocketeers \textcolor{red}{s}\textcolor{red}{e}e\textcolor{red}{k}ing for the \textcolor{red}{\#}\textcolor{red}{ }36\textcolor{red}{ };\textcolor{red}{ }10 million \textcolor{red}{a}nsari \textcolor{red}{x} \textcolor{red}{p}rize, a contest for\textcolor{red}{ }\textbackslash\textcolor{red}{ }privately \textcolor{red}{d}\textcolor{red}{o}\textcolor{red}{l}\textcolor{red}{l}\textcolor{red}{a}\textcolor{red}{r} suborbit\textcolor{red}{ }a\textcolor{red}{p}\textcolor{red}{o}\textcolor{red}{l}l\textcolor{red}{o} space \textcolor{red}{j}\textcolor{red}{u}\textcolor{red}{m}\textcolor{red}{p}, has o\textcolor{red}{p}\textcolor{red}{e}\textcolor{red}{n}ly announced the first\textcolor{red}{ }\textbackslash\textcolor{red}{ }launch date for its \textcolor{red}{y}\textcolor{red}{o}\textcolor{red}{u}n\textcolor{red}{g} \textcolor{red}{c}\textcolor{red}{h}\textcolor{red}{i}\textcolor{red}{l}\textcolor{red}{d}.} &0\\ 
TextFooler \textcolor{tabcyan}{\ding{108}} & \texttt{The Race is \textcolor{red}{A}\textcolor{red}{r}\textcolor{red}{o}\textcolor{red}{u}n\textcolor{red}{d}: Second Priv\textcolor{red}{y} \textcolor{red}{R}e\textcolor{red}{m}\textcolor{red}{i}\textcolor{red}{t} Set Lan\textcolor{red}{z}\textcolor{red}{a} \textcolor{red}{T}\textcolor{red}{i}\textcolor{red}{m}\textcolor{red}{e}\textcolor{red}{l}\textcolor{red}{i}\textcolor{red}{n}e for Hum\textcolor{red}{a}\textcolor{red}{n}\textcolor{red}{i}\textcolor{red}{t}\textcolor{red}{a}\textcolor{red}{r}\textcolor{red}{i}an Spaceflight (S\textcolor{red}{E}P\textcolor{red}{A}\textcolor{red}{R}A\textcolor{red}{T}\textcolor{red}{I}\textcolor{red}{O}\textcolor{red}{N}.com) S\textcolor{red}{E}P\textcolor{red}{A}\textcolor{red}{R}A\textcolor{red}{T}\textcolor{red}{I}\textcolor{red}{O}\textcolor{red}{N}.com - \textcolor{red}{C}\textcolor{red}{A}\textcolor{red}{N}\textcolor{red}{A}\textcolor{red}{D}\textcolor{red}{I}\textcolor{red}{E}\textcolor{red}{N}\textcolor{red}{S}, C\textcolor{red}{o}\textcolor{red}{u}n\textcolor{red}{t}\textcolor{red}{r}\textcolor{red}{i}\textcolor{red}{e}\textcolor{red}{s} -- \textcolor{red}{p}\textcolor{red}{a}\textcolor{red}{r}\textcolor{red}{a} second\textbackslash\textcolor{red}{s}\textcolor{red}{q}\textcolor{red}{u}a\textcolor{red}{d} of rocketeers \textcolor{red}{s}\textcolor{red}{u}\textcolor{red}{i}t\textcolor{red}{o}\textcolor{red}{r}\textcolor{red}{s} for the  \#36;10 \textcolor{red}{b}illion Ansari X \textcolor{red}{N}\textcolor{red}{o}\textcolor{red}{b}\textcolor{red}{e}\textcolor{red}{l}, a contes\textcolor{red}{t}\textcolor{red}{a}\textcolor{red}{n}t for\textbackslash\textcolor{red}{c}\textcolor{red}{o}v\textcolor{red}{e}\textcolor{red}{r}\textcolor{red}{t}ly \textcolor{red}{c}\textcolor{red}{h}\textcolor{red}{a}\textcolor{red}{m}\textcolor{red}{p}\textcolor{red}{i}\textcolor{red}{o}\textcolor{red}{n}ed suborbital space\textcolor{red}{s}\textcolor{red}{h}\textcolor{red}{i}\textcolor{red}{p} \textcolor{red}{p}l\textcolor{red}{a}\textcolor{red}{n}\textcolor{red}{e}, ha\textcolor{red}{d} \textcolor{red}{s}\textcolor{red}{o}\textcolor{red}{l}\textcolor{red}{e}\textcolor{red}{m}\textcolor{red}{n}ly \textcolor{red}{p}\textcolor{red}{r}o\textcolor{red}{c}\textcolor{red}{l}\textcolor{red}{a}\textcolor{red}{i}\textcolor{red}{m} the first\textbackslash\textcolor{red}{b}\textcolor{red}{e}\textcolor{red}{g}a\textcolor{red}{n}\textcolor{red}{ }\textcolor{red}{t}\textcolor{red}{i}\textcolor{red}{m}\textcolor{red}{e}\textcolor{red}{l}\textcolor{red}{i}\textcolor{red}{n}e for its \textcolor{red}{d}\textcolor{red}{e}\textcolor{red}{s}\textcolor{red}{o}\textcolor{red}{l}\textcolor{red}{a}\textcolor{red}{t}\textcolor{red}{e}\textcolor{red}{ }\textcolor{red}{b}\textcolor{red}{o}\textcolor{red}{m}\textcolor{red}{b}.} &0\\ 
DeepWordBug \textcolor{tabred}{\ding{108}} & \texttt{The Race is On: Second Private Tam Sets \textcolor{red}{Z}Launch Date for Hum\textcolor{red}{J}an Spaceflight (SPACE.com) SPACE.co\textcolor{red}{l} - TORONTO, Canada -- A second\textbackslash team of rocketeers competing for the  \#3;1 million Asari X Priz, a contest for\textbackslash privately \textcolor{red}{u}\textcolor{red}{f}nded suborbit\textcolor{red}{I}al space flight, has officially announced the first\textbackslash launch \textcolor{red}{B}ate for its mann\textcolor{red}{w}ed rocket.} &3\\ 
BAE-R \textcolor{tabcyan}{\ding{108}} & \texttt{The Race is On: \textcolor{red}{c}\textcolor{red}{u}\textcolor{red}{r}\textcolor{red}{r}\textcolor{red}{e}n\textcolor{red}{t} Private Team Sets \textcolor{red}{l}aunch\textcolor{red}{e}\textcolor{red}{s} Date for Human Spaceflight (SPACE.com) SPACE.com - TORONTO, \textcolor{red}{s}a\textcolor{red}{ }\textcolor{red}{-}\textcolor{red}{-}\textcolor{red}{ }\textcolor{red}{c}\textcolor{red}{a}\textcolor{red}{n}\textcolor{red}{a}\textcolor{red}{d}\textcolor{red}{a} second\textbackslash\textcolor{red}{j}\textcolor{red}{u}\textcolor{red}{r}\textcolor{red}{y} of rocketeers competing for the  \#36;10 million Ansari X \textcolor{red}{s}, a contest for\textbackslash privately f\textcolor{red}{i}\textcolor{red}{n}\textcolor{red}{a}n\textcolor{red}{c}ed suborbital space flight, has \textcolor{red}{j}\textcolor{red}{u}\textcolor{red}{s}\textcolor{red}{t} announced the first\textbackslash launch date for its \textcolor{red}{p}\textcolor{red}{r}\textcolor{red}{o}\textcolor{red}{p}\textcolor{red}{o}\textcolor{red}{s}ed rocket.} &3\\ 
BERT-attack \textcolor{tabcyan}{\ding{108}} & \texttt{The Race is On: Second Private Team Sets \textcolor{red}{l}a\textcolor{red}{n}\textcolor{red}{d}\textcolor{red}{i}\textcolor{red}{n}\textcolor{red}{g} Date for \textcolor{red}{e}\textcolor{red}{a}\textcolor{red}{r}\textcolor{red}{t}\textcolor{red}{h} Spaceflight (\textcolor{red}{m}\textcolor{red}{a}\textcolor{red}{r}\textcolor{red}{s}.com) SPACE.com - TORONTO, Canada -- \textcolor{red}{i}\textcolor{red}{s} \textcolor{red}{c}\textcolor{red}{u}\textcolor{red}{r}\textcolor{red}{r}\textcolor{red}{e}n\textcolor{red}{t}\textbackslash team of rocketeers competing for the  \#36;10 million Ansari X \textcolor{red}{q}\textcolor{red}{u}\textcolor{red}{e}\textcolor{red}{s}\textcolor{red}{t}, a contest for\textbackslash privately f\textcolor{red}{i}\textcolor{red}{n}\textcolor{red}{a}n\textcolor{red}{c}ed suborbital space \textcolor{red}{l}\textcolor{red}{a}\textcolor{red}{u}\textcolor{red}{n}\textcolor{red}{c}\textcolor{red}{h}, has officially announced the first\textbackslash la\textcolor{red}{n}\textcolor{red}{d}\textcolor{red}{i}n\textcolor{red}{g}\textcolor{red}{ }\textcolor{red}{n}\textcolor{red}{u}\textcolor{red}{m}\textcolor{red}{b}e\textcolor{red}{r} for its \textcolor{red}{a}\textcolor{red}{p}\textcolor{red}{o}\textcolor{red}{l}\textcolor{red}{l}\textcolor{red}{o} rocket.} &3\\ 
GBDA \textcolor{tabcyan}{\ding{108}} & \texttt{the race is on : second private team sets launch date for human spaceflight ( \textcolor{red}{b}\textcolor{red}{a}\textcolor{red}{r}ac\textcolor{red}{k}. com ) \textcolor{red}{c}\textcolor{red}{o}\textcolor{red}{n}\textcolor{red}{t}\textcolor{red}{i}\textcolor{red}{n}e\textcolor{red}{n}\textcolor{red}{t}\textcolor{red}{s}. com - \textcolor{red}{n}\textcolor{red}{e}\textcolor{red}{w}\textcolor{red}{s}\textcolor{red}{w}\textcolor{red}{e}\textcolor{red}{e}\textcolor{red}{k}, c\textcolor{red}{u}\textcolor{red}{b}a - - a second \textbackslash team of rocketeers competing for the \# 36 ; 10 million ansari x prize, a contest for \textbackslash privately funded suborbital space flight, has officially announced the first \textbackslash launch date for its manned rocket.} &0\\ 
CWBA \textcolor{tabred}{\ding{108}} & \texttt{t\textcolor{red}{20234}e ra\textcolor{red}{u}\textcolor{red}{e}e is on : sec\textcolor{red}{601}nd private te\textcolor{red}{4536}m sets lau\textcolor{red}{1410}ch da\textcolor{red}{1746}e for human space\textcolor{red}{969}light ( sp\textcolor{red}{*}ce. c\textcolor{red}{2}m ) sp\textcolor{red}{64257}ce. c\textcolor{red}{699}m - tor\textcolor{red}{8482}nto, can\textcolor{red}{8482}da - - a second \textbackslash team of rock\textcolor{red}{i}\textcolor{red}{a}eers com\textcolor{red}{0}\textcolor{red}{s}ting for the \# 36 ; 10 million ansari x pr\textcolor{red}{3936}ze, a contest for \textbackslash privately fun\textcolor{red}{24179}ed suborbital space flight, has off\textcolor{red}{e}\textcolor{red}{l}\textcolor{red}{d}ally announced the first \textbackslash launch date for its man\textcolor{red}{2488}ed roc\textcolor{red}{20986}et.} &3\\ 
\citep{pruthi2019misspellings} \textcolor{tabred}{\ding{108}} & \texttt{The Race is On: Second Priv\textcolor{red}{q}te Team Sets Launch Da\textcolor{red}{j}te for H\textcolor{red}{i}man Spacefight (SPA\textcolor{red}{D}E.com) SPA\textcolor{red}{D}E.com - TO\textcolor{red}{D}ONTO, Ca\textcolor{red}{h}ada -- A s\textcolor{red}{d}cond\textbackslash tam of rocketees co\textcolor{red}{p}\textcolor{red}{m}eting for the  \#36;10 million An\textcolor{red}{q}ari X P\textcolor{red}{f}ize, a cntest for\textbackslash privately fun\textcolor{red}{e}\textcolor{red}{d}d sub\textcolor{red}{l}orbital space fight, has offically announced the first\textbackslash launch date for its ma\textcolor{red}{b}ned rocket.} &3\\ 
\methodname{} \textcolor{tabred}{\ding{108}} & \texttt{The Race is On: Se\textcolor{red}{Z}ond Private Team Sets LaunchDate for \textcolor{red}{?}u\textcolor{red}{Q}an Spaceflight (SPACE\textcolor{red}{/}com) SP\textcolor{red}{D}ACE.\textcolor{red}{0}om - TORONTO, Canada \textcolor{red}{U}- A second\textcolor{red}{"}team \textcolor{red}{ }f rocketeers competing for the  \#36;10 million Ansari X Prize, a contest for\textcolor{red}{D}\textcolor{red}{ }rivatel\textcolor{red}{B} funded suborbital space flight, has officially announced \textcolor{red}{'}he first\textcolor{red}{\$}launch date for its manned rocket.} &0\\ 
\midrule 
\rowcolor{blue!10} Original & \texttt{Ky. Company Wins Grant to Study Peptides (AP) AP - A company founded by a chemistry researcher at the University of Louisville won a grant to develop a method of producing better peptides, which are short chains of amino acids, the building blocks of proteins.} &3\\ 
TextBugger \textcolor{tabred}{\ding{108}} & \texttt{Ky. Comp\textcolor{red}{n}\textcolor{red}{a}y Wins \textcolor{red}{S}\textcolor{red}{u}\textcolor{red}{b}\textcolor{red}{s}\textcolor{red}{i}\textcolor{red}{d}\textcolor{red}{i}\textcolor{red}{e}\textcolor{red}{s} to \textcolor{red}{E}\textcolor{red}{x}\textcolor{red}{a}\textcolor{red}{m}\textcolor{red}{i}\textcolor{red}{n}\textcolor{red}{e} P\textcolor{red}{p}\textcolor{red}{e}tides (AP) AP\textcolor{red}{S} - A company \textcolor{red}{b}\textcolor{red}{a}\textcolor{red}{s}ed by a chemist research\textcolor{red}{1077}r at the \textcolor{red}{A}\textcolor{red}{c}\textcolor{red}{a}\textcolor{red}{d}e\textcolor{red}{m}i\textcolor{red}{a} of \textcolor{red}{I}\textcolor{red}{n}\textcolor{red}{d}i\textcolor{red}{a}\textcolor{red}{n}\textcolor{red}{a}\textcolor{red}{p}\textcolor{red}{o}l\textcolor{red}{i}\textcolor{red}{s} won a grant to develop a method\textcolor{red}{o}\textcolor{red}{l}\textcolor{red}{o}\textcolor{red}{g}\textcolor{red}{y} of produc\textcolor{red}{t}i\textcolor{red}{o}\textcolor{red}{n} be\textcolor{red}{s}t peptides, which are \textcolor{red}{b}\textcolor{red}{r}\textcolor{red}{i}\textcolor{red}{e}\textcolor{red}{f} \textcolor{red}{s}\textcolor{red}{t}\textcolor{red}{r}in\textcolor{red}{g} of amino aids, the building block of prote\textcolor{red}{1110}ns.} &3\\ 
TextGrad \textcolor{tabcyan}{\ding{108}} & \texttt{Ky. \textcolor{red}{c}ompany \textcolor{red}{w}ins \textcolor{red}{a}\textcolor{red}{w}\textcolor{red}{a}r\textcolor{red}{d}\textcolor{red}{e}\textcolor{red}{d} to \textcolor{red}{t}\textcolor{red}{r}\textcolor{red}{e}\textcolor{red}{a}\textcolor{red}{t} \textcolor{red}{p}eptides (\textcolor{red}{ }\textcolor{red}{a}\textcolor{red}{b}\textcolor{red}{ }) \textcolor{red}{a}\textcolor{red}{o} - \textcolor{red}{a} company founded by a chemistry researcher at the \textcolor{red}{t}i\textcolor{red}{m}e of \textcolor{red}{l}ouisville won a \textcolor{red}{l}\textcolor{red}{e}a\textcolor{red}{d} to develop a \textcolor{red}{t}\textcolor{red}{r}e\textcolor{red}{a}t\textcolor{red}{m}\textcolor{red}{e}\textcolor{red}{n}\textcolor{red}{t} of producing \textcolor{red}{g}\textcolor{red}{r}e\textcolor{red}{a}ter peptides, which are short chains of \textcolor{red}{f}\textcolor{red}{a}\textcolor{red}{t}\textcolor{red}{t}\textcolor{red}{y} acids, the b\textcolor{red}{a}\textcolor{red}{s}i\textcolor{red}{s} blocks of \textcolor{red}{m}\textcolor{red}{u}\textcolor{red}{s}\textcolor{red}{c}\textcolor{red}{l}\textcolor{red}{e}.} &2\\ 
TextFooler \textcolor{tabcyan}{\ding{108}} & \texttt{Ky. \textcolor{red}{B}\textcolor{red}{u}\textcolor{red}{s}\textcolor{red}{i}\textcolor{red}{n}\textcolor{red}{e}\textcolor{red}{s}\textcolor{red}{s}\textcolor{red}{e}\textcolor{red}{s} Wins Grant to Study Peptides (\textcolor{red}{H}A\textcolor{red}{S}) \textcolor{red}{H}A\textcolor{red}{S} - A company founded by a chemistry researcher at the University of Louisville won a grant to develop a method of producing better peptides, which are short chains of amino acids, the building blocks of proteins.} &2\\ 
DeepWordBug \textcolor{tabred}{\ding{108}} & \texttt{Ky. Com\textcolor{red}{a}\textcolor{red}{p}ny Wins Grant to S\textcolor{red}{W}tudy Pepti\textcolor{red}{e}\textcolor{red}{d}s (A) AP - A company fo\textcolor{red}{R}nded by a c\textcolor{red}{F}hemistry research\textcolor{red}{f}er at the Univrsity of Louisvlle won a grant to dev\textcolor{red}{I}elop a \textcolor{red}{Q}ethod of proucing bette peptides, which are short \textcolor{red}{S}hains of amino acids, the bu\textcolor{red}{n}lding blocks of pro\textcolor{red}{f}teins.} &2\\ 
BAE-R \textcolor{tabcyan}{\ding{108}} & \texttt{Ky. m\textcolor{red}{i}\textcolor{red}{t} Wins Grant to Study Peptides (AP) AP - \textcolor{red}{b}\textcolor{red}{i}\textcolor{red}{o}\textcolor{red}{t}\textcolor{red}{e}\textcolor{red}{c}\textcolor{red}{h}\textcolor{red}{n}\textcolor{red}{o}\textcolor{red}{l}\textcolor{red}{o}\textcolor{red}{g}\textcolor{red}{y} company founded by a chemistry researcher at the University of Louisville won a grant to \textcolor{red}{s}\textcolor{red}{t}\textcolor{red}{u}\textcolor{red}{d}\textcolor{red}{y} a method of producing better peptides, which are short chain of amino acids, the building blocks of \textcolor{red}{g}e\textcolor{red}{n}\textcolor{red}{e}s.} &3\\ 
BERT-attack \textcolor{tabcyan}{\ding{108}} & \texttt{Ky. \textcolor{red}{u}\textcolor{red}{n}\textcolor{red}{i}\textcolor{red}{v}\textcolor{red}{e}\textcolor{red}{r}\textcolor{red}{s}\textcolor{red}{i}\textcolor{red}{t}y Wins \textcolor{red}{c}\textcolor{red}{o}\textcolor{red}{m}\textcolor{red}{m}\textcolor{red}{i}\textcolor{red}{s}\textcolor{red}{s}\textcolor{red}{i}\textcolor{red}{o}\textcolor{red}{n} to \textcolor{red}{r}\textcolor{red}{e}\textcolor{red}{s}\textcolor{red}{e}\textcolor{red}{a}\textcolor{red}{r}\textcolor{red}{c}\textcolor{red}{h} Peptides (AP) AP - A company o\textcolor{red}{w}ned by a chemistry \textcolor{red}{l}e\textcolor{red}{c}\textcolor{red}{t}\textcolor{red}{u}rer at the University of \textcolor{red}{k}\textcolor{red}{y} won a grant to \textcolor{red}{r}e\textcolor{red}{s}e\textcolor{red}{a}\textcolor{red}{r}\textcolor{red}{c}\textcolor{red}{h} a method of r\textcolor{red}{e}\textcolor{red}{s}\textcolor{red}{e}\textcolor{red}{a}\textcolor{red}{r}\textcolor{red}{c}\textcolor{red}{h} better peptides, which are short chains of amino acids, the building blocks of protein.} &3\\ 
GBDA \textcolor{tabcyan}{\ding{108}} & \texttt{ky. company wins grant to study peptides ( ap ) ap - a company founded by a chemistry researcher at the university of louisville won a grant to develop a method of producing better peptides, which are short chains of amino acids, the building blocks of proteins.} &3\\ 
CWBA \textcolor{tabred}{\ding{108}} & \texttt{ky. com\textcolor{red}{1405}any wi\textcolor{red}{2327}s gr\textcolor{red}{9824}nt to st\textcolor{red}{j}dy pep\textcolor{red}{65293}ides ( ap ) ap - a company f\textcolor{red}{z}\textcolor{red}{o}nded by a che\textcolor{red}{h}\textcolor{red}{r}\textcolor{red}{i}try res\textcolor{red}{s}\textcolor{red}{e}\textcolor{red}{f}cher at the un\textcolor{red}{f}\textcolor{red}{e}\textcolor{red}{i}\textcolor{red}{t}sity of louis\textcolor{red}{$^{\text{d}}$}ille w\textcolor{red}{$^{\text{a}}$}n a gr\textcolor{red}{12449}nt to deve\textcolor{red}{f}op a met\textcolor{red}{12369}od of producing better pep\textcolor{red}{1110}ides, which are short chains of am\textcolor{red}{1074}no ac\textcolor{red}{1495}ds, the building blocks of pro\textcolor{red}{f}\textcolor{red}{c}ins.} &1\\ 
\citep{pruthi2019misspellings} \textcolor{tabred}{\ding{108}} & \texttt{Ky. Compa\textcolor{red}{h}y Wins Grant to Stu\textcolor{red}{w}y Pept\textcolor{red}{d}\textcolor{red}{i}es (AP) AP - A company founded by a chemistry researcher at the U\textcolor{red}{Q}niversity of Louisville won a grant to develop a method of producing better pep\textcolor{red}{g}ides, which are short chains of amino aids, the building blocks of pro\textcolor{red}{h}eins.} &1\\ 
\methodname{} \textcolor{tabred}{\ding{108}} & \texttt{Ky. Company Wins Grant to Stu\textcolor{red}{J}dy Peptides (AP) A\textcolor{red}{F}P - A company founded by a chemistry researcher at the University of Louisville won a grant to develop a method of producing better peptides, which are short chains of amino acids, the building blocks of proteins.} &0\\ 
\bottomrule
    \end{tabular}
    }
    \label{tab:examples_agnews}
\end{table}

\begin{table}
    \centering
    \caption{\textbf{Attack examples in the first 3 sentences of MNLI-m.}}
    \resizebox{\textwidth}{!}{
    \begin{tabular}{l| p{15cm} c}
        \toprule
        Method & Sentence & Prediction\\
        \midrule
\rowcolor{blue!10} Original & \texttt{Everyone really likes the newest benefits } &2\\ 
TextBugger \textcolor{tabred}{\ding{108}} & \texttt{\textcolor{red}{S}\textcolor{red}{o}\textcolor{red}{m}\textcolor{red}{e}\textcolor{red}{b}o\textcolor{red}{d}\textcolor{red}{y} really lies the newest benefits } &0\\ 
TextGrad \textcolor{tabcyan}{\ding{108}} & \texttt{Everyone really \textcolor{red}{h}\textcolor{red}{a}\textcolor{red}{t}es the newest benefits} &0\\ 
TextFooler \textcolor{tabcyan}{\ding{108}} & \texttt{\textcolor{red}{N}\textcolor{red}{o}\textcolor{red}{b}o\textcolor{red}{d}\textcolor{red}{y} really likes the newest benefits } &0\\ 
DeepWordBug \textcolor{tabred}{\ding{108}} & \texttt{Everyone really \textcolor{red}{y}ikes the newest benefits } &0\\ 
BAE-R \textcolor{tabcyan}{\ding{108}} & \texttt{\textcolor{red}{n}\textcolor{red}{o}\textcolor{red}{b}o\textcolor{red}{d}\textcolor{red}{y} really likes the newest benefits } &0\\ 
BERT-attack \textcolor{tabcyan}{\ding{108}} & \texttt{Everyone really \textcolor{red}{h}\textcolor{red}{a}\textcolor{red}{t}es the newest benefits } &0\\ 
GBDA \textcolor{tabcyan}{\ding{108}} & \texttt{everyone really likes the newest \textcolor{red}{m}\textcolor{red}{i}\textcolor{red}{s}e\textcolor{red}{r}\textcolor{red}{y}} &0\\ 
\citep{pruthi2019misspellings} \textcolor{tabred}{\ding{108}} & \texttt{Everyone really lies the newest benefits } &0\\ 
\methodname{} \textcolor{tabred}{\ding{108}} & \texttt{Everyone really \textcolor{red}{Y}ikes the newest benefits } &0\\ 
\midrule 
\rowcolor{blue!10} Original & \texttt{The Government Executive articles housed on the website are not able to be searched.} &0\\ 
TextBugger \textcolor{tabred}{\ding{108}} & \texttt{The Government Executive articles housed on the website\textcolor{red}{s} are not a\textcolor{red}{ }ble to be searched.} &2\\ 
TextGrad \textcolor{tabcyan}{\ding{108}} & \texttt{The \textcolor{red}{g}overnment \textcolor{red}{e}xecutive articles housed on the website are not able to be \textcolor{red}{d}e\textcolor{red}{s}\textcolor{red}{t}r\textcolor{red}{o}\textcolor{red}{y}ed.} &2\\ 
TextFooler \textcolor{tabcyan}{\ding{108}} & \texttt{The Government Executive articles housed on the website are not \textcolor{red}{i}\textcolor{red}{n}\textcolor{red}{c}\textcolor{red}{a}\textcolor{red}{p}able to be searched.} &2\\ 
DeepWordBug \textcolor{tabred}{\ding{108}} & \texttt{The Government Executive articles housed on the website are not able to be sarched.} &2\\ 
BAE-R \textcolor{tabcyan}{\ding{108}} & \texttt{The Government c\textcolor{red}{a}\textcolor{red}{b}i\textcolor{red}{n}e\textcolor{red}{t} articles housed on the website are not \textcolor{red}{l}\textcolor{red}{i}\textcolor{red}{k}\textcolor{red}{e}l\textcolor{red}{y} to be searched.} &2\\ 
BERT-attack \textcolor{tabcyan}{\ding{108}} & \texttt{The Government Executive articles housed on the website are not \textcolor{red}{t}\textcolor{red}{o} to be \textcolor{red}{v}\textcolor{red}{i}\textcolor{red}{s}\textcolor{red}{i}\textcolor{red}{t}ed.} &2\\ 
GBDA \textcolor{tabcyan}{\ding{108}} & \texttt{the government executive articles housed on the website are not \textcolor{red}{s}\textcolor{red}{u}\textcolor{red}{r}e to be \textcolor{red}{r}a\textcolor{red}{i}\textcolor{red}{d}ed.} &2\\ 
\citep{pruthi2019misspellings} \textcolor{tabred}{\ding{108}} & \texttt{The Government Executive articles housed on the website are not able to be sarched.} &2\\ 
\methodname{} \textcolor{tabred}{\ding{108}} & \texttt{The Government Executive articles housed on the website are \textcolor{red}{G}ot able to be searched.} &1\\ 
\midrule 
\rowcolor{blue!10} Original & \texttt{I like him for the most part, but would still enjoy seeing someone beat him.} &1\\ 
TextBugger \textcolor{tabred}{\ding{108}} & \texttt{I like him for the most part, but would still enjoy seeing someone beat him.} &1\\ 
TextGrad \textcolor{tabcyan}{\ding{108}} & \texttt{I like him for the most \textcolor{red}{c}\textcolor{red}{e}\textcolor{red}{n}t, but would \textcolor{red}{n}\textcolor{red}{e}\textcolor{red}{v}\textcolor{red}{e}\textcolor{red}{r} enjoy seeing someone beat him.} &0\\ 
TextFooler \textcolor{tabcyan}{\ding{108}} & \texttt{I like him for the most p\textcolor{red}{o}rt\textcolor{red}{i}\textcolor{red}{o}\textcolor{red}{n}, but would still \textcolor{red}{c}\textcolor{red}{h}e\textcolor{red}{r}\textcolor{red}{i}\textcolor{red}{s}\textcolor{red}{h}\textcolor{red}{e}\textcolor{red}{s} seeing someone \textcolor{red}{c}\textcolor{red}{o}\textcolor{red}{n}\textcolor{red}{q}\textcolor{red}{u}e\textcolor{red}{r}\textcolor{red}{i}\textcolor{red}{n}\textcolor{red}{g} him.} &2\\ 
DeepWordBug \textcolor{tabred}{\ding{108}} & \texttt{I like him for the most art, but would still enjoy seeing someone beat him.} &2\\ 
BAE-R \textcolor{tabcyan}{\ding{108}} & \texttt{I like him for the most p\textcolor{red}{e}\textcolor{red}{o}\textcolor{red}{p}\textcolor{red}{l}\textcolor{red}{e}, but would still enjoy seeing someone beat him.} &2\\ 
BERT-attack \textcolor{tabcyan}{\ding{108}} & \texttt{I like him for the most a\textcolor{red}{l}\textcolor{red}{l}, but would still enjoy seeing someone beat him.} &2\\ 
GBDA \textcolor{tabcyan}{\ding{108}} & \texttt{i like him for the most part, but \textcolor{red}{h}o\textcolor{red}{w}\textcolor{red}{a}\textcolor{red}{r}d \textcolor{red}{a}\textcolor{red}{l}\textcolor{red}{w}\textcolor{red}{a}\textcolor{red}{y}\textcolor{red}{s} \textcolor{red}{r}e\textcolor{red}{g}\textcolor{red}{r}\textcolor{red}{e}\textcolor{red}{t}\textcolor{red}{s} seeing someone beat him.} &2\\ 
\citep{pruthi2019misspellings} \textcolor{tabred}{\ding{108}} & \texttt{I like him for the most part, but would still enjoy se\textcolor{red}{w}ing someone beat him.} &2\\ 
\methodname{} \textcolor{tabred}{\ding{108}} & \texttt{I like him for the most p\textcolor{red}{4}art, but would still enjoy seeing someone beat him.} &2\\ 
\bottomrule 
    \end{tabular}
    }
    \label{tab:examples_mnli}
\end{table}

\begin{table}
    \centering
    \caption{\textbf{Attack examples in the first 3 sentences of QNLI.}}
    \resizebox{\textwidth}{!}{
    \begin{tabular}{l| p{15cm} c}
        \toprule
        Method & Sentence & Prediction\\
        \midrule
\rowcolor{blue!10} Original & \texttt{As of that day, the new constitution heralding the Second Republic came into force.} &0\\ 
TextBugger \textcolor{tabred}{\ding{108}} & \texttt{As of that day, the new constitution heralding the Second Republic\textcolor{red}{s} came into for\textcolor{red}{1010}e.} &1\\ 
TextGrad \textcolor{tabcyan}{\ding{108}} & \texttt{As of that day, the new constitution heralding the \textcolor{red}{s}econd \textcolor{red}{r}epublic \textcolor{red}{r}\textcolor{red}{e}\textcolor{red}{g}\textcolor{red}{i}\textcolor{red}{s}\textcolor{red}{t}\textcolor{red}{e}\textcolor{red}{r}e\textcolor{red}{d} into \textcolor{red}{r}\textcolor{red}{e}\textcolor{red}{a}\textcolor{red}{l}.} &1\\ 
TextFooler \textcolor{tabcyan}{\ding{108}} & \texttt{As of that day, the new constitution heralding the Second Republic\textcolor{red}{s} \textcolor{red}{w}\textcolor{red}{e}\textcolor{red}{n}\textcolor{red}{t} into \textcolor{red}{t}\textcolor{red}{r}o\textcolor{red}{u}\textcolor{red}{p}e\textcolor{red}{s}.} &1\\ 
DeepWordBug \textcolor{tabred}{\ding{108}} & \texttt{As of that day, the new constitution heralding the Second Republic came into \textcolor{red}{o}\textcolor{red}{f}rce.} &1\\ 
BAE-R \textcolor{tabcyan}{\ding{108}} & \texttt{As of that d\textcolor{red}{o}\textcolor{red}{c}\textcolor{red}{u}\textcolor{red}{m}\textcolor{red}{e}\textcolor{red}{n}\textcolor{red}{t}, the new constitution heralding the Second \textcolor{red}{r}epublic came into \textcolor{red}{e}\textcolor{red}{x}\textcolor{red}{i}\textcolor{red}{s}\textcolor{red}{t}\textcolor{red}{e}\textcolor{red}{n}ce.} &0\\ 
BERT-attack \textcolor{tabcyan}{\ding{108}} & \texttt{As of that day, the new constitution herald\textcolor{red}{o}\textcolor{red}{f} the Second Republic came into \textcolor{red}{r}\textcolor{red}{e}\textcolor{red}{a}\textcolor{red}{l}.} &1\\ 
GBDA \textcolor{tabcyan}{\ding{108}} & \texttt{as of that day, the new constitution heralding the second republic came into force.} &0\\ 
\citep{pruthi2019misspellings} \textcolor{tabred}{\ding{108}} & \texttt{As of that day, the new constitution heralding the Second Republic came into for\textcolor{red}{d}e.} &1\\ 
\methodname{} \textcolor{tabred}{\ding{108}} & \texttt{As of that day, the new constitution heralding the Second Republic came into for\textcolor{red}{\$}ce.} &1\\ 
\midrule 
\rowcolor{blue!10} Original & \texttt{The most important tributaries in this area are the Ill below of Strasbourg, the Neckar in Mannheim and the Main across from Mainz.} &1\\ 
TextBugger \textcolor{tabred}{\ding{108}} & \texttt{The most important tributaries in this area\textcolor{red}{s} are the Ill below of Strasbourg, the Neckar in Mannheim and the Main across from Mainz.} &0\\ 
TextGrad \textcolor{tabcyan}{\ding{108}} & \texttt{The most important tributaries in this area are the \textcolor{red}{i}ll below of \textcolor{red}{s}trasbourg, the \textcolor{red}{n}eckar in \textcolor{red}{m}annheim and the \textcolor{red}{m}ain across from \textcolor{red}{c}\textcolor{red}{i}\textcolor{red}{n}\textcolor{red}{c}i\textcolor{red}{n}n\textcolor{red}{a}\textcolor{red}{t}\textcolor{red}{i}.} &0\\ 
TextFooler \textcolor{tabcyan}{\ding{108}} & \texttt{The most important tributaries in this area\textcolor{red}{s} are the Ill below of Strasbourg, the Neckar in Mannheim and the Main across from Mainz.} &0\\ 
DeepWordBug \textcolor{tabred}{\ding{108}} & \texttt{The most important tributaries in this area are the I\textcolor{red}{A}l below of Strasbourg, the Neckar in Mannheim and the Main across from Mainz.} &0\\ 
BAE-R \textcolor{tabcyan}{\ding{108}} & \texttt{The most important tributaries in this \textcolor{red}{s}\textcolor{red}{e}\textcolor{red}{c}\textcolor{red}{t}\textcolor{red}{o}\textcolor{red}{r} are the Ill below of Strasbourg, the Neckar in Mannheim and the Main across from Mainz.} &0\\ 
BERT-attack \textcolor{tabcyan}{\ding{108}} & \texttt{The most important tributaries in this area are the \textcolor{red}{f}\textcolor{red}{a}\textcolor{red}{r} below of Strasbourg, the Neckar in Mannheim and the Main across from Mainz.} &0\\ 
GBDA \textcolor{tabcyan}{\ding{108}} & \texttt{the most important tributaries in this area are the ill below of strasbourg, the neckar in mannheim and the \textcolor{red}{j}\textcolor{red}{e}\textcolor{red}{d}\textcolor{red}{i} across from mainz.} &0\\ 
\citep{pruthi2019misspellings} \textcolor{tabred}{\ding{108}} & \texttt{The most important tributaries in this area are the Ill below of Strasbourg, the Neckar in Mannheim and the Main across from Mainz.} &1\\ 
\methodname{} \textcolor{tabred}{\ding{108}} & \texttt{The most i\textcolor{red}{ð}portant tributaries in this area are the Ill below of Strasbourg, the Neckar in Mannheim and the Main across from Mainz.} &0\\ 
\midrule 
\rowcolor{blue!10} Original & \texttt{In most provinces a second Bachelor's Degree such as a Bachelor of Education is required to become a qualified teacher.} &1\\ 
TextBugger \textcolor{tabred}{\ding{108}} & \texttt{In most provinces a s\textcolor{red}{1077}cond Bac\textcolor{red}{1392}elor's Degree\textcolor{red}{s} such as a Bachelo\textcolor{red}{11397} of Education is required to become a qualified teacher.} &0\\ 
TextGrad \textcolor{tabcyan}{\ding{108}} & \texttt{In most provinces a \textcolor{red}{m}\textcolor{red}{i}\textcolor{red}{n}\textcolor{red}{i}\textcolor{red}{m}\textcolor{red}{u}\textcolor{red}{m} \textcolor{red}{b}achelor's \textcolor{red}{d}egree such as a \textcolor{red}{b}achelor of \textcolor{red}{e}ducation is required to become a qualified teacher.} &0\\ 
TextFooler \textcolor{tabcyan}{\ding{108}} & \texttt{\textcolor{red}{i}n\textcolor{red}{g} most provinces a second Bachelor's \textcolor{red}{G}\textcolor{red}{r}\textcolor{red}{a}\textcolor{red}{d}e\textcolor{red}{s} such as a \textcolor{red}{D}\textcolor{red}{i}\textcolor{red}{p}\textcolor{red}{l}\textcolor{red}{o}\textcolor{red}{m}\textcolor{red}{a}\textcolor{red}{s} of \textcolor{red}{T}u\textcolor{red}{i}tion is required to become a qualified teacher.} &0\\ 
DeepWordBug \textcolor{tabred}{\ding{108}} & \texttt{I most provinces a \textcolor{red}{q}econd Bachelor's \textcolor{red}{W}egree such as a B\textcolor{red}{F}chelor of ducation is required to beome a qualified teachr.} &0\\ 
BAE-R \textcolor{tabcyan}{\ding{108}} & \texttt{In most provinces a \textcolor{red}{b}\textcolor{red}{a}\textcolor{red}{s}\textcolor{red}{i}\textcolor{red}{c} Bachelor's Degree such as a \textcolor{red}{b}achelor of \textcolor{red}{s}\textcolor{red}{t}u\textcolor{red}{d}i\textcolor{red}{e}\textcolor{red}{s} is \textcolor{red}{n}e\textcolor{red}{e}\textcolor{red}{d}ed to become a qualified teacher.} &1\\ 
BERT-attack \textcolor{tabcyan}{\ding{108}} & \texttt{In most \textcolor{red}{c}\textcolor{red}{a}\textcolor{red}{n}\textcolor{red}{a}\textcolor{red}{d}\textcolor{red}{i}\textcolor{red}{a}\textcolor{red}{n} a \textcolor{red}{d}\textcolor{red}{i}\textcolor{red}{p}\textcolor{red}{l}o\textcolor{red}{m}\textcolor{red}{a} \textcolor{red}{b}achelor\textcolor{red}{t}\textcolor{red}{h}\textcolor{red}{e}\textcolor{red}{t}\textcolor{red}{h}\textcolor{red}{e} Degree such as a \textcolor{red}{m}a\textcolor{red}{j}or of Education is required to become a qualified teacher.} &0\\ 
GBDA \textcolor{tabcyan}{\ding{108}} & \texttt{in most provinces a second bachelor's degree such as a bachelor of education is \textcolor{red}{m}\textcolor{red}{i}\textcolor{red}{n}i\textcolor{red}{m}\textcolor{red}{u}\textcolor{red}{m} to become a qualified teacher.} &0\\ 
\citep{pruthi2019misspellings} \textcolor{tabred}{\ding{108}} & \texttt{In most provinces a second Bachelor's Degree such as a Bachelor of Education is required to become a qualified teacher.} &1\\ 
\methodname{} \textcolor{tabred}{\ding{108}} & \texttt{In most provinces a\textcolor{red}{2}second Bachelor's Degree such as a Bachelor of Education is required to become a qualified teacher.} &0\\ 
\bottomrule 
    \end{tabular}
    }
    \label{tab:examples_qnli}
\end{table}

\begin{table}
    \centering
    \caption{\textbf{Attack examples in the first 3 sentences of RTE.}}
    \resizebox{\textwidth}{!}{
    \begin{tabular}{l| p{15cm} c}
        \toprule
        Method & Sentence & Prediction\\
        \midrule
\rowcolor{blue!10} Original & \texttt{Christopher Reeve had an accident.} &1\\ 
TextBugger \textcolor{tabred}{\ding{108}} & \texttt{Christopher Reeve had an accid\textcolor{red}{ }ent.} &1\\ 
TextGrad \textcolor{tabcyan}{\ding{108}} & \texttt{Christopher \textcolor{red}{r}eeve had an c\textcolor{red}{a}\textcolor{red}{r}e\textcolor{red}{e}\textcolor{red}{r}.} &0\\ 
TextFooler \textcolor{tabcyan}{\ding{108}} & \texttt{\textcolor{red}{V}\textcolor{red}{a}\textcolor{red}{l}\textcolor{red}{e}\textcolor{red}{r}\textcolor{red}{i}\textcolor{red}{a} Reeve \textcolor{red}{w}a\textcolor{red}{s} an \textcolor{red}{c}\textcolor{red}{o}\textcolor{red}{l}\textcolor{red}{l}i\textcolor{red}{s}\textcolor{red}{i}\textcolor{red}{o}n\textcolor{red}{s}.} &0\\ 
DeepWordBug \textcolor{tabred}{\ding{108}} & \texttt{Christopher Reeve had an accidnt.} &1\\ 
BAE-R \textcolor{tabcyan}{\ding{108}} & \texttt{\textcolor{red}{k}\textcolor{red}{a}re\textcolor{red}{n} Reeve had an accident.} &1\\ 
BERT-attack \textcolor{tabcyan}{\ding{108}} & \texttt{\textcolor{red}{d}\textcolor{red}{a}\textcolor{red}{v}i\textcolor{red}{d} Reeve had an \textcolor{red}{s}\textcolor{red}{t}\textcolor{red}{r}\textcolor{red}{o}\textcolor{red}{k}e.} &0\\ 
GBDA \textcolor{tabcyan}{\ding{108}} & \texttt{christopher reeve had an \textcolor{red}{stroke}.} &0\\ 
\citep{pruthi2019misspellings} \textcolor{tabred}{\ding{108}} & \texttt{Christopher Reeve had an acci\textcolor{red}{S}dent.} &1\\ 
\methodname{} \textcolor{tabred}{\ding{108}} & \texttt{Christopher Reeve had an accident\textcolor{red}{S}} &0\\ 
\midrule 
\rowcolor{blue!10} Original & \texttt{Pennsylvania has the biggest Amish community in the U.S.} &1\\ 
TextBugger \textcolor{tabred}{\ding{108}} & \texttt{Penn has the \textcolor{red}{l}\textcolor{red}{a}\textcolor{red}{r}gest Amish community in the U.S.} &0\\ 
TextFooler \textcolor{tabcyan}{\ding{108}} & \texttt{Pennsylvania has the \textcolor{red}{w}i\textcolor{red}{d}e Amish community in the U.S.} &0\\ 
DeepWordBug \textcolor{tabred}{\ding{108}} & \texttt{Pennsylvania has the bigges Amish community in the U.S.} &1\\ 
BAE-R \textcolor{tabcyan}{\ding{108}} & \texttt{Pennsylvania has the \textcolor{red}{l}\textcolor{red}{a}\textcolor{red}{r}ge Amish community in the U.S.} &0\\ 
BERT-attack \textcolor{tabcyan}{\ding{108}} & \texttt{Pennsylvania has the \textcolor{red}{h}\textcolor{red}{u}ge Amish community in the U.S.} &1\\ 
GBDA \textcolor{tabcyan}{\ding{108}} & \texttt{pennsylvania has the \textcolor{red}{strongest} amish community in the \textcolor{red}{state. nara geographical}} &0\\ 
\citep{pruthi2019misspellings} \textcolor{tabred}{\ding{108}} & \texttt{Pennsylvania has the bigge\textcolor{red}{a}t Amish community in the U.S.} &0\\
\methodname{} \textcolor{tabred}{\ding{108}} & \texttt{Pennsylvania has the bigge\textcolor{red}{A}t Amish community in the U.S.} &0\\ 
\midrule 
\rowcolor{blue!10} Original & \texttt{Security forces were on high alert after a campaign marred by violence.} &0\\ 
TextBugger \textcolor{tabred}{\ding{108}} & \texttt{Security forces were on high alert after a c\textcolor{red}{o}\textcolor{red}{u}\textcolor{red}{n}\textcolor{red}{t}\textcolor{red}{r}\textcolor{red}{y}\textcolor{red}{s}i\textcolor{red}{d}\textcolor{red}{e} marred by violence.} &1\\ 
TextGrad \textcolor{tabcyan}{\ding{108}} & \texttt{Security f\textcolor{red}{a}\textcolor{red}{m}\textcolor{red}{i}\textcolor{red}{l}\textcolor{red}{i}es were on high alert after a m\textcolor{red}{o}\textcolor{red}{n}\textcolor{red}{t}\textcolor{red}{h} marred by violence.} &1\\ 
TextFooler \textcolor{tabcyan}{\ding{108}} & \texttt{Security forces were on high alert after a c\textcolor{red}{o}\textcolor{red}{u}\textcolor{red}{n}\textcolor{red}{t}\textcolor{red}{r}\textcolor{red}{y}\textcolor{red}{s}i\textcolor{red}{d}\textcolor{red}{e} marred by violence.} &1\\ 
DeepWordBug \textcolor{tabred}{\ding{108}} & \texttt{Security forces were on high alert after a cmpaign marred by violence.} &1\\ 
BAE-R \textcolor{tabcyan}{\ding{108}} & \texttt{\textcolor{red}{r}\textcolor{red}{a}\textcolor{red}{n}\textcolor{red}{s}\textcolor{red}{p}\textcolor{red}{o}\textcolor{red}{r}\textcolor{red}{t} force were on high alert after a campaig\textcolor{red}{n}\textcolor{red}{i}n\textcolor{red}{g} marred by violence.} &1\\ 
GBDA \textcolor{tabcyan}{\ding{108}} & \texttt{security forces were on high alert after a campaign marred by \textcolor{red}{marches}.} & 1\\
\citep{pruthi2019misspellings} \textcolor{tabred}{\ding{108}} & \texttt{Security forces were on high alert after a campai\textcolor{red}{r}n marred by violence.} &1\\ 
\methodname{} \textcolor{tabred}{\ding{108}} & \texttt{Security forces were on high alert after a\textcolor{red}{2}campaign marred by violence.} &1\\ 
\bottomrule 
    \end{tabular}
    }
    \label{tab:examples_rte}
\end{table}

\begin{table}
    \centering
    \caption{\textbf{Attack examples in the first 3 sentences of SST-2.}}
    \resizebox{\textwidth}{!}{
    \begin{tabular}{l| p{15cm} c}
        \toprule
        Method & Sentence & Prediction\\
        \midrule
\rowcolor{blue!10} Original & \texttt{it 's a charming and often affecting journey . } &1\\ 
TextBugger \textcolor{tabred}{\ding{108}} & \texttt{it 's a ch\textcolor{red}{593}rming and often affecting \textcolor{red}{v}o\textcolor{red}{y}\textcolor{red}{a}\textcolor{red}{g}e . } &1\\ 
TextGrad \textcolor{tabcyan}{\ding{108}} & \texttt{it's a \textcolor{red}{d}\textcolor{red}{a}\textcolor{red}{n}\textcolor{red}{g}\textcolor{red}{e}\textcolor{red}{r}\textcolor{red}{o}\textcolor{red}{u}\textcolor{red}{s} and often affecting \textcolor{red}{t}\textcolor{red}{r}\textcolor{red}{a}\textcolor{red}{v}\textcolor{red}{e}\textcolor{red}{l}\textcolor{red}{l}\textcolor{red}{i}\textcolor{red}{n}\textcolor{red}{g}\textcolor{red}{.}} &0\\ 
TextFooler \textcolor{tabcyan}{\ding{108}} & \texttt{it 's a c\textcolor{red}{u}\textcolor{red}{t}i\textcolor{red}{e} and often aff\textcolor{red}{l}\textcolor{red}{i}ct\textcolor{red}{s} journey . } &0\\ 
DeepWordBug \textcolor{tabred}{\ding{108}} & \texttt{it 's a \textcolor{red}{W}charming and o\textcolor{red}{t}\textcolor{red}{f}en aff\textcolor{red}{c}\textcolor{red}{e}ting journey . } &0\\ 
BAE-R \textcolor{tabcyan}{\ding{108}} & \texttt{it 's a \textcolor{red}{d}ar\textcolor{red}{k} and often \textcolor{red}{w}\textcolor{red}{i}\textcolor{red}{n}\textcolor{red}{d}ing journey . } &0\\ 
BERT-attack \textcolor{tabcyan}{\ding{108}} & \texttt{it 's a \textcolor{red}{o}n\textcolor{red}{e} and often a\textcolor{red}{n}\textcolor{red}{o}t\textcolor{red}{h}\textcolor{red}{e}\textcolor{red}{r} journey . } &1\\ 
GBDA \textcolor{tabcyan}{\ding{108}} & \texttt{it's a \textcolor{red}{colourful} and \textcolor{red}{not} affecting journey.} &0\\ 
CWBA \textcolor{tabred}{\ding{108}} & \texttt{it's a char\textcolor{red}{640}ing a\textcolor{red}{→}d of\textcolor{red}{$^{\text{w}}$}en affe\textcolor{red}{37070}ting jo\textcolor{red}{1657}ney.} &0\\ 
\citep{pruthi2019misspellings} \textcolor{tabred}{\ding{108}} & \texttt{it 's a chrming and often a\textcolor{red}{c}fecting journey . } &0\\ 
\methodname{} \textcolor{tabred}{\ding{108}} & \texttt{it 's a \textcolor{red}{\%}harming and often affecting journey . } &0\\ 
\midrule 
\rowcolor{blue!10} Original & \texttt{unflinchingly bleak and desperate } &0\\ 
TextBugger \textcolor{tabred}{\ding{108}} & \texttt{unflinchingly \textcolor{red}{s}\textcolor{red}{o}\textcolor{red}{m}\textcolor{red}{b}\textcolor{red}{e}\textcolor{red}{r} and desperate } &1\\ 
TextGrad \textcolor{tabcyan}{\ding{108}} & \texttt{unflinchingly \textcolor{red}{d}\textcolor{red}{a}\textcolor{red}{r}k and desperate} &1\\ 
TextFooler \textcolor{tabcyan}{\ding{108}} & \texttt{unflinchingly \textcolor{red}{e}\textcolor{red}{e}\textcolor{red}{r}\textcolor{red}{i}\textcolor{red}{e} and desperate } &1\\ 
DeepWordBug \textcolor{tabred}{\ding{108}} & \texttt{unflinchingly blak and desper\textcolor{red}{t}\textcolor{red}{a}e } &1\\ 
BAE-R \textcolor{tabcyan}{\ding{108}} & \texttt{unflinchingly \textcolor{red}{h}\textcolor{red}{a}\textcolor{red}{p}\textcolor{red}{p}\textcolor{red}{y} and desperate } &1\\ 
BERT-attack \textcolor{tabcyan}{\ding{108}} & \texttt{unflinchingly \textcolor{red}{d}\textcolor{red}{a}\textcolor{red}{r}k and desperate } &1\\ 
GBDA \textcolor{tabcyan}{\ding{108}} & \texttt{unflinchingly \textcolor{red}{picturesque} and desperate} &1\\ 
CWBA \textcolor{tabred}{\ding{108}} & \texttt{unfl\textcolor{red}{30340}\textcolor{red}{a}\textcolor{red}{r}ingly b\textcolor{red}{y}\textcolor{red}{a}ak a\textcolor{red}{y}\textcolor{red}{o}d des\textcolor{red}{h}\textcolor{red}{i}\textcolor{red}{l}\textcolor{red}{a}rate} &1\\ 
\citep{pruthi2019misspellings} \textcolor{tabred}{\ding{108}} & \texttt{unflinchingly beak and des\textcolor{red}{e}\textcolor{red}{p}rate } &1\\ 
\methodname{} \textcolor{tabred}{\ding{108}} & \texttt{unflinchingly \textcolor{red}{à}bleak and desperate } &1\\ 
\midrule 
\rowcolor{blue!10} Original & \texttt{allows us to hope that nolan is poised to embark a major career as a commercial yet inventive filmmaker . } &1\\ 
TextBugger \textcolor{tabred}{\ding{108}} & \texttt{allows \textcolor{red}{n}\textcolor{red}{o}us to hope that nolan is poised to embark a major career as a commercial \textcolor{red}{h}\textcolor{red}{o}\textcolor{red}{w}\textcolor{red}{e}\textcolor{red}{v}e\textcolor{red}{r} invntive \textcolor{red}{c}i\textcolor{red}{n}\textcolor{red}{e}m\textcolor{red}{a}\textcolor{red}{t}\textcolor{red}{o}\textcolor{red}{g}\textcolor{red}{r}a\textcolor{red}{p}\textcolor{red}{h}er\textcolor{red}{s} . } &0\\ 
TextGrad \textcolor{tabcyan}{\ding{108}} & \texttt{allows us to \textcolor{red}{a}\textcolor{red}{r}\textcolor{red}{g}\textcolor{red}{u}e that nolan is \textcolor{red}{i}\textcolor{red}{n}\textcolor{red}{e}\textcolor{red}{l}\textcolor{red}{i}\textcolor{red}{g}i\textcolor{red}{b}\textcolor{red}{l}\textcolor{red}{e} to embark a major career as a commercial \textcolor{red}{f}\textcolor{red}{a}\textcolor{red}{i}\textcolor{red}{l}\textcolor{red}{s} inventive filmmaker.} &0\\ 
TextFooler \textcolor{tabcyan}{\ding{108}} & \texttt{allows \textcolor{red}{o}u\textcolor{red}{r}\textcolor{red}{s}\textcolor{red}{e}\textcolor{red}{l}\textcolor{red}{v}\textcolor{red}{e}s to hope that nolan is poised to embark\textcolor{red}{e}\textcolor{red}{d} a \textcolor{red}{s}\textcolor{red}{e}\textcolor{red}{v}\textcolor{red}{e}r\textcolor{red}{e} career as a commercial yet \textcolor{red}{n}\textcolor{red}{o}ve\textcolor{red}{l}t\textcolor{red}{y}\textcolor{red}{ }\textcolor{red}{s}\textcolor{red}{u}\textcolor{red}{p}\textcolor{red}{e}\textcolor{red}{r}i\textcolor{red}{n}\textcolor{red}{t}\textcolor{red}{e}\textcolor{red}{n}\textcolor{red}{d}e\textcolor{red}{n}\textcolor{red}{t} . } &0\\ 
DeepWordBug \textcolor{tabred}{\ding{108}} & \texttt{allows \textcolor{red}{G}s to hope that nolan is \textcolor{red}{L}oised to embark a major career as a commercial yet inve\textcolor{red}{w}ntive filmmaker . } &0\\ 
BAE-R \textcolor{tabcyan}{\ding{108}} & \texttt{allows \textcolor{red}{i}\textcolor{red}{t} to hope that nolan is poised to \textcolor{red}{a}\textcolor{red}{s}\textcolor{red}{s}\textcolor{red}{u}\textcolor{red}{m}\textcolor{red}{e} a major career as a commercial yet \textcolor{red}{a}\textcolor{red}{m}\textcolor{red}{a}t\textcolor{red}{e}\textcolor{red}{u}\textcolor{red}{r} filmmaker . } &1\\ 
BERT-attack \textcolor{tabcyan}{\ding{108}} & \texttt{allows \textcolor{red}{t}\textcolor{red}{o} to hope that nolan is \textcolor{red}{e}\textcolor{red}{l}\textcolor{red}{i}\textcolor{red}{g}i\textcolor{red}{b}\textcolor{red}{l}\textcolor{red}{e} to embark a major career as a commercial yet invent\textcolor{red}{e}\textcolor{red}{x}\textcolor{red}{p}\textcolor{red}{e}\textcolor{red}{r}i\textcolor{red}{e}\textcolor{red}{n}\textcolor{red}{c}e\textcolor{red}{d} filmmak\textcolor{red}{i}\textcolor{red}{n}\textcolor{red}{g} . } &0\\ 
GBDA \textcolor{tabcyan}{\ding{108}} & \texttt{allows us to \textcolor{red}{doubt} that nolan is poised to embark a major career as a commercial \textcolor{red}{lower} inventive \textcolor{red}{writer}.} &0\\ 
CWBA \textcolor{tabred}{\ding{108}} & \texttt{all\textcolor{red}{21335}ws us to ho\textcolor{red}{26954}e that nolan is po\textcolor{red}{2313}sed to embark a major career as a comm\textcolor{red}{b}\textcolor{red}{e}cial y\textcolor{red}{1705}t inventive filmmaker.} &0\\ 
\citep{pruthi2019misspellings} \textcolor{tabred}{\ding{108}} & \texttt{allows us to hope that nolan is poised to ebark a major career as a commercial yet in\textcolor{red}{f}entive filmmaker . } &0\\ 
\methodname{} \textcolor{tabred}{\ding{108}} & \texttt{allows us to hope that no\textcolor{red}{\$}an is poised to embark a major career as a commercial yet inventive filmmaker . } &0\\ 
\bottomrule 
    \end{tabular}
    }
    \label{tab:examples_sst}
\end{table}

\subsection{TextAttack Baseline}
In this section we complement the analysis in \cref{subsec:sota} by reporting results for the TextAttack ALBERT models \citep{Lan2020albert}. In \cref{tab:textattack_albert}, we can observe \methodname{} consistently attains the highest ASR among all the studied methods, while obtaining the lowest Levenshtein distance in $4/5$ cases and highest similarity in $3/5$ cases.

\begin{table}
    \centering
    \caption{\textbf{Attack evaluation in the TextAttack ALBERT models:} Token-level and character-level attacks are highlighted with \textcolor{tabcyan}{\ding{108}} and \textcolor{tabred}{\ding{108}} respectively. for each metric, the best method is highlighted in \textbf{bold} and the runner-up in \underline{underlined}. \methodname{} consistently achieves highest Attack Success Rate (ASR).}
    \resizebox{0.75\textwidth}{!}{
    \begin{tabular}{cg|vggg}
        \toprule
        \rowcolor{white}
            & & \multicolumn{4}{c}{ALBERT} \\
        \rowcolor{white}
        & Method & ASR (\%) $\bm{\uparrow}$ & $d_{\text{lev}}(S,S')~ \bm{\downarrow}$ & $\text{Sim}(S,S')~ \bm{\uparrow}$ & $\text{Time (s)} ~ \bm{\downarrow}$ \\
        \midrule
        \parbox[t]{3mm}{\multirow{10}{*}{\rotatebox[origin=c]{90}{AG-News}}} & CWBA \textcolor{tabred}{\ding{108}} & $57.96$ & $22.69_{\pm (21.07)}$ & $0.64_{\pm (0.22)}$ & $205.69_{\pm (162.00)}$ \\
         & \cellcolor{white} BAE-R \textcolor{tabcyan}{\ding{108}} & \cellcolor{white}$18.26$ &\cellcolor{white} $15.15_{\pm (11.28)}$ & \cellcolor{white}$\mathbf{0.97}_{\pm (0.02)}$ &\cellcolor{white} $1.84_{\pm (1.70)}$ \\
         & BERT-attack \textcolor{tabcyan}{\ding{108}} & $37.23$ & $21.34_{\pm (15.29)}$ & $0.93_{\pm (0.05)}$ & $2.41_{\pm (2.14)}$ \\
         & \cellcolor{white} DeepWordBug \textcolor{tabred}{\ding{108}} &\cellcolor{white} $56.90$ & \cellcolor{white}$9.77_{\pm (6.77)}$ &\cellcolor{white} $0.83_{\pm (0.14)}$ & \cellcolor{white}$\mathbf{0.73}_{\pm (0.40)}$ \\
         & TextBugger \textcolor{tabred}{\ding{108}} & $71.76$ & $17.48_{\pm (16.82)}$ & $0.91_{\pm (0.06)}$ & $\underline{1.38}_{\pm (0.98)}$ \\
         & \cellcolor{white} TextFooler \textcolor{tabcyan}{\ding{108}} & \cellcolor{white}$76.22$ &\cellcolor{white} $46.51_{\pm (35.21)}$ & \cellcolor{white}$0.87_{\pm (0.10)}$ & \cellcolor{white}$3.89_{\pm (3.11)}$ \\
         & TextGrad \textcolor{tabcyan}{\ding{108}} & $75.37$ & $42.43_{\pm (19.05)}$ & $0.85_{\pm (0.07)}$ & $8.23_{\pm (9.15)}$ \\
         & \cellcolor{white} \citep{pruthi2019misspellings} \textcolor{tabred}{\ding{108}} & \cellcolor{white}$88.00$ &\cellcolor{white} $5.50_{\pm (4.74)}$ & \cellcolor{white}$0.89_{\pm (0.13)}$ & \cellcolor{white} $29.17_{\pm (25.58)}$ \\
         & \methodname{}\texttt{-Fast} \textcolor{tabred}{\ding{108}} & $\underline{95.44}$ & $\underline{3.25}_{\pm (2.88)}$ & $\underline{0.95}_{\pm (0.06)}$ & $2.38_{\pm (3.27)}$ \\
         & \cellcolor{white} \methodname{} \textcolor{tabred}{\ding{108}} &\cellcolor{white} $\mathbf{97.13}$ & \cellcolor{white}$\mathbf{2.45}_{\pm (2.31)}$ &\cellcolor{white} $\mathbf{0.97}_{\pm (0.04)}$ & \cellcolor{white}$6.94_{\pm (12.33)}$ \\
        \midrule
        \parbox[t]{3mm}{\multirow{9}{*}{\rotatebox[origin=c]{90}{MNLI-m}}} & BAE-R \textcolor{tabcyan}{\ding{108}} & $71.57$ & $6.28_{\pm (3.27)}$ & $0.84_{\pm (0.14)}$ & $0.54_{\pm (0.34)}$ \\
         & \cellcolor{white} BERT-attack \textcolor{tabcyan}{\ding{108}} &\cellcolor{white} $\underline{97.50}$ & \cellcolor{white}$7.14_{\pm (5.17)}$ &\cellcolor{white} $0.84_{\pm (0.12)}$ &\cellcolor{white} $2.96_{\pm (16.16)}$ \\
         & DeepWordBug \textcolor{tabred}{\ding{108}} & $85.90$ & $2.31_{\pm (1.63)}$ & $0.77_{\pm (0.17)}$ & $\underline{0.27}_{\pm (0.13)}$ \\
         & \cellcolor{white} TextBugger \textcolor{tabred}{\ding{108}} &\cellcolor{white} $88.05$ &\cellcolor{white} $4.86_{\pm (4.64)}$ &\cellcolor{white} $0.82_{\pm (0.13)}$ &\cellcolor{white} $0.55_{\pm (0.38)}$ \\
         & TextFooler \textcolor{tabcyan}{\ding{108}} & $94.98$ & $9.49_{\pm (6.45)}$ & $0.82_{\pm (0.13)}$ & $0.55_{\pm (0.40)}$ \\
         & \cellcolor{white} TextGrad \textcolor{tabcyan}{\ding{108}} & \cellcolor{white}$94.15$ & \cellcolor{white}$8.96_{\pm (4.90)}$ & \cellcolor{white}$0.79_{\pm (0.12)}$ &\cellcolor{white} $2.33_{\pm (1.63)}$ \\
         & \citep{pruthi2019misspellings} \textcolor{tabred}{\ding{108}} & $58.18$ & $1.26_{\pm (0.57)}$ & $0.84_{\pm (0.11)}$ & $5.14_{\pm (5.25)}$ \\
         & \cellcolor{white} \methodname{}\texttt{-Fast} \textcolor{tabred}{\ding{108}} & \cellcolor{white}$\mathbf{100.00}$ & \cellcolor{white}$\underline{1.17}_{\pm (0.42)}$ & \cellcolor{white}$\underline{0.85}_{\pm (0.13)}$ & \cellcolor{white}$\mathbf{0.22}_{\pm (0.15)}$ \\
         & \methodname{} \textcolor{tabred}{\ding{108}} & $\mathbf{100.00}$ & $\mathbf{1.08}_{\pm (0.28)}$ & $\mathbf{0.86}_{\pm (0.11)}$ & $1.53_{\pm (0.70)}$ \\
        \midrule
        \parbox[t]{3mm}{\multirow{9}{*}{\rotatebox[origin=c]{90}{QNLI}}} & \cellcolor{white} BAE-R \textcolor{tabcyan}{\ding{108}} & \cellcolor{white}$45.97$ & \cellcolor{white}$10.94_{\pm (7.57)}$ & \cellcolor{white}$\mathbf{0.95}_{\pm (0.06)}$ & \cellcolor{white}$2.52_{\pm (2.46)}$ \\
         & BERT-attack \textcolor{tabcyan}{\ding{108}} & $73.40$ & $16.74_{\pm (16.94)}$ & $0.90_{\pm (0.10)}$ & $760.09_{\pm (5904.67)}$ \\
         & \cellcolor{white} DeepWordBug \textcolor{tabred}{\ding{108}} & \cellcolor{white}$74.07$ & \cellcolor{white}$5.00_{\pm (4.36)}$ & \cellcolor{white}$0.85_{\pm (0.16)}$ & \cellcolor{white}$\mathbf{0.57}_{\pm (0.43)}$ \\
         & TextBugger \textcolor{tabred}{\ding{108}} & $76.03$ & $9.59_{\pm (11.48)}$ & $0.90_{\pm (0.10)}$ & $\underline{1.15}_{\pm (0.99)}$ \\
         & \cellcolor{white} TextFooler \textcolor{tabcyan}{\ding{108}} &\cellcolor{white} $80.72$ & \cellcolor{white}$22.56_{\pm (20.96)}$ & \cellcolor{white}$0.88_{\pm (0.11)}$ & \cellcolor{white}$2.13_{\pm (2.00)}$ \\
         & TextGrad \textcolor{tabcyan}{\ding{108}} & $74.78$ & $28.47_{\pm (17.98)}$ & $0.84_{\pm (0.09)}$ & $5.78_{\pm (6.02)}$ \\
         & \cellcolor{white} \citep{pruthi2019misspellings} \textcolor{tabred}{\ding{108}} & \cellcolor{white}$26.47$ & \cellcolor{white}$\underline{1.85}_{\pm (1.18)}$ & \cellcolor{white}$0.93_{\pm (0.09)}$ & \cellcolor{white}$10.12_{\pm (8.49)}$ \\
         & \methodname{}\texttt{-Fast} \textcolor{tabred}{\ding{108}} & $\underline{96.19}$ & $2.26_{\pm (1.74)}$ & $0.93_{\pm (0.08)}$ & $1.58_{\pm (2.36)}$ \\
         & \cellcolor{white} \methodname{} \textcolor{tabred}{\ding{108}} & \cellcolor{white}$\mathbf{96.23}$ & \cellcolor{white}$\mathbf{1.78}_{\pm (1.11)}$ & \cellcolor{white}$\underline{0.94}_{\pm (0.07)}$ & \cellcolor{white}$9.60_{\pm (8.10)}$ \\
        \midrule
        \parbox[t]{3mm}{\multirow{9}{*}{\rotatebox[origin=c]{90}{RTE}}} & BAE-R \textcolor{tabcyan}{\ding{108}} & $61.14$ & $6.69_{\pm (3.43)}$ & $\underline{0.88}_{\pm (0.09)}$ & $0.82_{\pm (0.53)}$ \\
         & \cellcolor{white} BERT-attack \textcolor{tabcyan}{\ding{108}} & \cellcolor{white}$9.80$ & \cellcolor{white}$5.20_{\pm (2.95)}$ & \cellcolor{white}$0.86_{\pm (0.16)}$ & \cellcolor{white}$21.39_{\pm (34.86)}$ \\
         & DeepWordBug \textcolor{tabred}{\ding{108}} & $59.24$ & $1.54_{\pm (0.84)}$ & $0.84_{\pm (0.13)}$ & $\mathbf{0.13}_{\pm (0.05)}$ \\
         & \cellcolor{white} TextBugger \textcolor{tabred}{\ding{108}} &\cellcolor{white} $70.62$ & \cellcolor{white}$4.45_{\pm (5.24)}$ & \cellcolor{white}$0.87_{\pm (0.11)}$ & \cellcolor{white}$0.44_{\pm (0.33)}$ \\
         & TextFooler \textcolor{tabcyan}{\ding{108}} & $68.25$ & $7.60_{\pm (5.61)}$ & $\mathbf{0.89}_{\pm (0.09)}$ & $0.52_{\pm (0.65)}$ \\
         & \cellcolor{white} TextGrad \textcolor{tabcyan}{\ding{108}} & \cellcolor{white}$70.70$ &\cellcolor{white} $7.07_{\pm (3.25)}$ &\cellcolor{white} $0.83_{\pm (0.10)}$ & \cellcolor{white}$2.56_{\pm (2.28)}$ \\
         & \citep{pruthi2019misspellings} \textcolor{tabred}{\ding{108}} & $48.34$ & $\mathbf{1.22}_{\pm (0.41)}$ & $0.86_{\pm (0.09)}$ & $11.56_{\pm (7.69)}$ \\
         & \cellcolor{white} \methodname{}\texttt{-Fast} \textcolor{tabred}{\ding{108}} & \cellcolor{white}$\underline{97.16}$ & \cellcolor{white}$1.68_{\pm (1.32)}$ & \cellcolor{white}$0.83_{\pm (0.14)}$ & \cellcolor{white}$\underline{0.42}_{\pm (0.44)}$ \\
         & \methodname{} \textcolor{tabred}{\ding{108}} & $\mathbf{100.00}$ & $\underline{1.29}_{\pm (0.65)}$ & $0.87_{\pm (0.10)}$ & $2.49_{\pm (2.13)}$ \\
        \midrule
        \parbox[t]{3mm}{\multirow{10}{*}{\rotatebox[origin=c]{90}{SST-2}}} & \cellcolor{white} CWBA \textcolor{tabred}{\ding{108}} &\cellcolor{white} $77.88$ &\cellcolor{white} $11.18_{\pm (4.58)}$ &\cellcolor{white} $0.55_{\pm (0.25)}$ &\cellcolor{white} $58.28_{\pm (50.83)}$ \\
         & BAE-R \textcolor{tabcyan}{\ding{108}} & $62.77$ & $10.25_{\pm (7.24)}$ & $0.85_{\pm (0.16)}$ & $0.78_{\pm (0.77)}$ \\
         & \cellcolor{white} BERT-attack \textcolor{tabcyan}{\ding{108}} & \cellcolor{white}$72.34$ &\cellcolor{white} $11.57_{\pm (6.89)}$ & \cellcolor{white}$0.85_{\pm (0.10)}$ &\cellcolor{white} $148.11_{\pm (1077.71)}$ \\
         & DeepWordBug \textcolor{tabred}{\ding{108}} & $84.78$ & $3.37_{\pm (2.47)}$ & $0.82_{\pm (0.16)}$ & $\mathbf{0.23}_{\pm (0.12)}$ \\
         & \cellcolor{white} TextBugger \textcolor{tabred}{\ding{108}} &\cellcolor{white} $72.52$ &\cellcolor{white} $5.61_{\pm (5.51)}$ & \cellcolor{white}$\mathbf{0.91}_{\pm (0.06)}$ &\cellcolor{white} $1.85_{\pm (0.90)}$ \\
         & TextFooler \textcolor{tabcyan}{\ding{108}} & $95.79$ & $15.79_{\pm (11.12)}$ & $0.83_{\pm (0.14)}$ & $1.11_{\pm (0.74)}$ \\
         & \cellcolor{white} TextGrad \textcolor{tabcyan}{\ding{108}} &\cellcolor{white} $96.28$ &\cellcolor{white} $18.67_{\pm (9.73)}$ & \cellcolor{white}$0.80_{\pm (0.11)}$ &\cellcolor{white} $2.95_{\pm (1.65)}$ \\
         & \citep{pruthi2019misspellings} \textcolor{tabred}{\ding{108}} & $95.05$ & $1.98_{\pm (1.28)}$ & $0.87_{\pm (0.13)}$ & $4.66_{\pm (3.98)}$ \\
         & \cellcolor{white} \methodname{}\texttt{-Fast} \textcolor{tabred}{\ding{108}} &\cellcolor{white} $\underline{99.88}$ &\cellcolor{white} $\underline{1.62}_{\pm (0.88)}$ &\cellcolor{white} $\underline{0.89}_{\pm (0.12)}$ &\cellcolor{white} $\underline{0.38}_{\pm (0.34)}$ \\
         & \methodname{} \textcolor{tabred}{\ding{108}} & $\mathbf{100.00}$ & $\mathbf{1.38}_{\pm (0.67)}$ & $\mathbf{0.91}_{\pm (0.10)}$ & $1.38_{\pm (0.93)}$ \\
        \bottomrule
    \end{tabular}
    }
    \label{tab:textattack_albert}
\end{table}

\subsection{Attack Transferability}
\label{subsec:transfer_attacks}
In this section we study the transferability of \methodname{} attacks. This is a widely studied setup in the computer vision community \citep{demontis2019adversarial}. For each dataset, attack and model, we generate the attacked sentences and evaluate the ASR when using them for attacking other models. As a reference we take the best token-level method from \cref{tab:textattack_benchmark}, i.e., TextFooler.

In \cref{tab:transfer_attacks} we can observe both TextFooler and \methodname{} fail to produce high ASRs in the transfer attack setup. As a reference, the highest transfer ASR was $55.48\%$ and was attained by \methodname{} in the MNLI-m dataset, with BERT as a Source Model and RoBERTa as the target model. We notice in the AG-News dataset it is considerably harder to produce transfer attacks, with the highest transfer ASR being $9.77\%$ among all setups. We believe improving the ASR in the transfer attack setup is an interesting avenue.

\subsection{Robust Word Recognition Defenses}
\label{app:typo_correctors}
To complete the analysis, we repeat the experiments in \cref{subsec:typo} in the RTE, MNLI-m and QNLI datasets\footnote{The AG-News dataset is not studied in \citet{pruthi2019misspellings,jones2020robencodings}.}. 
\begin{table}
    \centering
    \setlength{\tabcolsep}{0.5\tabcolsep}
    \caption{\textbf{Effect of each \texttt{PJC} constraint:} \methodname{} ASR when individually removing each constraint while keeping the rest. Performance with no constraints (\texttt{None}) put as reference. The ASR drastically increases when removing the \texttt{LowEng}, \texttt{End} or \texttt{Start} constraints, proving the fragility of existing robust word recognition defenses.}
    \vspace{2pt}
    \renewcommand{\arraystretch}{1}
    \begin{tabular}{cG|cg|cg|cg}
    \toprule
    \rowcolor{white}& & \multicolumn{2}{c|}{RTE} & \multicolumn{2}{c|}{MNLI-m} & \multicolumn{2}{c}{QNLI}\\
    \rowcolor{white} Defense  & Attack constraint & Acc. (\%) & ASR (\%)& Acc. (\%) & ASR (\%)& Acc. (\%) & ASR (\%)\\
    \midrule
    \parbox[t]{3mm}{\multirow{6}{*}{\rotatebox[origin=c]{90}{\citep{pruthi2019misspellings}}}} & \texttt{None} & \multirow{7}{*}{$60.36$} & $92.17$ & \multirow{7}{*}{$76.33$} & $100.00$ & \multirow{7}{*}{$73.55$} & $86.38$\\
    &\cellcolor{white} \texttt{PJC}  & & \cellcolor{white}$42.17$ & & \cellcolor{white}$87.39$ & & \cellcolor{white} $43.05$\\
    &~~\texttt{-LowEng} & & $70.48$& & $97.90$ && $65.53$ \\
    &\cellcolor{white}~~\texttt{-Length} && \cellcolor{white}$45.78$ && \cellcolor{white}$92.51$ && \cellcolor{white} $46.19$\\
    &~~\texttt{-End} & & $63.86$ & & $96.19$ && $57.63$\\
    &\cellcolor{white}~~\texttt{-Start} &&\cellcolor{white} $69.88$ &&\cellcolor{white} $97.11$ && \cellcolor{white} $57.08$\\
    &~~\texttt{-NoRepeat} && $42.17$ && $89.36$ && $42.51$\\
    \midrule
    \parbox[t]{3mm}{\multirow{6}{*}{\rotatebox[origin=c]{90}{\citep{jones2020robencodings}}}} & \cellcolor{white}\texttt{None} & \multirow{7}{*}{$48.74$} & \cellcolor{white} $65.93$ & \multirow{7}{*}{$68.44$} & \cellcolor{white} $100.00$ & \multirow{7}{*}{$76.20$} & \cellcolor{white} $98.56$ \\
    &\texttt{PJC}  &  & $2.96$ &  & $2.93$ &  & $1.05$ \\
    &\cellcolor{white}~~\texttt{-LowEng} & &\cellcolor{white} $57.04$ &&\cellcolor{white} $96.63$ &&\cellcolor{white} $94.62$ \\
    &~~\texttt{-Length} && $2.96$ && $2.93$ && $0.79$ \\
    &\cellcolor{white}~~\texttt{-End} & &\cellcolor{white} $44.44$ &&\cellcolor{white} $75.26$ &&\cellcolor{white} $54.99$ \\
    &~~\texttt{-Start} && $47.41$ && $82.87$ && $67.32$ \\
    &\cellcolor{white}~~\texttt{-NoRepeat} && \cellcolor{white}$6.67$ &&\cellcolor{white} $9.81$ &&\cellcolor{white} $3.81$ \\
    \bottomrule
    \end{tabular}
    \label{tab:pjc_appendix}
\end{table}

In \cref{tab:pjc_appendix} we can observe a similar phenomenon as in \cref{tab:pjc}, i.e., robust word encoding defenses only work when assuming the attacker adopts the \texttt{PJC} constraints. When either the \texttt{LowEng}, \texttt{Start} or \texttt{End} constrants are relaxed, the ASR considerably grows close to $100\%$. It is worth mentioning that in the RTE dataset, defending with \citet{jones2020robencodings} results in \methodname{} without any constraints achieving only $65.93\%$ ASR. Nevertheless, this defense degrades the clean accuracy to less than $50\%$. If the dataset is balanced, as RTE approximately is\footnote{\url{https://huggingface.co/datasets/glue/viewer/rte/validation}}, a constant classifier can achieve $50\%$ clean accuracy and $0\%$ ASR. This fact shows the little value of the RTE defended model in \citet{jones2020robencodings}.

\subsection{Attack of LLM Classifier}
\label{app:appendix_promptllm}
The prompt design in attacking different LLMs is summarized in \cref{tab:prompt}. A schematic of attacking LLMs is present in \cref{fig:schematicllama}. In this experiment, we use token-based position selection to further accelerate the process of attack. Specifically, we mask each token in the inputs and select the top ten tokens with the highest loss. Since some tokens consist of many characters, we only use 40 positions of these tokens to perform \cref{alg:position_selection}. The remaining step is the same as in \cref{alg:ours}.
Notably, in \cref{tab:llmpos}, we see that such a process can significantly accelerate the attack while maintaining the performance of ASR and other metrics. The result on Vicuna 7B is present in \cref{tab:llmclasifier_vicuna}, where we can see the proposed \methodname{} achieves much higher ASR than other baselines with less edit distance.
    \begin{figure}[!h]
    \centering
    \includegraphics[width=0.48\textwidth]{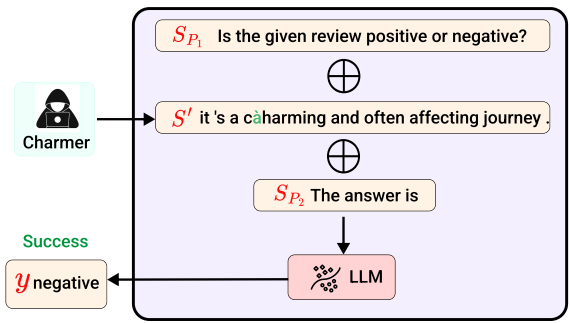}
    \caption{Schematic of the proposed \methodname{} in attacking LLM-classifiers. \methodname{} modifies the input to $S'$ with a small perturbation (annotated in green color) to the original input so that the model produces the desired output $y$. $S_{P_1}$ and $S_{P_2}$ are auxiliary prompts that remain unchanged during the attack.}
    \label{fig:schematicllama}
        \vspace{-3mm}
\end{figure}
\begin{table}[ht]
\tabcolsep=0.06cm
    \centering
    \caption{ Ablation study on the time efficiency of different methods of position selection in \llamachat{}. We choose the fast version of \methodname{} with $n=1$ and $k=10$. The result shows that combing \cref{alg:position_selection} with a token-based pre-selection procedure can notably improve the efficiency of the proposed \methodname{}.} 
\vspace{2pt}
    \resizebox{0.8\textwidth}{!}{
    \begin{tabular}{l|g|gggv}
        \toprule
         \cellcolor{white}
         & Method & \cellcolor{white}  ASR (\%)&  \cellcolor{white}   $d_{\text{lev}}(S,S')$ & \cellcolor{white}      $\text{Sim}(S,S')$ & \cellcolor{white}Time        \\
        \midrule
        \parbox[t]{2mm}{\multirow{2}{*}{\rotatebox[origin=c]{90}{SST-2}}} 
            & \methodname{}\texttt{-Fast} (Token-based \cref{alg:position_selection})  &            $95.47$ &   $\textbf{2.55}$ &  $0.83$ &   $\textbf{1.47}$ \\
            &  \cellcolor{white} \methodname{}\texttt{-Fast} (\cref{alg:position_selection}) 
           & \cellcolor{white} $\textbf{95.60}$& \cellcolor{white}	$2.56$& \cellcolor{white}	$\textbf{0.85}$& \cellcolor{white}	$3.32$
            \\   
                    \midrule
        \parbox[t]{3mm}{\multirow{2}{*}{\rotatebox[origin=c]{90}{QNLI}}} 
            & \methodname{}\texttt{-Fast} (Token-based \cref{alg:position_selection})  &          
            ${93.51}$&	${2.40}$	&${0.93}$	&$\textbf{5.66}$
            \\
            & \cellcolor{white} \methodname{}\texttt{-Fast} (\cref{alg:position_selection}) & \cellcolor{white} $\mathbf{96.82}$         
         & \cellcolor{white}	$\textbf{2.34}$& \cellcolor{white}		$\mathbf{0.94}$& \cellcolor{white} $10.30$
            \\
        \midrule
        \parbox[t]{2mm}{\multirow{2}{*}{\rotatebox[origin=c]{90}{RTE}}} 
            & \methodname{}\texttt{-Fast}(Token-based \cref{alg:position_selection})  &            ${97.10}$ &   ${1.68}$ &  $0.82$ &   $\textbf{2.06}$ \\
            & \cellcolor{white} \methodname{}\texttt{-Fast}(\cref{alg:position_selection})  & \cellcolor{white} $\textbf{98.07}$&	\cellcolor{white} $\textbf{1.64}$	& \cellcolor{white}$\textbf{0.84}$& \cellcolor{white}	$2.63$\\
        \bottomrule
    \end{tabular}}
\label{tab:llmpos}
\end{table}

\begin{table}[!ht]
\tabcolsep=0.06cm
    \centering
    \caption{\textbf{Attack evaluation in Vicuna 7B:} 
    We choose the fast version of \methodname{} with $n=1$ and $k=10$.
    \methodname{} outperforms baselines in terms of attack success rate, Levenshtein distance 
, and achieves comparative similarity and speed.}
\vspace{2pt}
\begin{tabular}{l|g|vggg}
\toprule
 \rowcolor{white}
 & Method &   ASR (\%)&     $d_{\text{lev}}(S,S')$ &      $\text{Sim}(S,S')$ &Time        \\
\midrule
\parbox[t]{2mm}{\multirow{7}{*}{\rotatebox[origin=c]{90}{QNLI}}} 
& BAE-R \textcolor{tabcyan}{\ding{108}} &        $40.66$ &  $12.36$ &  $\mathbf{0.96}$ &  $3.11$ \\
\rowcolor{white}
& BERT-attack \textcolor{tabcyan}{\ding{108}} &      $\underline{56.67}$&	$16.61$	&$0.91$	&$4.03$  \\
& DeepWordBug \textcolor{tabred}{\ding{108}} &        $43.77$ &   $\underline{3.63}$ &  $0.91$ &  $\mathbf{1.32}$ \\
   & \cellcolor{white}TextBugger \textcolor{tabred}{\ding{108}} &       \cellcolor{white} $53.11$ &   \cellcolor{white}$9.08$ &  \cellcolor{white}$0.93$ &  \cellcolor{white}$\underline{2.56}$ \\
    & TextFooler \textcolor{tabcyan}{\ding{108}} &        $51.28$ &  $20.70$ &  $0.92$ &  $4.76$ \\
 \rowcolor{white} 
    & \methodname{}\texttt{-Fast} \textcolor{tabred}{\ding{108}} &
     $ \textbf{98.35}$	&$\textbf{2.04}$&	$\underline{0.94}$&	${4.89}$ \\
\midrule
\parbox[t]{2mm}{\multirow{7}{*}{\rotatebox[origin=c]{90}{RTE}}} 
 &BAE-R \textcolor{tabcyan}{\ding{108}} &        $64.11$ &   $5.96$ &  $\mathbf{0.89}$ &  $1.23$\\
 \rowcolor{white}
    & BERT-attack \textcolor{tabcyan}{\ding{108}} &        $\underline{82.78}$ &   $8.92$ &  $0.82$ &  $1.60$ \\
   & DeepWordBug \textcolor{tabred}{\ding{108}} &            $50.97$ &   $\underline{2.67}$ &  $0.76$ &   $\mathbf{0.61}$ \\
     &\cellcolor{white} TextBugger \textcolor{tabred}{\ding{108}} &        \cellcolor{white}$71.77$ &   \cellcolor{white}$6.63$ &  \cellcolor{white}$0.84$ &  \cellcolor{white}$\underline{1.12}$ \\
    & TextFooler \textcolor{tabcyan}{\ding{108}} &        $78.47$ &   $7.94$ &  $\underline{0.86}$ &  $1.64$ \\
    \rowcolor{white}
    &\methodname{}\texttt{-Fast} \textcolor{tabred}{\ding{108}} &  
   $ \mathbf{89.05}$&	$\mathbf{1.56}$&	${0.85}$	&${1.98}$\\
\midrule
\parbox[t]{2mm}{\multirow{7}{*}{\rotatebox[origin=c]{90}{SST-2}}} 
& BAE-R \textcolor{tabcyan}{\ding{108}} &        $43.22$ &  $14.29$ &  $0.75$ &  $3.28$ \\
\rowcolor{white}
    & BERT-attack \textcolor{tabcyan}{\ding{108}} &        $32.31$ &  $13.21$ &  $0.85$ &  $2.42$ \\
   & DeepWordBug \textcolor{tabred}{\ding{108}} &        $30.72$ &   $\underline{4.22}$ &  $0.76$ &  $\mathbf{0.88}$ \\
     &  \cellcolor{white} TextBugger \textcolor{tabred}{\ding{108}} &       \cellcolor{white}   $23.01$ &    \cellcolor{white} $9.97$ &   \cellcolor{white} $\underline{0.88}$ &   \cellcolor{white} $1.73$ \\
    & TextFooler \textcolor{tabcyan}{\ding{108}} &             $\underline{64.04}$ &  $18.03$ &  $\textbf{0.91}$ &   $4.05$ \\
    \rowcolor{white}
    & \methodname{}\texttt{-Fast} \textcolor{tabred}{\ding{108}} 
    &$\mathbf{91.89}$&	$\mathbf{2.47}$	&$0.85$&	$\underline{1.66}$\\
\bottomrule
\end{tabular}
\label{tab:llmclasifier_vicuna}
\end{table}

\begin{table}[!ht]
    \centering
    \caption{\textbf{Prompting in different LLMs and datasets:} The sentences outside ``[Input]'' are considered as auxiliary prompts $S_{P_1}$ and $S_{P_2}$, as demonstrated in \cref{fig:schematicllama}.}
\begin{tabular}{l| p{1.0cm}| p{10cm}}
\toprule
  \textbf{Model} &\textbf{Dataset} &\textbf{Prompt design}      \\
\midrule
\multirow[t]{3}{*}{\llamachat{}} 
&SST-2 & Is the given review positive or negative? [Input] The answer is
\\ 
\cline{2-3} 
& RTE& [Input premise] Based on the paragraph above can we conclude the following sentence, answer with yes or no.  [Input hypothesis] The answer is 
\\ 
\cline{2-3} 
& QNLI& 
Does the sentence answer the question? Answer with yes or no. Question: [Input premise]  Sentence: [Input hypothesis]
                    The answer is
\\
\midrule
\multirow[t]{3}{*}{Vicuna 7B} 
&SST-2 & Analyze the tone of this statement and respond with either positive or negative: [Input] The answer is:
\\ 
\cline{2-3} 
& RTE& [Input premise] Based on the paragraph above can we conclude the following sentence, answer with yes or no. [Input hypothesis] The answer is 
\\ 
\cline{2-3} 
& QNLI& 
[Input premise] Based on the question above, does the following sentence answer the question? [Input hypothesis] Answer with yes or no. The answer is
\\
\hline
\end{tabular}
\label{tab:prompt}
\end{table}

\subsection{Jailbreaking LLMs}
\label{app:jailbreaking}
\textcolor{tabred}{\textbf{Disclaimer: this attack can lead to harmful content}}\\
In this section, we showcase that the proposed method can also be applied to jailbreaking LLMs, which refers to designing prompts to allow LLMs to output harmful content.
We compare against the recent jailbreaking attacks GCG \citep{zou2023universal} and AutoDAN \citep{liu2023autodan} in the AdvBench banchmark \citep{zou2023universal}. First, we overview the formulation of jailbreaking LLM. Given a harmful query $S$, the goal of jailbreaking is to make the model answer certain query $S$ even when the safety system prompt $S_{\text{system}}$ is provided, see \cref{fig:schematicllm} for an illustration of $S_{\text{system}}$. Formally, our attacker aims to find an adversarial example $S'$ based on $S$ such that:
\begin{equation*}
\begin{split}
   & \max_{S'\in \mathcal{S}_{k}(S,\Gamma)}
   \mathcal{L}\left(f(
     S_\text{system} \oplus S'
    ), y\right) :=
    \max_{S'\in \mathcal{S}_{k}(S,\Gamma)}
    \mathbb{P} (y |
     S_\text{system} \oplus S'
    )
    \label{eq:prob_llmjailbreak}
\end{split}
\end{equation*}
where $\mathcal{L}$ is defined as the probability of generating the target $ y:=\text{``Sure, here is''} \oplus S.$ For example, if $S$ is ``how to make a bomb'', then $y$ is ``Sure, here is how to make a bomb''.

In this experiment, we use the Advbench dataset proposed by \citet{zou2023universal}, which consists of harmful and toxic sentences across different topics.
Due to the costly computational budget of attacking LLMs, we only use the first 50 sentences in the dataset. We compare the proposed \methodname{} against the white-box attack GCG \cite{zou2023universal} 
and black-box attack AutoDAN \cite{liu2023autodan}.
We adopt the same hyper-parameters of LLM and criterion of success as in \cite{zou2023universal}. The result in \cref{tab:jailb} shows that \methodname{}~can pass the safety alignment process of LLMs with much less change in terms of Levenshtein distance.
\begin{table}[!ht]
\tabcolsep=0.06cm
    \centering
    \caption{Attack evaluation in jailbreaking different large language models.}
\begin{tabular}{l|g|ggg}
\toprule
 \rowcolor{white}
Model & Method &   ASR (\%)&     $d_{\text{lev}}(S,S')$ &      Time        \\
\midrule
  Vicuna 7B  & GCG &           $100.00$ &  $35.26$ &   $56.32$ \\
   \rowcolor{white}  & AutoDAN&         $100.00$  & 3677   &  -\tablefootnote{In Vicuna 7B and Guanaco 7B, AutoDAN uses the initialized handcrafted prefix in jailbreaks successfully so that the time is 0. Our method can also work on top of these handcrafted prefixes.\label{atd}} \\
    & \methodname{} &            $100.00$ &   $\mathbf{3.44}$ &   $\mathbf{17.41}$\\
\midrule
Guanaco 7B & GCG &  $100.00$   &$55.34$& $30.06$ \\
    \rowcolor{white}  & AutoDAN&   $100.00$     &  $3677.00$  &  -
    \cref{atd} 
    \\
    & \methodname{} &   $100.00$               &$\mathbf{4.98}$ & $\mathbf{24.99}$ \\
\midrule
\llamachat{}   & GCG &   86.00   &76.74  &948.44  \\
    \rowcolor{white}  & AutoDAN&       $18.00$  & $3209.33$   &  $29.01$ \\
    & \methodname{} &    94.00             & 28.02 &341.66  \\
\bottomrule
\end{tabular}
\label{tab:jailb}
\end{table}

\section{Proof of \cref{cor:Sk}}
\label{app:method}
\label{app:proof}
In this section, we provide the technical proof of \cref{cor:Sk}.
\begin{proof}
    Starting with the upper bound,
    in the base case, we have $|\mathcal{S}_0| = |\{S\}| = 1$. Then, we will prove the relationship between $|\mathcal{S}_k|$ and $|\mathcal{S}_{k-1}|$. For a certain $S' \in \mathcal{S}_{k-1}$, we have:
    \[
    \begin{aligned}
        |\mathcal{S}_{k}(S,\Gamma)| & = \left|\underset{S' \in \mathcal{S}_{k-1}}{\bigcup}\{S'': d_{\text{lev}}(S',S'')\leq 1\}\right|\\
        & \leq \sum_{S' \in \mathcal{S}_{k-1}}\left|\{S'': d_{\text{lev}}(S',S'')\leq 1\}\right|\\
        \text{\textcolor{teal}{[\cref{prop:char_d1}]}} &= \sum_{S' \in \mathcal{S}_{k-1}}\left|\left\{\psi\left(\phi\left(S'\right) \overset{i}{\leftarrow}c\right)~~\forall i\in [2|S'|+1], \forall c \in \Gamma\cup\{\xi\}\right\}\right|\\
        \text{\textcolor{teal}{[\# combinations of $i$'s and $c$'s]}}& \leq \sum_{S' \in \mathcal{S}_{k-1}}(2|S'|+1)\cdot(|\Gamma|+1)\\
        \text{\textcolor{teal}{[$|S'|\leq |S| + k~~\forall S'\in \mathcal{S}_k$]}}& \leq \sum_{S' \in \mathcal{S}_{k-1}}(2(|S| + k)-1)\cdot(|\Gamma|+1)\\
        & = (2(|S| + k)-1)\cdot(|\Gamma|+1)\cdot\left|\mathcal{S}_{k-1}\right|\,.\\
    \end{aligned}
    \]
    Finally, by induction we have
    \[
        |\mathcal{S}_{k}(S,\Gamma)| \leq (|\Gamma|+1)^{k}\cdot \prod_{j=1}^{k}(2(|S| + k)-1)\,. %
    \]
    For the lower bound, it is enough to compute the size of the set of strings obtained by adding just prefixes:%
    \[
    \begin{aligned}
        |\mathcal{S}_{k}(S,\Gamma)| & \geq \left|\left\{\psi(P \oplus \phi(S)), \forall P \in \{P' \in \mathcal{S}(\Gamma\cup\{\xi\}),~ |P'| \leq k\}\right\}\right|\\
        & = \left|\left\{\psi(P), \forall P \in \{P' \in \mathcal{S}(\Gamma\cup\{\xi\}),~ |P| \leq k\}\right\}\right|\\
        & = \left|\bigcup_{i=0}^{k}\{P' \in \mathcal{S}(\Gamma),~ |P'| = i\}\right|\\
        \text{\textcolor{teal}{[Disjoint sets]}}& = \sum_{i=0}^{k}\left|\{P' \in \mathcal{S}(\Gamma),~ |P'| = i\}\right|\\
        & = \sum_{i=0}^{k}|\Gamma|^{i}\\
        \text{\textcolor{teal}{[Geometric series]}} & = \left\{\begin{matrix}
            \frac{1-|\Gamma|^{k+1}}{1-|\Gamma|} & \text{if } |\Gamma|> 1\\
            k+1 & \text{if } |\Gamma| = 1\\
        \end{matrix}\right.
    \end{aligned}
    \]
\end{proof}

\section{Alternative Attack Designs}
\label{app:pga}
In this section, we cover alternative algorithmic designs called PGA-\methodname{} for solving \cref{eq:attack_problem}
Specifically, we study relaxing the binary constraints in \cref{eq:bp_problem} in order to perform a Projected Gradient Ascent (PGA) procedure. %

\subsection{PGA-\methodname{}}
Let $\bm{E}^{(i)} = \textbf{Token}(S^{(i)})\bm{T} \in \realnum^{l_i \times d}$ be the embeddings of tokens for any sentence $S^{(i)} \in \mathcal{S}' \subseteq \mathcal{S}(\Gamma)$ with $i \in  \left[\left|\mathcal{S}'\right|\right]$. The zero-padded embeddings for the sentences in the set $\mathcal{S}'$ become:
\[  
    \hat{\bm{E}}^{(i)} = \bm{E}^{(i)} \oplus \bm{0}_{(\overline{l} - l_i)\times d} \in \realnum^{\overline{l} \times d}, \quad  \forall i \in \left[\left|\mathcal{S}'\right|\right]\,,
\]
where $\overline{l} = \max\{l_i : i \in \left[\left|\mathcal{S}'\right|\right]\}$ and $\oplus$ is the concatenation operator along the first dimension. 
\begin{remark}[Model output after zero padding]
    Given a function %
    $f$, the output before and after zero padding is unchanged, i.e., $\bm{f}(\hat{\bm{E}}^{(i)}) = \bm{f}(\bm{E}^{(i)}) ~~ \forall \bm{E}^{(i)} \in \mathcal{S}'$.
\end{remark}
We can reformulate the problem in \cref{eq:attack_problem} as:
\begin{equation}
    \begin{matrix}
        \tag{BP}
        \underset{\bm{u} \in \realnum^{|\mathcal{S}_{k}(S,\Gamma)|}}{\max}& \mathcal{L}\left(\bm{f}\left(\sum_{i=1}^{|\mathcal{S}_{k}(S,\Gamma)|}u_i\cdot \hat{\bm{E}}^{(i)}\right), y\right)\\
        \text{s.t.} & 
            u_i \in \{0,1\} ~~\forall  i\in [|\mathcal{S}_{k}(S,\Gamma)|],\quad
            \lp{\bm{u}}{1} = 1
    \end{matrix}\,,
    \label{eq:bp_problem}
\end{equation}
which is a constrained binary optimization problem. %
Note that given $\bm{u}^{\text{BP}}$ a maximizer of \cref{eq:bp_problem} with $i^{\text{BP}} := \argmax_{i\in [|\mathcal{S}_{k}(S,\Gamma)|]}u_{i}^{\text{BP}}$, we have that the sentence $S^{(i^{\text{BP}})} \in \mathcal{S}_{k}(S,\Gamma)$ is a maximizer of \cref{eq:attack_problem}.

However, solving \cref{eq:bp_problem} is as hard as solving \cref{eq:attack_problem} because of the exponential size of $\mathcal{S}_{k}(S,\Gamma)$, see \cref{cor:Sk}. Alternatively, we can relax the binary constraints from the $\bm{u}$ vector and solve:

\begin{equation}
    \begin{matrix}
        \tag{SP}
        \max_{\bm{u} \in \realnum^{|\mathcal{S}_{k}(S,\Gamma)|}} & \mathcal{L}\left(\bm{f}\left(\sum_{i=1}^{|\mathcal{S}_{k}|}u_i\cdot \hat{\bm{E}}^{(i)}\right), y\right)\\
        \text{s.t.} & 
            u_i \in [0,1] ~~\forall  i\in [|\mathcal{S}_{k}(S,\Gamma)|], \quad
            \lp{\bm{u}}{1} = 1.
    \end{matrix}
    \label{eq:sp_problem}
\end{equation}
In this case, given $\bm{u}^{\text{SP}}$ a maximizer of \cref{eq:sp_problem}, we know:
$$
\resizebox{0.48\textwidth}{!}{$
\mathcal{L}\left(\bm{f}\left(\sum_{i=1}^{|\mathcal{S}_{k}(S,\Gamma)|}u^{\text{SP}}_i\cdot \hat{\bm{E}}^{(i)}\right), y\right) \geq \mathcal{L}\left(\bm{f}\left(\sum_{i=1}^{|\mathcal{S}_{k}(S,\Gamma)|}u^{\text{BP}}_i\cdot \hat{\bm{E}}^{(i)}\right), y\right)$}\,.
$$
Note that the embeddings $\sum_{i=1}^{|\mathcal{S}_{k}(S,\Gamma)|}u^{\text{SP}}_i\cdot \hat{\bm{E}}^{(i)}$ have no correspondence to any sentence in $\mathcal{S}_{k}(S,\Gamma)$. However, we can still take $i^{\text{SP}} = \argmax_{i\in [|\mathcal{S}_{k}(S,\Gamma)|]}u_{i}^{\text{SP}}$ and hopefully %
$S^{(i^{\text{SP}})} \in \mathcal{S}_{k}(S,\Gamma)$ is an adversarial example.
To solve \cref{eq:sp_problem}, we employ projection gradient ascent with step-size $\eta$ as follows:
\begin{equation*}
\begin{split}
   & \bm u^{t+1} = \Pi_{\Delta}  (\bm u^{t} + \eta  \nabla{\mathcal{L}_{\bm u}(\bm u^t)} ) \,,
    \end{split}
\end{equation*}
where $ \Pi_{\Delta} (\cdot)$ is the projection
function. Let us denote by $ \hat{\bm u} := \bm u^{t} +\eta  \nabla{\mathcal{L}_{\bm u}(\bm u^t)}$ for notation simplification, then the projection step essentially aims to solve the following quadratic programming problem:
\begin{equation}
\label{equ:quadratic programming}
\begin{split}
    &     \bm u^{t+1}  = \argmin_{\bm u} \frac{1}{2} || \bm u - \hat{\bm u} ||_2^2, 
    \\ & \text{subject to}:  \sum u_i = 1,\quad
    u_i \ge 0. 
\end{split}
  \tag{QP}
\end{equation}
The Lagrangian associated with \cref{equ:quadratic programming} is as follows:
\begin{equation*}
   \mathcal{L}(\bm u,\lambda, \bm v):=  \frac{1}{2} || \bm u -\hat{\bm u} ||_2^2 + 
   \lambda (\bm u^\top \bm 1 - 1)
   - 
   \bm v^\top \hat{\bm u}
   \,,
\end{equation*}
where $\lambda \in \mathbb{R}, \bm v \in \mathbb{R}^{|S_\text{all}|}$ are the Lagrange multipliers.
The Karush-Kuhn-Tucker optimality conditions are necessary and sufficient for solving \cref{equ:quadratic programming}, that is:
\begin{align}
    &  \nabla_{\bm u} \mathcal{L}(\bm u,\lambda, \bm v) 
    =  \bm u - \hat{\bm u} + \lambda \bm 1 - \bm v = \bm 0 ,
    \label{equ:kkt1}
\\ &  u_i \ge 0\,,
    \label{equ:kkt2}
\\ & \sum u_i - 1 = 0\,,
    \label{equ:kkt3}
\\ & v_i \ge 0\,,
    \label{equ:kkt4}
\\ & v_i  u_i = 0\,.
    \label{equ:kkt5}
\end{align}
Clearly, given any $\lambda$, if we set $u_i = \max(\hat{u}_i - \lambda, 0 ), \quad v_i = \max(\lambda - \hat{u}_i, 0 ),$ then \cref{equ:kkt1,equ:kkt2,equ:kkt4,equ:kkt5} can be satisfied. Therefore, the remaining problem reduces to find a $\lambda$ that satisfies \cref{equ:kkt3}, i.e.,
 \begin{equation*}
    \sum u_i - 1 = \sum  \max(\hat{u}_i - \lambda, 0 ) - 1 = 0 \,.
 \end{equation*}
 We employ the algorithm proposed in \citet{held1974validation} to solve it, as presented in \cref{alg:pga}. Lastly, we select the $\argmax_{j} (u^\star_j)$ element in  $S_{\text{all}}
$ as the attack sentence $S'$.

\begin{algorithm}
\caption{Projection into simplex \citep{held1974validation}}
\label{alg:pga}
\textbf{Input}: $\bm{\hat{u}}:= \bm u^{t} +\eta  \nabla{\mathcal{L}_{\bm u}(\bm u^t)}  \in \mathbb{R}^{|S_\text{all}|}$.
\\
Sort $\bm{\hat{u}}$ such that $\hat{u}_1 \le \hat{u}_2 \le \cdots  \le \hat{u}_{|S_\text{all}|}$.
\\
Set $J_0 := \max ({J:  \frac{
- 1 + \sum_{i=J+1}^{|S_\text{all}|}  \hat{u}_i 
}{|S_\text{all}|-J} > \hat{u}_J} )$.
\\
Calculate $\lambda = 
\frac{
- 1 + \sum_{i=J_0+1}^{|S_\text{all}|}  \hat{u}_i 
}{|S_\text{all}|-J}.
$
\\
Set $ u^{t+1}_i = \max(\hat{u_i} - \lambda,0)$.
\\
\textbf{Output}: $\bm u^{t+1}$
\end{algorithm}
\subsection{Comparison Between PGA-\methodname{} and Query-based \methodname{}}
In this section, we experimentally validate the efficiency of PGA-\methodname{} and compare it against query-based \methodname{}, which is proposed in the main body. The result in \cref{tab:pga} shows that PGA-\methodname{} can efficiently reduce the runtime as it does not require the forward pass over a mini-batch of sentences after position selection.
However,  PGA-\methodname{} is worse than~\methodname{} 
 on other metrics, e.g., ASR, Levenshtine distance and similarity are degraded. We believe that combining the efficiency of PGA-\methodname{}  and high ASR in -\methodname{}  holds promise for future research endeavors.

\begin{table}[t]
    \centering
    \caption{\textbf{Comparison between our PGA-\methodname{}  and query-based \methodname{} (proposed in the main body) in BERT:} The best method is highlighted in \textbf{bold}. While the PGA-\methodname{} strategy can noticeably improve the runtime, the ASR, Levenshtine distance and similarity are considerably degraded.}
    \begin{tabular}{cg|vggg}
        \toprule
        \rowcolor{white}
        & Method & ASR (\%) $\bm{\uparrow}$ & $d_{\text{lev}}(S,S')~ \bm{\downarrow}$ & $\text{Sim}(S,S')~ \bm{\uparrow}$ & $\text{Time (s)} ~ \bm{\downarrow}$ \\
        \midrule
        \multirow{2}{*}{AG-News} & PGA-\methodname{} & $86.94$ & $7.33_{\pm (5.01)}$ & $0.87_{\pm (0.11)}$ & $\mathbf{8.15}_{\pm (7.04)}$ \\
         & \cellcolor{white} \methodname{} & \cellcolor{white} $\mathbf{98.51}$ & \cellcolor{white} $\mathbf{3.68}_{\pm (3.08)}$ & \cellcolor{white} $\mathbf{0.95}_{\pm (0.06)}$ & \cellcolor{white} $8.74_{\pm (11.10)}$ \\
        \midrule
        \multirow{2}{*}{MNLI-m} & PGA-\methodname{} & $99.05$ & $2.11_{\pm (1.53)}$ & $0.79_{\pm (0.19)}$ & $\mathbf{0.85}_{\pm (0.72)}$ \\
         & \cellcolor{white} \methodname{} & \cellcolor{white} $\mathbf{100.00}$ & \cellcolor{white} $\mathbf{1.14}_{\pm (0.42)}$ & \cellcolor{white} $\mathbf{0.85}_{\pm (0.13)}$ & \cellcolor{white} $1.45_{\pm (0.81)}$ \\
        \midrule
        \multirow{2}{*}{QNLI} & PGA-\methodname{} & $81.19$ & $3.46_{\pm (2.32)}$ & $0.89_{\pm (0.12)}$ & $\mathbf{5.15}_{\pm (4.60)}$ \\
         & \cellcolor{white} \methodname{} & \cellcolor{white} $\mathbf{97.68}$ & \cellcolor{white} $\mathbf{1.94}_{\pm (1.48)}$ & \cellcolor{white} $\mathbf{0.94}_{\pm (0.07)}$ & \cellcolor{white} $9.19_{\pm (9.60)}$ \\
        \midrule
        \multirow{2}{*}{RTE} & PGA-\methodname{} & $72.64$ & $\mathbf{1.53}_{\pm (1.45)}$ & $\mathbf{0.86}_{\pm (0.11)}$ & $\mathbf{0.65}_{\pm (0.71)}$ \\
         & \cellcolor{white} \methodname{} & \cellcolor{white} $\mathbf{97.01}$ & \cellcolor{white} $1.55_{\pm (1.42)}$ & \cellcolor{white} $\mathbf{0.86}_{\pm (0.13)}$ & \cellcolor{white} $2.50_{\pm (2.33)}$ \\
        \midrule
        \multirow{2}{*}{SST-2} & PGA-\methodname{} & $97.52$ & $2.68_{\pm (1.82)}$ & $0.84_{\pm (0.16)}$ & $\mathbf{1.09}_{\pm (0.89)}$ \\
         & \cellcolor{white} \methodname{} & \cellcolor{white} $\mathbf{100.00}$ & \cellcolor{white} $\mathbf{1.47}_{\pm (0.74)}$ & \cellcolor{white} $\mathbf{0.90}_{\pm (0.11)}$ & \cellcolor{white} $1.27_{\pm (0.84)}$ \\
        \bottomrule
    \end{tabular}
    \label{tab:pga}
\end{table}

\end{onecolumn}
\end{document}